\documentclass[acmlarge, authorversion=true, nonacm=true]{acmart}

\usepackage{multirow}
\usepackage{soul}
\usepackage{tabu}

\AtBeginDocument{%
  }

\begin{document}

\title{WEAR: An Outdoor Sports Dataset for Wearable and Egocentric Activity Recognition}

\author{Marius Bock}
\email{marius.bock@uni-siegen.de}
\orcid{1234-5678-9012}
\affiliation{%
  \department{Ubiquitous Computing}
  \department{Computer Vision}
  \institution{University of Siegen}
  \city{Siegen}
  \country{Germany}
}

\author{Hilde Kuehne}
\orcid{0000-0003-1079-4441}
\email{h.kuehne@uni-tuebingen.de}
\affiliation{%
  \department{Multimodal Learning}
  \institution{University of Tuebingen}
  \city{Tuebingen}
  \country{Germany}
}

\author{Kristof Van Laerhoven}
\orcid{0000-0001-5296-5347}
\email{kvl@eti.uni-siegen.de}
\affiliation{%
  \department{Ubiquitous Computing}
  \institution{University of Siegen}
  \city{Siegen}
  \country{Germany}
}

\author{Michael Moeller}
\orcid{0000-0002-0492-6527}
\email{michael.moeller@uni-siegen.de}
\affiliation{%
  \department{Computer Vision}
  \institution{University of Siegen}
  \city{Siegen}
  \country{Germany}
}

\begin{abstract}
  Research has shown the complementarity of camera- and inertial-based data for modeling human activities, yet datasets with both egocentric video and inertial-based sensor data remain scarce. In this paper, we introduce WEAR, an outdoor sports dataset for both vision- and inertial-based human activity recognition (HAR). Data from 22 participants performing a total of 18 different workout activities was collected with synchronized inertial (acceleration) and camera (egocentric video) data recorded at 11 different outside locations. WEAR provides a challenging prediction scenario in changing outdoor environments using a sensor placement, in line with recent trends in real-world applications. Benchmark results show that through our sensor placement, each modality interestingly offers complementary strengths and weaknesses in their prediction performance. Further, in light of the recent success of single-stage Temporal Action Localization (TAL) models, we demonstrate their versatility of not only being trained using visual data, but also using raw inertial data and being capable to fuse both modalities by means of simple concatenation. The dataset and code to reproduce experiments is publicly available via: \url{mariusbock.github.io/wear/}.
\end{abstract}




\begin{teaserfigure}
    \centering
    \includegraphics[width=.95\linewidth]{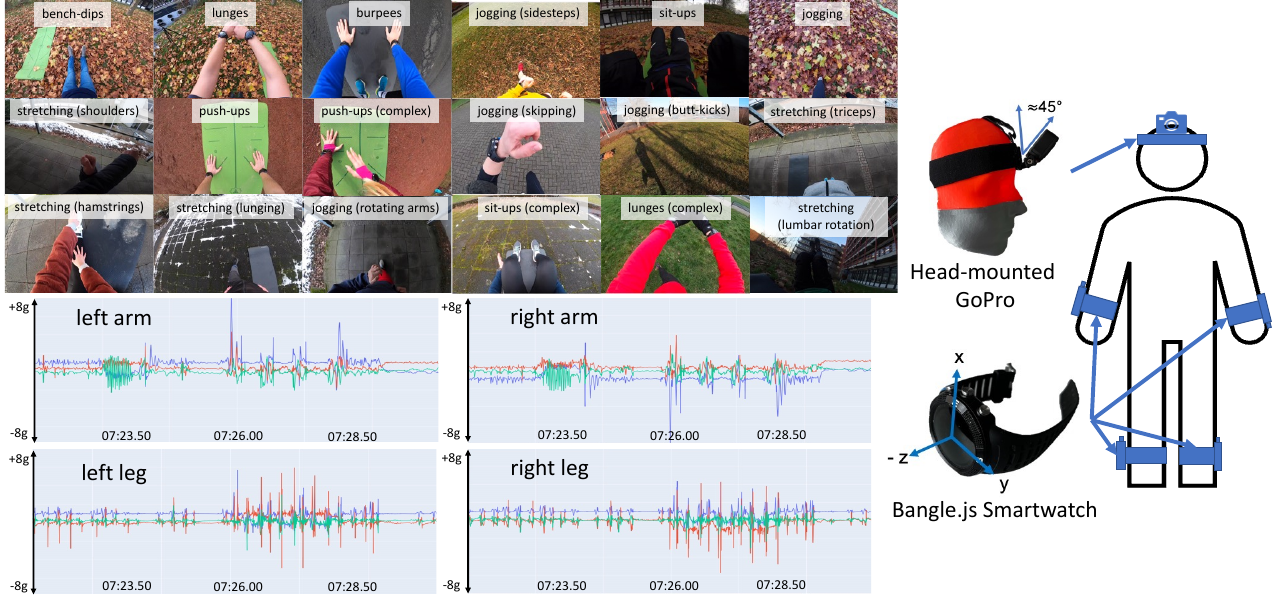}
    \caption{
    Overview of the sensor setup and example data of the two types of wearable sensors employed in our WEAR dataset. 22 participants each performed 18 activities at 11 different outdoor locations while being equipped with four open-source smartwatches (one per limb) and a head-mounted camera.}
    \label{fig:banner}
    \Description{Visualization of the different data types found in the WEAR dataset. The figure shows sample images of each activity from different participants captured by the egocentric camera as well as sample data from the four inertial sensors worn on each participant's limbs.
   }
  \end{teaserfigure}

\maketitle

\section{Introduction \& Motivation}
The physical activities that we perform in our daily lives have been identified as valuable information for a number of research fields and applications %, such as work processes support, preventive healthcare, cognitive science or workout monitoring 
\cite{baoActivityRecognitionUserAnnotated2004, pattersonFineGrainedActivityRecognition2005, wardActivityRecognitionAssembly2006}. Research efforts have shown that physical activities can be detected using either wearable inertial sensors or camera-based approaches. The inertial sensors can continuously observe motion and gestures at particular body locations, whereas camera-based systems can typically observe the user's entire body, but can be hindered by (self-)occlusions. Inertial data manifests itself as  multidimensional time series, but is unlike image data almost impossible to be interpreted by a human annotator after recording. Even though research has shown that both modalities are complementary to each other \cite{spriggsTemporalSegmentationActivity2009}, available benchmark datasets that provide both camera and inertial-based sensor data remain scarce. 
We therefore introduce WEAR, an outdoor sports human activity recognition (HAR) dataset featuring workout activities performed by 22 participants while wearing inertial sensors on both wrists and ankles as well as a head-mounted camera capturing egocentric vision - see Figure~\ref{fig:banner}. 
Unlike previous egocentric datasets \cite{graumanEgo4DWorld0002022, damenRescalingEgocentricVision2022, graumanEgoExo4DUnderstandingSkilled2024}, WEAR provides an inertial sensor placement which is in line with recent trends in real-world application scenarios, providing complementary information to the captured video stream. Further, amongst egocentric datasets which utilize limb-placed IMU sensors, WEAR is the first dataset being collected at different outdoor recording locations, with each location introducing different visual and surface conditions, yet not providing cues about the activity being performed.
Our contributions in this paper are three-fold:
\begin{enumerate}
    \item We introduce a new inertial- and vision-based HAR dataset, called WEAR, consisting of data from 22 participants performing 18 different outdoor sports activities.
    \item Results of our performed benchmark analysis reveal that the application-driven sensor placement along with the challenging outdoor recording scenario makes WEAR an a suitable dataset to assess methods on how to combine strengths of both inertial- and vision-based approaches.
    \item We demonstrate that state-of-the-art TAL models from computer vision are particularly suited to not only process raw inertial data, but even successfully fuse multi-modal information significantly outperforming the best single-modality approach as well as beating the best possible (oracle) late fusion approach in terms of mAP.
\end{enumerate}

\section{Related Work}
\label{sec:related_work}

\subsection{Inertial- and Video-based HAR datasets} In Table~\ref{tab:datasets} we show a curated list of datasets which provide both egocentric vision- (e.g. RGB, depth) and IMU-based (e.g. accelerometer, gyroscope, magnetometer) modalities in the context of HAR. We compare datasets regarding their recency, size, number of participants, number and type of activities performed, recording environment, limb placement position of IMU sensors and whether the dataset is provided on a clip-basis or a continuous stream. The three largest multi-modal corpuses currently available, the Ego4D \cite{graumanEgo4DWorld0002022}, EPIC-Kitchens \cite{damenRescalingEgocentricVision2022} and Ego-Exo4D \cite{graumanEgoExo4DUnderstandingSkilled2024} datasets only provide inertial data from the built-in IMU of the head-worn camera. Given the captured inertial data thus only decodes the head movement of participants, this position of an inertial sensor is likely to provide only limited complementary information to the egocentric vision modality. As evident by the rise in popularity of commercial head-mounted cameras and wrist-worn smartwatches for tracking sports, we thus decided to position the camera and IMU sensors used during collection of the WEAR dataset in line with the recent trends in real-world application scenarios. With the head and limbs being positions which do not limit participants in their freedom of movement, we deem said positions to further be most suited in capturing how participants interact with their environment and/ or objects. Further note that with the intended audience of the datasets residing mostly in the vision-based community, the CMU-MMAC \cite{delatorrefernandoDetailedHumanData2009}, Ego4D, Ego-Exo4D and EPIC-Kitchens datasets are focusing on providing activity annotations following a vision-centric definition, e.g. \textit{pick up object X}, or are close to a narration-style annotation, e.g. \textit{adjust the stove heat}. Though these type of labels are differentiable from a vision-side, they are near impossible to distinguish using body-worn sensors. To nevertheless give an estimate of activities present in the dataset, we provide activity verb counts (Ego4D and EPIC-Kitchens) and high-level activity counts (Ego-Exo4D) in Table~\ref{tab:datasets}.
Our chosen sensor placement thus makes the CMU-MMAC \cite{delatorrefernandoDetailedHumanData2009}, ADL \cite{dieteVisionAccelerationModalities2019} and ActionSense \cite{delpretoActionSenseMultimodalDataset2022} dataset most comparable to the WEAR dataset. Compared to these three datasets, the WEAR dataset is the largest in size and participant count. Furthermore, the three existing benchmark datasets were all recorded in a non-changing indoor (kitchen) environment, which, in case of the ActionSense and CMU-MMAC dataset, was artificially created in a laboratory. Unlike the outdoor setup of the WEAR dataset, the indoor setup thus limits each dataset's amount of variety captured in the visual data, as lighting conditions and surroundings remain the same throughout all participants. Lastly, as activities are mostly object-centric activities (e.g. cooking recipes) and action is mostly taking place in the POV of the user, prediction scenarios of the three datasets are more biased towards vision rather than inertial data. 

\begin{table}
  \small
  \centering
  \caption{List of available egocentric vision datasets, which provide inertial data, compared with the WEAR dataset. We differentiate between number (\#Class) and type of activity classes (S = Sports, D = Daily Living, C = Cooking), number of participants (\#Participants), recording environment (outside or inside), limb placement of IMU sensors (LW = left wrist, RW = right wrist, LA = left ankle, RA = right ankle) and type of recorded videos (T = trimmed or U = untrimmed video sequences). *Subset which provides both IMU and activity annotations **Exact participant counts are not provided; labels summarize key-steps $^\dag$ Activity verb classes. $^\ddag$ Excludes calibration periods. }
  \label{tab:datasets}
  \begin{tabular}{@{}lcccccccccccc@{}}
    Dataset &  \multicolumn{4}{c}{General} & & \multicolumn{4}{c}{Limb Placement} & & \\ \cmidrule{2-5} \cmidrule{7-10}
     & \#Participants & \#Class & \#Hours & Type & Where & LW & RW & LA & RA & Video \\
    \midrule
    MEAD \cite{songEgocentricActivityRecognition2016}               & 2    & 20          & \textless1 & D,S & In,Out   & & & & & T \\
    DataEgo \cite{possasEgocentricActivityRecognition2018}           & $\approx$10 & 20          & 4 & D,S & In,Out   & & & & & U\\
    Ego4D \cite{graumanEgo4DWorld0002022}                            & $\approx$60  & 87$^\dag$         & 160$^*$ & D,S,C & In,Out   & & & & & T \\
    Ego-Exo4D \cite{graumanEgoExo4DUnderstandingSkilled2024}                                          & $\leq$555$^{**}$  & 689$^{**}$         & 68$^*$ & D,S,C & In,Out   & & & & & T \\
    EPIC-Kitchens \cite{damenRescalingEgocentricVision2022}          & 37   & 97$^\dag$          & 100 & C & In       & & & & & U \\
    UESTC-MMEA-CL \cite{xuContinualEgocentricActivity2023}           & 10   & 32          & 30 & D & In,Out   & & & & & T \\
    CMU-MMAC \cite{delatorrefernandoDetailedHumanData2009}           & 13   & 16$^\dag$          & 2 & C & In      & \checkmark & \checkmark & \checkmark & \checkmark & U  \\
    ADL Dataset \cite{dieteVisionAccelerationModalities2019}         & 2    & 6           & \textless1 & D & In       & \checkmark & \checkmark & & & U \\
    ActionSense \cite{delpretoActionSenseMultimodalDataset2022}      & 9   & 20          & 9$^\ddag$ & C & In      & \checkmark & \checkmark & \checkmark & \checkmark & U \\
    \bottomrule
    \textbf{WEAR} & \textbf{22} & \textbf{18} & \textbf{19} & \textbf{S} & \textbf{Out} & \textbf{\checkmark} & \textbf{\checkmark} & \textbf{\checkmark} & \textbf{\checkmark} & \textbf{U} \\
  \end{tabular}
\end{table}

\subsection{Inertial-based HAR}
Compared to video-based modalities body-worn sensor systems bear a great potential in analyzing our daily activities with minimal intrusion, yielding various applications from the provision of medical support to supporting complex work processes \cite{bullingTutorialHumanActivity2014}. Within the last decade deep learning based-methods have established themselves as the de facto standard in inertial-based HAR as they have shown to outperform classical machine learning algorithms \cite{ordonezDeepConvolutionalLSTM2016, hammerlaDeepConvolutionalRecurrent2016, guanEnsemblesDeepLSTM2017}. One of the most well-known deep learning approaches for inertial-based HAR is the \emph{DeepConvLSTM} which is a hybrid model combining both convolutional and recurrent layers \cite{ordonezDeepConvolutionalLSTM2016}. By combining both types of layers the network is able to automatically extract discriminative features and model temporal dependencies. Following the success of the original DeepConvLSTM, researchers worked on extending the architecture \cite{murahariAttentionModelsHuman2018, xiDeepDilatedConvolution2018} or build up on the idea of combining convolutional and recurrent layers by proposing their own architectures \cite{xuInnoHARDeepNeural2019, abedinAttendDiscriminateStateoftheart2021, yukiActivityRecognitionUsing2018, zhouTinyHARLightweightDeep2022}.
Within this publication we are reporting benchmark scores using the WEAR dataset inertial sensor-streams as input for two popular HAR models \cite{bockImprovingDeepLearning2021, abedinAttendDiscriminateStateoftheart2021}. 
Contrary to the belief that one needs to employ multiple recurrent layers when dealing with sequential data \cite{karpathyVisualizingUnderstandingRecurrent2015}, \citet{bockImprovingDeepLearning2021} proposed an altered \emph{shallow DeepConvLSTM} architecture which proved to outperform the original architecture by a significant margin.
Differently, \citet{abedinAttendDiscriminateStateoftheart2021} chose to build up on the idea of the DeepConvLSTM and introduced the \emph{Attend-and-Discriminate} architecture which exploits interactions among different sensor modalities by introducing self-attention through a cross-channel interaction encoder and adding attention to the recurrent parts of the network.

\subsection{Vision-based HAR}
Predicting activities performed by humans based on visual-cues can broadly be categorized into three main application scenarios: action recognition, localization and anticipation. Action recognition systems \cite{liuSwinTransformerHierarchical2021, wangActionCLIPNewParadigm2021, liMViTv2ImprovedMultiscale2022} aim to assign a set of trimmed action segments an activity label. Contrarily, TAL systems \cite{zhangActionFormerLocalizingMoments2022, yangBasicTADAstoundingRGBOnly2023, liuEndtoendTemporalAction2022} are tasked to identify start and end times of all activities in a untrimmed video by predicting a set of activity triplets \textit{(start, end, activity label)}. Lastly, action anticipation systems \cite{girdharAnticipativeVideoTransformer2021, royPredictingNextAction2022} aim to predict the label of a future activity having observed a segment preceding its occurrence. Though sensor-based HAR systems are employed using a sliding window approach and thus assign activity labels to a set of trimmed inertial-sequences, their ultimate goal is to identify a set of activities within a continuous timeline. We therefore deem vision-based TAL to be most comparable to inertial-based HAR and will focus on it in our benchmark analysis.
Existing TAL methods can be divided into two categories: two- and single-stage approaches. Two-stage approaches 
\cite{linBMNBoundarymatchingNetwork2019, linFastLearningTemporal2020, xuGTADSubgraphLocalization2020, baiBoundaryContentGraph2020, zhaoBottomupTemporalAction2020, zengGraphConvolutionalNetworks2019, gongScaleMattersTemporal2020, liuMultishotTemporalEvent2021, qingTemporalContextAggregation2021, sridharClassSemanticsbasedAttention2021, zhuEnrichingLocalGlobal2021, zhaoVideoSelfstitchingGraph2021, tanRelaxedTransformerDecoders2021} divide the process of TAL into two subtasks. First, during the action segment proposal generation, candidate video segments are generated which are then, classified with an activity label as well as refined regarding their temporal boundaries. Contrarily, single-stage approaches \cite{yangBasicTADAstoundingRGBOnly2023, shiReActTemporalAction2022, nagProposalfreeTemporalAction2022, liuEndtoendTemporalAction2022, liuProgressiveBoundaryRefinement2020, longGaussianTemporalAwareness2019, linLearningSalientBoundary2021, chenDCANImprovingTemporal2022, zhangActionFormerLocalizingMoments2022, shiTriDetTemporalAction2023} aim to localize actions in a single shot without using action proposals. 

In light with the success of transformer architectures in natural language processing \cite{vaswaniAttentionAllYou2017, devlinBERTPretrainingDeep2019} and computer vision \cite{kolesnikovImageWorth16x162021, yuanTokenstotokenViTTraining2021, liuSwinTransformerHierarchical2021}, researchers have demonstrated their applicability for TAL \cite{chengTallFormerTemporalAction2022, liuEmpiricalStudyEndtoend2022, liuEndtoendTemporalAction2022, shiReActTemporalAction2022, tanRelaxedTransformerDecoders2021, zhangActionFormerLocalizingMoments2022} breaking previously held benchmark scores of numerous popular datasets \cite{heilbronActivityNetLargescaleVideo2015, damenRescalingEgocentricVision2022, jiangTHUMOSChallengeAction2014} without any additional training data by a significant margin.
One of such architectures is the \emph{ActionFormer} proposed by \citet{zhangActionFormerLocalizingMoments2022}, which is an end-to-end trainable transformer-based architecture, which unlike other single-stage approaches, does not rely on pre-defined anchor windows. The architecture combines multiscale feature representations with local self-attention and is trained through a classification and regression loss calculated by a light-weighted decoder. Building up on ActionFormer architecture, \citet{shiTriDetTemporalAction2023} proposed the \emph{TriDet} model which suggest to replace the transformer layers of the ActionFormer with fully-convolutional, so-called SGP layers, as well as use a trident regression head which claims to improve imprecise boundary predictions via an estimated relative probability distribution around the boundary. Given the rapid rise in popularity of single-stage TAL such as the ActionFormer, we decided said models to be a suited option to deliver a first benchmark for the WEAR dataset.  

\subsection{Multimodal (Inertial and RGB Video) HAR}
With early works such that of \citet{spriggsTemporalSegmentationActivity2009} having shown the complementarity of inertial- and camera-based features, research has followed up by exploring different ways of combining the two modalities. One can categorize such methods broadly by the point in time at which the fusion of both modalities is performed. Late fusion approaches usually follow a two-stream architecture training both vision- and inertial-based modalities separately before merging together outputs of each stream through such as produced softmax probabilities e.g. via a weighted combination \citep{weiSimultaneousUtilizationInertial2020}, attention mechanisms \cite{gaoMMTSAMultiModalTemporal2023}, pooling operations \citep{songMultimodalMultistreamDeep2016, imranMultimodalEgocentricActivity2020}, majority voting \citep{dieteImprovingMotionbasedActivity2018}, a concurrent classifier \citep{wuAnticipatingDailyIntention2017, dieteFusingObjectInformation2019, ijazMultimodalTransformerNursing2022, strombackMMFitMultimodalDeep2020} or knowledge transfer between separate networks \cite{raduVision2SensorKnowledgeTransfer2019}. Early fusion approaches aim at jointly learning from both modalities by using feature embeddings calculated on one (or both) modalities to e.g. use the concatenation of both to train a concurrent network \citep{imranEvaluatingFusionRGBD2020, xuContinualEgocentricActivity2023, nakamuraJointlyLearningEnergy2017, luHumanActivityClassification2018, huMultimodalHumanActivity2023, ehatisham-ul-haqRobustHumanActivity2019, dieteFusingObjectInformation2019, dieteVisionAccelerationModalities2019, songEgocentricActivityRecognition2016, yuHierarchicalDeepFusion2019, chenFusionDepthSkeleton2016, islamMuMuCooperativeMultitask2022, islamMultiGATGraphicalAttentionbased2021}, enhance softmax probabilities used during late fusion \citep{dieteFusingObjectInformation2019, dieteVisionAccelerationModalities2019} or adding intermediate cross-view connections amongst the two modality streams \citep{ijazMultimodalTransformerNursing2022}. 
With experiments showing that single-stage TAL models are able to produce competitive results on raw inertial data, this paper also tests the applicability of two state-of-the-art models, namely the ActionFormer and TriDet model, to fuse and combine cues of both modalities in an early-fusion style. Unlike other early fusion techniques, our approach is the first to directly use the raw inertial data by means of simple concatenation together with a vision-based feature embedding. 

\section{Methodology}
\label{sec:methodology}

\subsection{Study Design \& Scalable Pipeline}
\label{subsec:datacollection}

Participants of the WEAR dataset were recorded during separate recording sessions. Prior to their first session, participants were handed a recording plan which outlined the study protocol as well informed about any risks of harm, data collection, usage, anonymisation and publication, as well as how to revoke their data usage rights at any point in the future. The study design involving human participants was reviewed and approved by the University of Siegen (reference number: 03/2023 VVT). All participants were briefed and provided their written informed consent. Each participant was asked to perform all 18 workout activities detailed in the recording plan. The location and the time of day at which the sessions were performed, were not fixed and thus vary across participants (see Section~B in the supplementary material). Participants were suggested to follow a two-session setup, i.e. 9 activities per session. Nevertheless, it was allowed to differ from this setup and split the 18 activities across as many (or as few) sessions as participants liked. This caused the amount of recording sessions to vary across participants, but also increased the amount of captured variability in weather conditions and recording locations. In order to avoid misunderstandings in the execution of the activities, the authors discussed all activities prior to each session and encouraged participants to ask questions during the session if something remained unclear. Participants were tasked to perform each activity for roughly 90 seconds. As activities varied in their intensity, it was not required to perform activities for 90 seconds straight and participants could include breaks as needed. Furthermore, to ensure that each participant was able to perform all workout activities properly, the recording plan detailed how activities could be altered in their execution, for instance so that they required less physical strength. 

\subsection{Experimental Protocol}

In order to properly explain to participants the activities they needed to perform and give insights on the overall study design a recording plan (see Section~D in the supplementary material) was provided to participants prior to their first session. The recording plan details all necessary materials and is written in such a way that the can easily be reproduced by persons other than the authors. The plan further outlines the study protocol as well informs about any risks of harm, data collection, usage, anonymisation and publication, as well as how to revoke data usage rights at any point in the future. Besides a written description of each activity, the original document provides short video-clips of each activity, showing the correct execution of exercises. To avoid any misunderstandings, the participants further received a one-on-one session with the researchers being able to ask their questions about the plan and activities in it. 

The recording plan provided with our dataset includes all necessary materials and is written in such a way that all activities and sessions can easily be reproduced by persons other than the authors. Besides the used sensors for video and acceleration recording, the exercises only require a yoga mat and a chair (or similar items). Sessions can be recorded at any location outside as long as the privacy of the participants and bystanders is ensured. The code repository of the dataset provides a detailed guide on how to work with the equipment and how to collect, postprocess and annotate newly collected data. We argue that this facilitates reproducibility, and with a minimal setup ensures that it is possible for others to extend our dataset at a later date.

\subsection{Participant Information}

We recorded data for 22 participants (13 male, 9 female) at 11 different locations and under varying weather conditions. The first 18 participants were recorded over a stretch of 5 months (October to February), totalling more than 15 hours. In an effort to provide a test set along with the WEAR dataset an additional 6 participants were recorded, totalling around 4 hours of additional data. To allow researchers to explore also personalised prediction approaches, the test dataset includes two participants which were already part of the first recordings, and which volunteered to complete the workout an additional time. Recordings of the test dataset took place during spring (new participants) as well as summer (re-recordings). Egocentric video data of the four new participants was recorded using a GoPro Hero 11 as opposed to a Hero 8 and two new participants were recorded at a previously unseen location. On average each participant contributed roughly 50 minutes of data. At the time of recording, the participants had a mean age of 27.59 years (standard deviation (SD): 4.69), a mean height of 175.2 cm (SD: 9.83), and a mean weight of 69.74 kg (SD: 11.23). In order to assess their sports level, participants filled in a post-session questionnaire. The questionnaire contained questions related to vital information (such as body height, weight and age), weekly workout frequency (min. 15 minutes duration) and experience in particular workout activities. The participants in the study worked out for a mean of 3.73 times per week (SD: 2.31), were familiar with a mean of 15 out of the 18 activities in advance (SD: 3.57), and regularly performed a mean of 5.71 of the recorded activities (SD: 4.14) as part of their private workouts. Participants reported for their personal workout schedules a wide-range of cardio- (running, hiking, cycling, dancing), strength- (weight lifting, freeletics, rowing, bouldering), team- (volleyball, basketball, (table-)tennis) and flexibility-focused (yoga, ballet) exercise types. Individual answers of each participant can be found in the supplementary material (see Section~B).

\subsection{Dataset Collection \& Structure}

The WEAR dataset provides participant-wise raw and processed acceleration and egocentric-video data (see Figure~\ref{fig:banner}). We focus on 3D accelerometers especially as they cover a substantial amount of commercial fitness devices worn at the wrists and ankles. They furthermore are used in a large set of existing research and datasets focusing on wearable data for activity recognition, and they do not suffer from noise, drift, and other device-specific characteristics. 3D accelerometer data was collected at 50~Hz with a sensitivity of $\pm$ 8g using four open-source Bangle.js smartwatches running a custom, open-source firmware \cite{vanlaerhovenValidationOpensourceAmbulatory2022}. The watches were placed by the researchers in a fixed orientation on the left and right wrists and ankles of each participant. Egocentric video data was captured using either a GoPro Hero 8 or Hero 11 action camera, which was mounted using a head strap on each participant's head. The resulting `.mp4'-videos were recorded at 1080p resolution with 60 frames per second and the camera being tilted downwards in a 45 degree angle. A second tripod-mounted camera was placed within the proximity of each participant to facilitate annotation {recording the environment in which the workout was performed from a third-person-perspective. Using again a large FOV setting, the second camera was placed in a way such that as much area as possible was captured. To allow for even more freedom of movement, participants were allowed to move out of the FOV of the second camera, but were asked to start and end their activities within the camera's FOV. This allowed participants, especially during running exercises, to run straight distances and overall commence activities in a more natural way. For privacy reasons, the second camera's video and all audio captured are not part of the WEAR dataset. 

The open-source firmware \cite{vanlaerhovenValidationOpensourceAmbulatory2022} running on each Bangle.js smartwatch stores the lossless, delta-compressed inertial data in separate files on the internal memory of each watch. During post-processing, said compressed files were extracted, uncompressed and concatenated to a single `.csv'-file per session. Being a common issue with accelerometers sampling at a high sampling rate, the Bangle.js smartwatch is not able to maintain an exact sampling rate of 50 Hz at all times throughout the experiment, with the true sampling rate being closer to 48 Hz with fluctuations ranging between $\pm$ 1 Hz. The firmware provides for each file a timestamp that was set by the on-board real-time clock, which allows correcting individual times of all delta-compressed samples. Therefore, in order to obtain the true sampling rate and correct the timestamps of the concatenated `.csv'-file, synchronisation jumps were performed by each participant at the start and end of each session. The synchronization jumps involved participants move in front of the tripod-mounted camera, stand still for approximately 10 seconds, jump three times along with raising the arms while jumping, and stand still for another 10 seconds. This allowed to map peaks in the inertial sensor streams to be mapped to points in the video stream and thus obtain a start and end point within both modality data streams. Lastly, assuming recorded inertial data records are equidistant, all records within the span of the start and end-point were evenly distributed across the experiment's duration and, as a final step, resampled to have a sampling rate of 50 Hz via linear interpolation. Similar to the inertial data, the video data recorded by the head-mounted GoPro was not recording at a true frame rate of 60 FPS, but slightly deviated from that (i.e. 59.94 FPS). We therefore also resampled the egocentric videos to be of a frame rate of 60 FPS. 

In order to validate our synchronization process we made use of the similarities between sensor and audio data and converted each axis of the 3D accelerometer as well as their combined magnitude to four separate `.wav'-files. This approach is inspired by the works of \citet{schollMultimediaExchangeFormat2019} and \citet{binmorshedPersonalizedApproachDeveloping2022}. We calculated the magnitude as the summed norm of each individual inertial sensor channels, i.e. $\sqrt{x^2 + y^2 + z^2}$ with $x$, $y$ and $z$ being the x-, y- and z-axis of the 3D accelerometer data. Having converted the `.csv'-data to `.wav'-files allowed us to import both video data and inertial data into a standard video editing tool, in our case we used Final Cut Pro (see Figure~\ref{fig:finalcut}). The user interface of Final Cut offers to see previews of sound files being in our case equivalent to a graph-like visualization of the acceleration data. This allowed us to verify our applied synchronization during annotation throughout the whole duration of each participant’s session. On average, the combined magnitude proved to be most useful when verifying the correctness of our synchronization across time. Labels of the activities were added by a single expert annotator as subtitles in `.srt'-format. A final script then converted the exported `.srt'-file to `.csv'-format, filling gaps within the subtitles with a \textit{NULL} label and appended this to the respective final inertial sensor data `.csv'-file.

\begin{figure}
\begin{center}
   \includegraphics[width=1\linewidth]{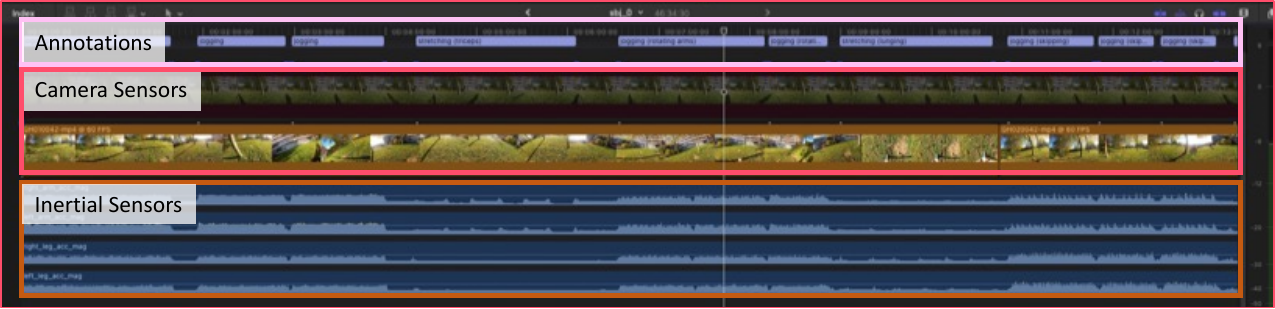}
\end{center}
   \caption{Snapshot along with descriptions of the annotation process using Final Cut Pro. Importing the converted video and inertial data (as `.wav'-files) allowed for an easy validation of the synchronization process. Labels were added via subtitles, exported as `.srt'-files and converted such that they can be appended to the respective `.csv'-files.}
\label{fig:finalcut}
\Description{Snapshot of the tool, i.e. Final Cut, used for annotating the WEAR dataset. The figure shows an overview of the User Interface consisting of the two video streams (3rd person and ego vision), the four magnitude streams (converted as audio files) and the activity annotations, i.e. subtitles.}
\end{figure}

\section{Benchmarks and Baseline Results}
\label{sec:experiments}

Though the WEAR dataset provides the possibility for a multitude of HAR use cases, this paper focuses on introducing one sample application scenario per data modality, namely: (1) inertial-based wearable activity recognition, (2) vision-based TAL, as well as, (3) a combined approach using both data modalities as input simultaneously. We chose to use said application scenarios because of their similarities with each other as they both aim to detect a set of activities in an untrimmed sequence of data. Nevertheless, other HAR-specific (e.g. action anticipation and classification) and non-HAR application scenarios (e.g. hand detection, pose estimation or simultaneous localization and mapping (SLAM)) are applicable. 

During each experiment we employ a Leave-One-Subject-Out (LOSO) cross-validation which has each participant being used as the validation set exactly one time while all other participants are used as trainig data. In order to minimize the risk of performance differences between experiments being the result of statistical variance, all experiments mentioned in this publication were repeated three times employing three different random seeds (1, 2 and 3). The final validation results of an experiment are then reported as the average across the three repeated runs. As mentioned in Chapter~\ref{subsec:datacollection} a separate test dataset was collected which consists of 6 additional participants performing the same workout as all the other participants. The test dataset further includes additional sessions of two participants which volunteered to be recorded a second time. We use the test data to get an unbiased assessment of our chosen set of hyperparameters and applied postprocessing, which are determined during the single-modality and multimodal validation experiments.

Note that all models part of our benchmark analysis were applied without pretraining on existing benchmark datasets. Nevertheless, feature embeddings to train the vision models were extracted using a two-stream I3D feature extractor \cite{carreiraQuoVadisAction2017} which was pre-trained on Kinetics-400 \cite{kayKineticsHumanAction2017}. With the standard error of evaluation metrics amongst runs being at maximum 2.5\% and the majority of runs being below 1\%, we only report average evaluation metrics in this paper. All mentioned experiments were conducted on a single NVIDIA Tesla V100 GPU and lasted no longer than 24 hours. Though sharing inherent similarities, vision-based action localization algorithms predict a collection of activity segments defined by a start and end time, while, contrarily, inertial-based HAR systems provide labels based on the pre-defined windowed segmentation. Given their difference in prediction output, different evaluation metrics are applied, with mean average precision (mAP) being most prominent metric in vision-based TAL and F1-score being the most prominent metric in inertial-based activity recognition. Therefore, to guarantee comparability amongst application scenarios and architectures, predictions of each algorithm are converted such that both vision- and inertial-based evaluation metrics can be calculated. More specifically, our reported benchmark evaluation metrics are (1) a record-based calculated recall, precision and F1-score, and (2) segment-based mean average precision (mAP) at different temporal intersection over union (tIoU) thresholds, commonly used to evaluate TAL datasets. To ensure readability we only provide  a selection of visualized results. For a complete collection visualizations of all mentioned experiments, please see Chapter C.7 within the supplementary material.

\subsection{Single-Stage Temporal Action Localization for Inertial Data}

Though originally intended to be applied to video data, our work demonstrates that vision-based TAL models, such as the TriDet \cite{shiTriDetTemporalAction2023} and ActionFormer \cite{zhangActionFormerLocalizingMoments2022} models, prove to be applicable to inertial data as well as a means to fuse both modalities. Both TAL architectures take as input a collection of clip-wise feature embeddings, which are obtained by applying a sliding window approach on top of an input video. Using both classification and regression losses, the ActionFormer and TriDet model then try to localize activity segments, defined by an activity label, start, and end time, within a complete video. To obtain discriminative feature embeddings of each sliding video clip, the TAL community has resorted to using feature extraction methods such as a two-stream I3D feature extractor \cite{carreiraQuoVadisAction2017}. The feature extractors, usually pretrained on a larger vision corpus like the Kinetics-400 dataset \cite{kayKineticsHumanAction2017}, summarize the raw visual data into a one-dimensional feature embedding.

As both TAL and inertial-based architectures rely on a sliding window approach to preprocess data, our paper suggests a simple yet effective preprocessing technique such that raw inertial data can be used to train vision-based TAL models. Illustrated in Figure~\ref{fig:preprocessing}, we propose vectorizing the two-dimensional sliding window data commonly found in inertial-based architectures, creating one-dimensional raw inertial feature embeddings per sliding window that can be used to train TAL models such as ActionFormer and TriDet. Specifically, starting with the windowed inertial data of dimensions $[\textit{no. windows}, \textit{window length}, \textit{no. sensor axes}]$, we concatenate the individual sensor axes of each window, resulting in a one-dimensional feature vector of size $[\textit{window length} \times \textit{no. sensor axes}]$. In the case of the WEAR dataset, which provides 12 individual sensor axes per participant, we obtain (depending on the window length) feature vectors of size 300 (0.5 seconds), 600 (1 second), and 1200 (2 seconds) per video clip, i.e., sliding window. Even though our concatenation approach results in varying input dimensions, this change does not come at increased computational costs. More specifically, while the number of learnable parameters marginally increases (not more than 10\%) with an increased input dimension, unlike other approaches, no additional embedding needs to be extracted from the inertial data, and raw data streams can be directly used. Furthermore, as demonstrated in the multimodal experiments in this paper, we show that by concatenating one-dimensional inertial data and I3D feature embeddings, the TAL architectures are capable of successfully fusing multi-modal information, significantly outperforming single-modality approaches.

By using our approach, the TAL architectures process the inertial data in a different format compared to, for example, the shallow DeepConvLSTM. However, their performed feature extraction is comparable. Specifically, both architectures shift convolutional filters separately across each sensor axis, with the only difference being that the TAL models do not pad the start and end of each sensor axis, allowing for a slight overlap amongst axes at their point of concatenation. Nevertheless, we argue that our approach draws inspiration from visual feature embeddings such as the I3D features, which have proven effective in training architectures, yet, by design, will also cause an overlap when transitioning between flow and RGB features.

\begin{figure}[!ht]
    \begin{center}
   \includegraphics[width=0.8\linewidth]{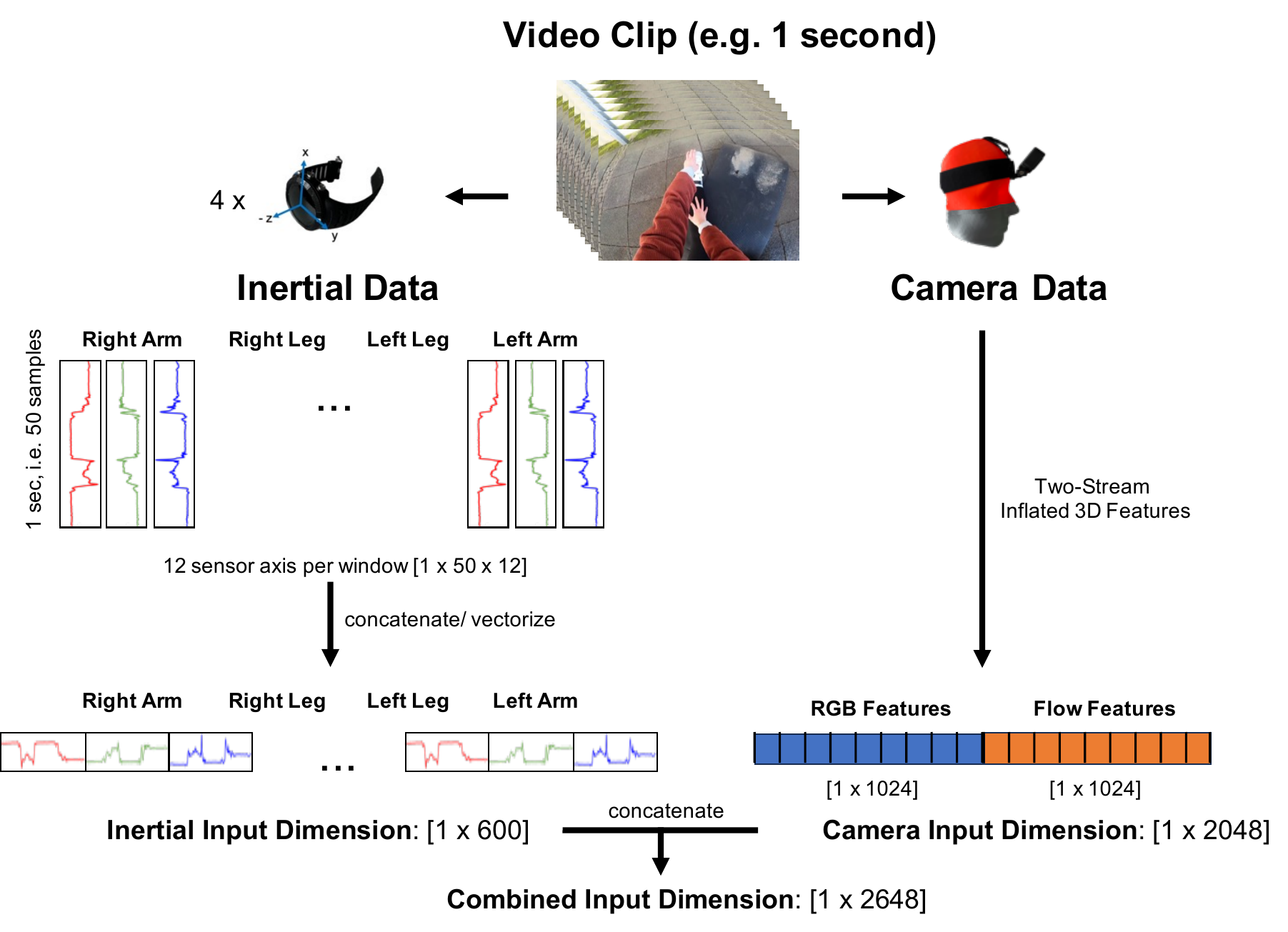}
    \end{center}
   \caption{Visualization of the applied preprocessing on inertial and camera data in order to make to create a feature embedding which can be used to train the TriDet and ActionFormer network. }
\label{fig:preprocessing}
\Description{Visualization of the applied preprocessing of the raw video and inertial to create feature embeddings which can be used as input for the ActionFormer and TriDet model.}
\end{figure}

\subsection{Single-modality Experiments}
\label{subsec:single}

Similar to \citet{zhangActionFormerLocalizingMoments2022} and \citet{shiTriDetTemporalAction2023}, we opted to train the vision-based benchmark models using two-stream I3D feature embeddings with three different clip lengths (0.5, 1, and 2 seconds), employing a 50\% overlap between clips. Besides extending the number of epochs to 100, we adopted the same training strategy that yielded the best results on the EPIC-Kitchens dataset \cite{damenRescalingEgocentricVision2022}, as reported by both architectures. In contrast to inertial-based approaches, TAL models are not trained to predict an explicitly modeled NULL-class. With both models set to predict up to 2000 action segments per video, each timestamp ended up being classified by an action segment, resulting in a prediction performance for the NULL-class close to or at 0\% accuracy. To address this, we eliminated low-scoring segments by increasing the scoring threshold of both models to 0.1, significantly enhancing the accuracy of the NULL-class while only marginally affecting prediction performance of all other activity classes.

As our inertial-based benchmark algorithms of choice, we employed the shallow DeepConvLSTM proposed by \citet{bockImprovingDeepLearning2021} and the Attend-and-Discriminate model proposed by \citet{abedinAttendDiscriminateStateoftheart2021}. Throughout all experiments, we followed the training strategy suggested by \citet{bockImprovingDeepLearning2021}, which has shown to yield reliable results on various inertial-based HAR datasets, only extending the number of epochs to match those of the vision-based experiments. To compensate for longer training times, we applied a step-wise learning rate schedule. Additionally, incorporating architecture changes recommended by \citet{bockImprovingDeepLearning2021}, we modified the Attend-and-Discriminate model to utilize a one-layered instead of a two-layered recurrent module and adjusted the convolutional kernel size based on the sliding window and sampling rate of the WEAR dataset. Given that inertial-based architectures provide predictions on a per-window basis, frequent intermediate, short-lasting activity switches occur along the time axis, resulting in small coherent segments and ultimately lower mAP scores compared to the vision-based models presented in this paper. To mitigate these switches, predictions made by the inertial-based architectures were smoothed using a majority-vote filter of 10 seconds. For detailed ablation results on the design choices and chosen hyperparameters see Section C of the supplementary material.

Examining the results presented in Table~\ref{tab:results}, one can observe that for the TAL models, a clip length of 1 second yielded the best predictive performance, while for the inertial-based architectures, a larger window size of 2 seconds produced higher classification results. Analyzing per-class results, it is evident that when using only visual data as input, the models struggle to differentiate between different running styles, activities outside the field of view of the participant (e.g., triceps stretches), as well as normal and complex sit-ups. In contrast, since inertial sensors are not affected by occlusions and accelerometer data provides an encoded representation of limb orientation due to the Earth's gravitational force, algorithms trained on inertial data do not exhibit these weaknesses and are particularly reliable during activities where limb orientation is the main discriminator (e.g., stretches). However, as indicated by a larger NULL-class accuracy compared to the inertial-based models, vision-based approaches are better at differentiating activities from breaks, which is a desirable trait of HAR systems given the often substantial size of the NULL-class \cite{bullingTutorialHumanActivity2014}. 

Interestingly, using the same hyperparameters as those used during vision-based experiments, a plain ActionFormer and TriDet network not only produce competitive classification results based on inertial input data but also show less confusion among activity classes compared to inertial-based architectures. Comparing per-class results between the TAL models and classic inertial-based architectures, it is evident that the confusion in the TAL models mostly lies within the activity categories themselves (jogging, stretching, and strength), whereas inertial-based models exhibit a higher degree of overall confusion between the activities and the NULL-class. Given that inertial-based models like the DeepConvLSTM are trained to classify sliding windows individually, we assume that this overall larger confusion and intermediate activity switches are caused by the inertial models' inability to leverage context information beyond the window they are tasked to classify. Consequently, due to the intermediate activity switches, calculated mAP scores of the inertial architectures are significantly lower than those of the camera-based approaches. Nonetheless, inertial-based models are on average able to predict all workout activities more consistently and produce the highest classification metrics across all experiments.

\subsection{Multimodal (Inertial and Egocentric Video) Experiments}
\label{subsec:multi}

As demonstrated in previous chapters, both the vision and inertial modalities present various strengths and weaknesses. Therefore, in an effort to leverage the strengths of both modalities, we evaluated a combined, multimodal training approach using the TriDet and ActionFormer model. To fuse the two-stream I3D feature embeddings with the inertial data early on, we vectorized the inertial data, as described in the previous section,  such that each sliding window is represented by a one-dimensional inertial embedding vector. We then concatenated said vector with the I3D feature embedding of the corresponding video clip, resulting in an early fused representation of the a sliding window consisting of both visual and inertial information. Through simple concatenation of both modalities, both architectures achieved the highest average mAP and close-to-best F1-scores across all experiments (see Table~\ref{tab:results}).
Comparing confusion matrices of all three approaches (see Figure~\ref{fig:confmats}) reveals that both vision models, applied in a plain fashion, are able to successfully combine inertial and vision data and leveraging the previously mentioned strengths of each modality. To assess how our early-fusion approach compares to voting-based late-fusion approaches such as proposed by \citet{ijazMultimodalTransformerNursing2022}, we implemented an \emph{Oracle}-based late fusion, which creates perfectly late fused predictions of different models. The predictions are merged by comparing each of them with the ground truth data and only keeping, if predicted by one of the networks, the correct prediction. Interestingly, the first \emph{Oracle}-late-fusion \emph{O-LF(I, C)}, which late fuses predictions of the best inertial and best vision model, produces lower mAP scores than that of the best TAL model being trained on both modalities simultaneously. Furthermore, late-fusing the best inertial, vision and early-fusion approach (\emph{O(I, C, I + C)}), increases classification and mAP scores of \emph{O(I, C)} by as much as 10\%, suggesting the early-fusion-based approach is capable of learning to differentiate activities both single-modality models failed to classify correctly (see e.g. misclassified stretching exercises in Figure~\ref{fig:oracle_viz}). Nevertheless, classification results of the \emph{Oracle}-based late fusion significantly outperform both single- and combined-modality approaches, indicating that the data set is far from being saturated.

\begin{table}
  \centering
  \small
  \caption{LOSO validation results of human activity recognition approaches based on body-worn IMU (Inertial), vision (Camera) and combined (Inertial + Camera) features for different clip lengths (CL) on the first 18 participants of the WEAR dataset evaluated in terms of precision (P), recall (R), F1-score and mean average precision (mAP) for different temporal intersection over union (tIoU) thresholds. The results underline the complementarity of the inertial and camera modalities. \emph{O-LF()} corresponds to the \emph{Oracle}-based late fusion, which creates perfectly late fused predictions of different models.} Best results per modality are in \textbf{bold}.
  \label{tab:results}
  \begin{tabular}{@{}llcccccccccc@{}}
    & Model & CL & P & R & F1 & \multicolumn{6}{c}{mAP} \\ \cmidrule{7-12} 
    & & & & & & 0.3 & 0.4 & 0.5 & 0.6 & 0.7 & Avg \\
    \toprule
    \multirow{12}{*}{\rotatebox[origin=c]{90}{Inertial}}
    & Shallow D.              & 0.5s                & 74.81 & 74.93 & 72.26 & 60.03 & 57.91 & 55.54 & 54.19 & 52.31 & 55.99 \\
    & A-and-D                 & 0.5s                & 76.54 & 73.04 & 72.10 & 57.51 & 54.59 & 52.22 & 49.47 & 46.94 & 52.15 \\
    & ActionFormer            & 0.5s                & 63.09 & 78.17 & 66.80 & 81.12 & 77.47 & 69.08 & 53.98 & 38.56 & 64.04 \\
    & TriDet                  & 0.5s                & 68.58 & 78.30 & 70.41 & 79.66 & 76.07 & 69.78 & 60.57 & 49.95 & 67.21 \\
    & Shallow D.              & 1s                  & 77.71 & 77.41 & 75.44 & 63.37 & 61.35 & 58.91 & 57.02 & 55.05 & 59.14 \\
    & A-and-D                 & 1s                  & 79.97 & 75.92 & 74.67 & 61.72 & 58.77 & 56.29 & 53.83 & 51.59 & 56.44 \\
    & ActionFormer            & 1s                  & 65.88 & 78.44 & 68.40 & \textbf{81.91} & 79.75 & 76.99 & 72.09 & 64.33 & 75.01 \\
    & TriDet                  & 1s                  & 68.18 & \textbf{78.58} & 70.36 & 81.89 & \textbf{80.46} & \textbf{78.47} &  \textbf{74.59} & \textbf{68.91} & \textbf{76.86} \\
    & Shallow D.              & 2s                  & 78.00 & 77.19 & 75.43 & 65.14 & 63.58 & 60.98 & 59.15 & 57.39 & 61.25 \\
    & A-and-D                 & 2s                  & \textbf{80.96} & 78.42 & \textbf{77.10} & 63.52 & 61.57 & 58.82 & 56.88 & 54.41 & 59.04 \\
    & ActionFormer            & 2s                  & 61.56 & 76.63 & 64.57 & 80.14 & 77.24 & 74.41 & 70.24 & 63.18 & 73.04 \\
    & TriDet                  & 2s                  & 63.14 & 75.64 & 65.94 & 78.69 & 77.02 & 73.89 & 70.68 & 65.31 & 73.12 \\
    \midrule
    \multirow{6}{*}{\rotatebox[origin=c]{90}{Camera}}
    & ActionFormer            & 0.5s                & 60.86 & 72.98 & 62.54 & 81.07 & 78.48 & 72.13 & 57.34 & 41.27 & 66.06 \\
    & TriDet                  & 0.5s                & 64.69 & 72.89 & 64.84 & 80.46 & 76.95 & 72.62 & 64.46 & 55.52 & 70.00 \\
    & ActionFormer            & 1s                  & 65.82 & 75.34 & 66.40 & \textbf{85.19} & 83.31 & 80.94 & 77.22 & 69.96 & 79.32 \\
    & TriDet                  & 1s                  & \textbf{67.42} & 75.21 & \textbf{66.88} & 84.88 & \textbf{83.49} & \textbf{82.12} & \textbf{79.75} & \textbf{76.24} & \textbf{81.30} \\
    & ActionFormer            & 2s                  & 63.40 & \textbf{76.25} & 65.39 & 84.93 & 82.95 & 80.58 & 77.55 & 72.07 & 79.62 \\
    & TriDet                  & 2s                  & 65.53 & 75.36 & 65.91 & 82.94 & 82.00 & 80.22 & 78.13 & 75.20 & 79.70 \\
    \midrule
    \multirow{12}{*}{\rotatebox[origin=c]{90}{Inertial + Camera}}
    & ActionFormer            & 0.5s                & 71.45 & \textbf{83.77} & 74.51 & 86.04 & 83.96 & 77.80 & 64.41 & 43.76 & 71.19 \\
    & TriDet                  & 0.5s                & \textbf{75.74} & 82.79 & \textbf{76.83} & 84.45 & 82.07 & 77.12 & 68.81 & 57.16 & 73.92 \\
    & ActionFormer            & 1s                  & 72.87 & 83.24 & 75.26 & \textbf{86.82} & 85.27 & 82.46 & 78.26 & 72.33 & 81.03 \\
    & TriDet                  & 1s                  & 73.39 & 82.00 & 75.09 & 86.42 & \textbf{85.48} & \textbf{83.35} & \textbf{80.39} & \textbf{75.64} & \textbf{82.26} \\
    & ActionFormer            & 2s                  & 65.86 & 80.17 & 68.82 & 84.18 & 81.66 & 78.17 & 74.59 & 68.74 & 77.47 \\
    & TriDet                  & 2s                  & 69.19 & 81.09 & 71.98 & 84.66 & 82.79 & 80.66 & 78.11 & 74.15 & 80.07 \\ \cmidrule{2-12} 
    & \emph{O-LF(I, C)}             & \textit{0.5s}   & \textit{90.23} & \textit{91.52} & \textit{89.60} & \textit{74.79} & \textit{73.73} & \textit{72.44} & \textit{69.94} & \textit{68.53} & \textit{71.88} \\
    & \emph{O-LF(I, C)}             & \textit{1s}     & \textit{91.26} & \textit{92.72} & \textit{90.68} & \textit{75.37} & \textit{74.42} & \textit{73.64} & \textit{72.74} & \textit{71.41} & \textit{73.52} \\
    & \emph{O-LF(I, C)}             & \textit{2s}     & \textit{91.26} & \textit{92.94} & \textit{90.81} & \textit{73.46} & \textit{72.37} & \textit{71.70} & \textit{70.50} & \textit{69.60} & \textit{71.53} \\ \cmidrule{2-12} 
    & \emph{O-LF(I, C, I + C)}      & \textit{0.5s}   & \textit{93.29} & \textit{94.37} & \textit{93.10} & \textit{84.41} & \textit{84.17} & \textit{83.37} & \textit{81.61} & \textit{80.40} & \textit{82.79} \\
    & \emph{O-LF(I, C, I + C)}      & \textit{1s}     & \textit{93.99} & \textit{94.89} & \textit{93.38} & \textit{84.14} & \textit{83.81} & \textit{83.34} & \textit{82.54} & \textit{81.81} & \textit{83.13} \\
    & \emph{O-LF(I, C, I + C)}      & \textit{2s}     & \textit{93.69} & \textit{95.13} & \textit{93.38} & \textit{81.79} & \textit{81.37} & \textit{81.02} & \textit{79.90} & \textit{78.94} & \textit{80.60} \\
    \bottomrule
  \end{tabular}
\end{table}

\begin{figure}
\begin{center}
   \includegraphics[width=1.\linewidth]{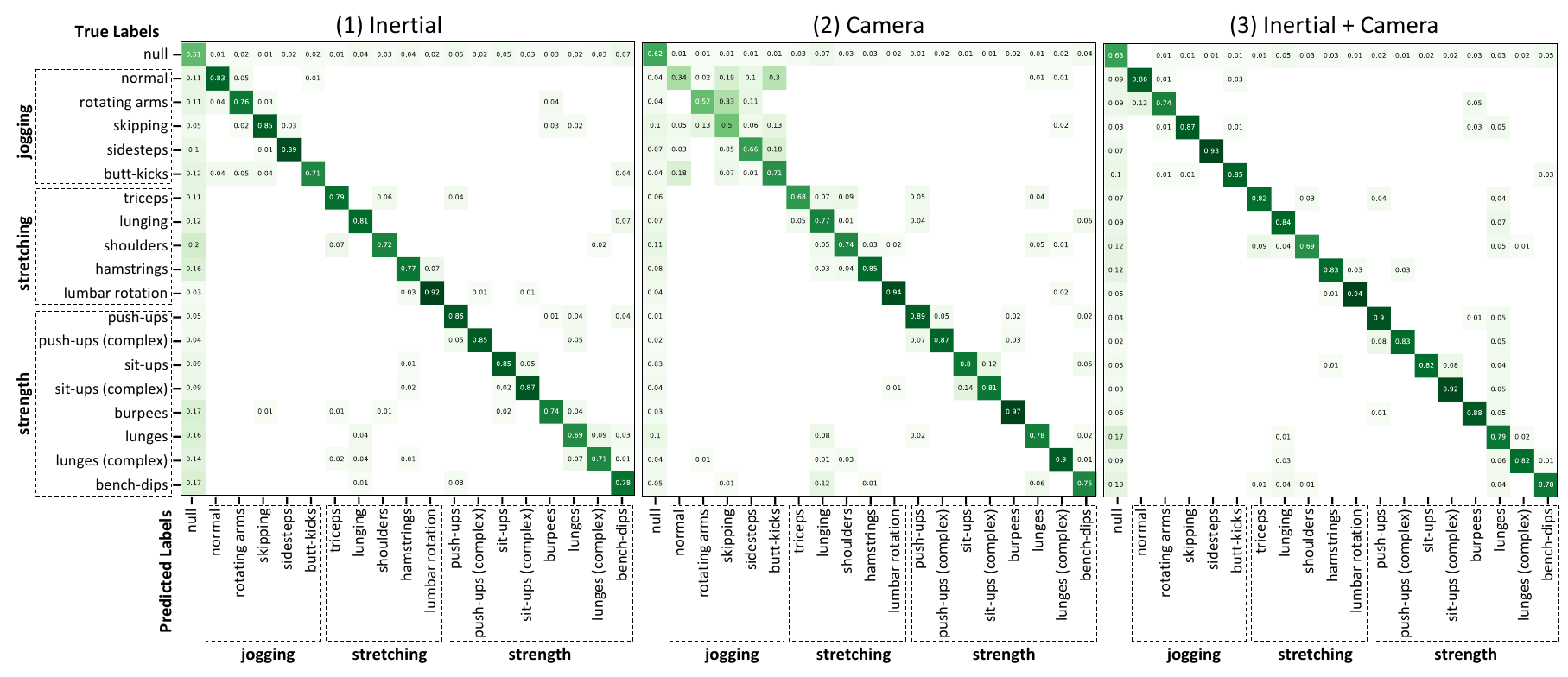}
\end{center}
   \caption{Confusion matrices of the TriDet model \citep{shiTriDetTemporalAction2023} being applied using inertial, vision (camera) and both combined (inertial + camera) with a one second sliding window and 50\% overlap. Compared to inertial-based architectures \citep{bockImprovingDeepLearning2021, abedinAttendDiscriminateStateoftheart2021} overall confusion (except for the NULL-class) is decreased. After combination strengths of each architecture are leveraged with e.g. jogging activities not getting confused anymore and overall confusion with the NULL-class decreases. Note that confusions which are 0 are omitted.}
\label{fig:confmats}
 \Description{Visualization of the confusion matrices of the TriDet model being applied on the inertial, vision and combined data. The confusion matrices show how the confusion, i.e. misclassification of the algorithms, are distributed amongst the activity classes of the WEAR dataset. Each cell within the matrix contains the percentage of confusion (unless the confusion is 0).}
\end{figure} 

\begin{figure}
\begin{center}
   \includegraphics[width=0.9\linewidth]{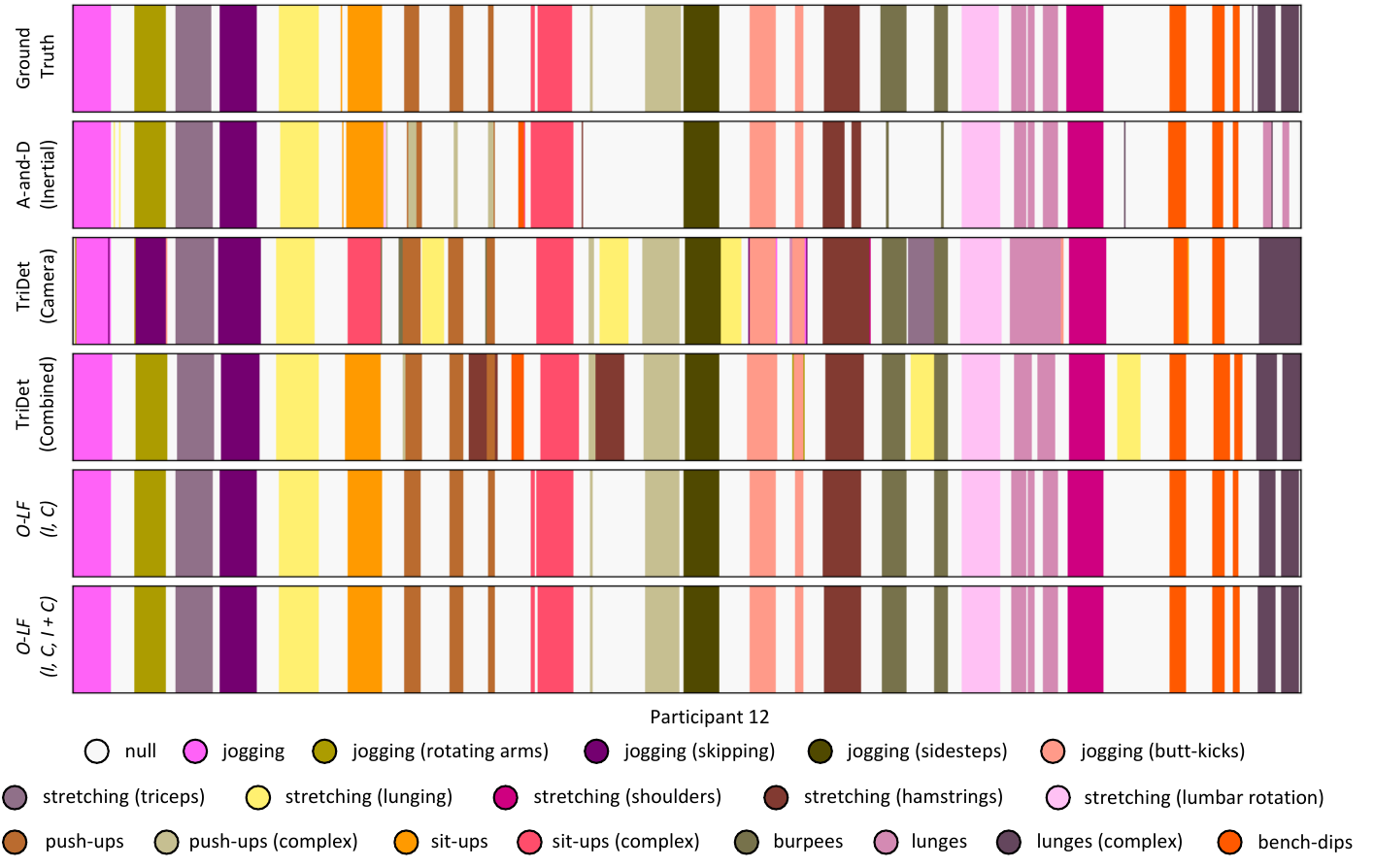}
\end{center}
   \caption{Color-coded comparison of the ground truth data of a sample participant with the best inertial-based (A-and-D), camera-based (TriDet) and fusion-based model (TriDet) along with an oracle combination of the best fusion-based model (\emph{O-LF(I, C)}) as well as an oracle combination the best camera, inertial and fusion-based-model (\emph{O-LF(I, C, I + C)}) using a sliding window approach of 1.0 seconds with a 50\% overlap. The visualisation underlines the similarities amongst the predictive streams of \emph{O-LF(I, C)} and the fusion-approach as well advantages of learning from both modalities simultaneously.}
\label{fig:oracle_viz}
\Description{Color-coded visualization of tested deep learning algorithms compared with the different oracle combinations of them. The figure shows colored timelines representing predictions made by the algorithms for a sample participant compared with the actual color-coded activity timeline.}
\end{figure}

\subsection{Test Set Results}

To verify chosen hyperparameters and suitability of the applied postprocessing as determined in the LOSO cross-validation, we collected an additional test set of six additional participants. The participants constitute of four previously unseen participants as well as two participants, already present in the training data and which volunteered to be recorded a second time. Unlike recordings of the training data, the test data was recorded during spring and summer. Egocentric video data of the four new participants was recorded using a GoPro Hero 11 as opposed to a Hero 8 and two new participants were recorded at a previously unseen location. Figure \ref{fig:test_results} summarizes results of the benchmark algorithms applied on the test dataset using a 1 second sliding window with a 50\% overlap. Training was performed identical to that described in previous chapters with the input data being the data of the 18 participants used for LOSO validation. Each experiment was repeated three times with different random seeds (i.e., 1, 2 or 3). One can see that observed trends are similar to those seen during LOSO cross-validation. Nevertheless, one can observe a increase in average mAP and classification metrics for both the inertial- and vision-based models, which does also translate to mAP and class increasing when combining camera and inertial data.

\begin{figure}
\begin{center}
   \includegraphics[width=0.7\linewidth]{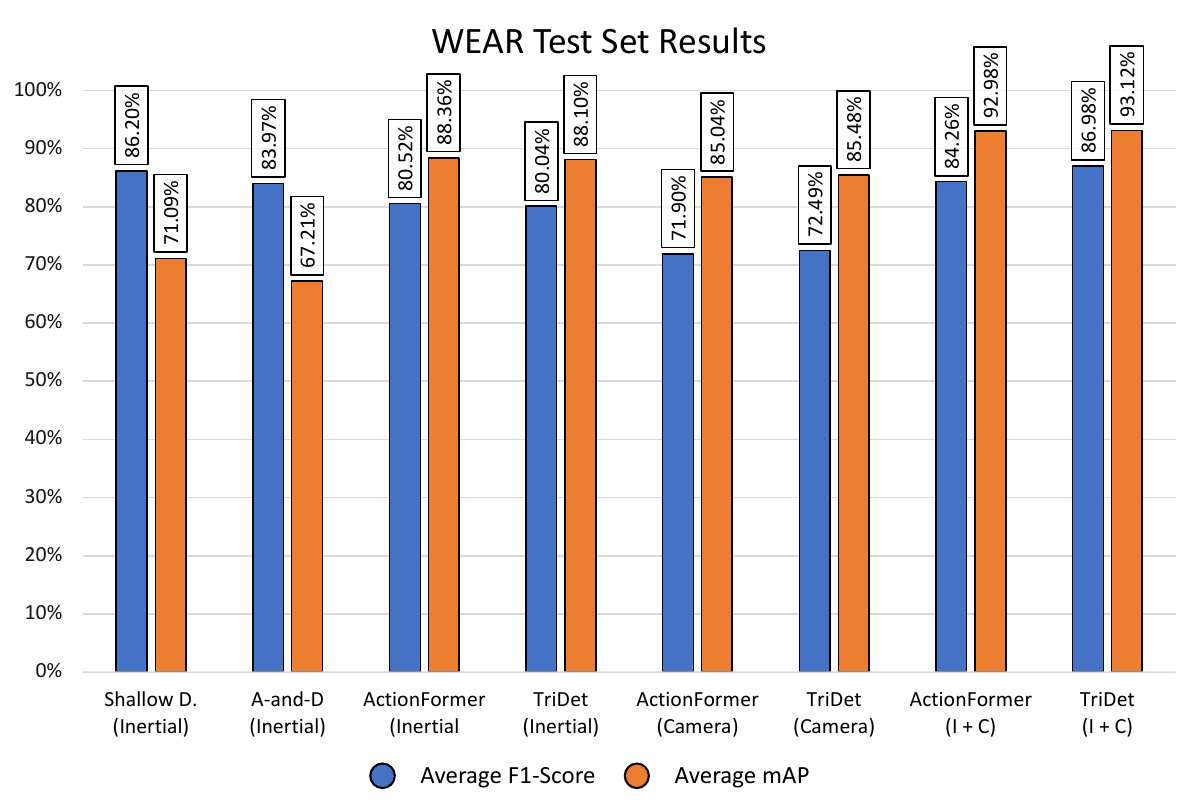}
\end{center}
   \caption{Average F1-score and mAP on the WEAR test set. The test features 4 unseen participants as well 2 reoccuring ones, a unseen location, different weather conditions and a new camera sensor. One can see that observed trends are similar to those seen during LOSO cross-validation.}
\label{fig:test_results}
\Description{Bar-plot visualization of the test results for the different benchmarking algorithms. We provide both a bar plot for F1-score as well as average mAP. One can see similar trends as observed during the LOSO cross-validation}
\end{figure}

\section{Ablation Experiments}

\subsection{Selection of Inertial Sensors}

Although commercial fitness products worn on the ankle or foot (e.g., anklets, sock-embedded devices, or shoe-embedded units) have gained popularity in recent years, wrist-worn sensors like smartwatches remain the most popular body position for inertial sensors in the context of wearable activity recognition. The following experiments (see Figure~\ref{fig:sensor_selection_results}) investigate how reported LOSO results in Chapter~\ref{subsec:single} and \ref{subsec:multi} are influenced by using only a subset of the inertial sensors, specifically by using only (1) acceleration recorded from the right wrist and (2) acceleration recorded from both the right wrist and right ankle. Results show that using only acceleration data obtained from the right wrist significantly decreases predictive performance across all algorithms and metrics. Moreover, the value of additionally measuring acceleration at the ankles of participants is clearly underlined, as results again significantly increase, mostly on par compared to using all four inertial sensor locations. Interestingly, unlike the inertial-based architectures, results of the vision-based TAL models improve when excluding data captured by the left wrist and left ankle inertial sensors, which could be due to the dataset being biased towards right-handed participants (see Table 3 in the supplementary material) and dominant hand movement being overall more consistent. 

\begin{figure}
\begin{center}
   \includegraphics[width=1.\linewidth]{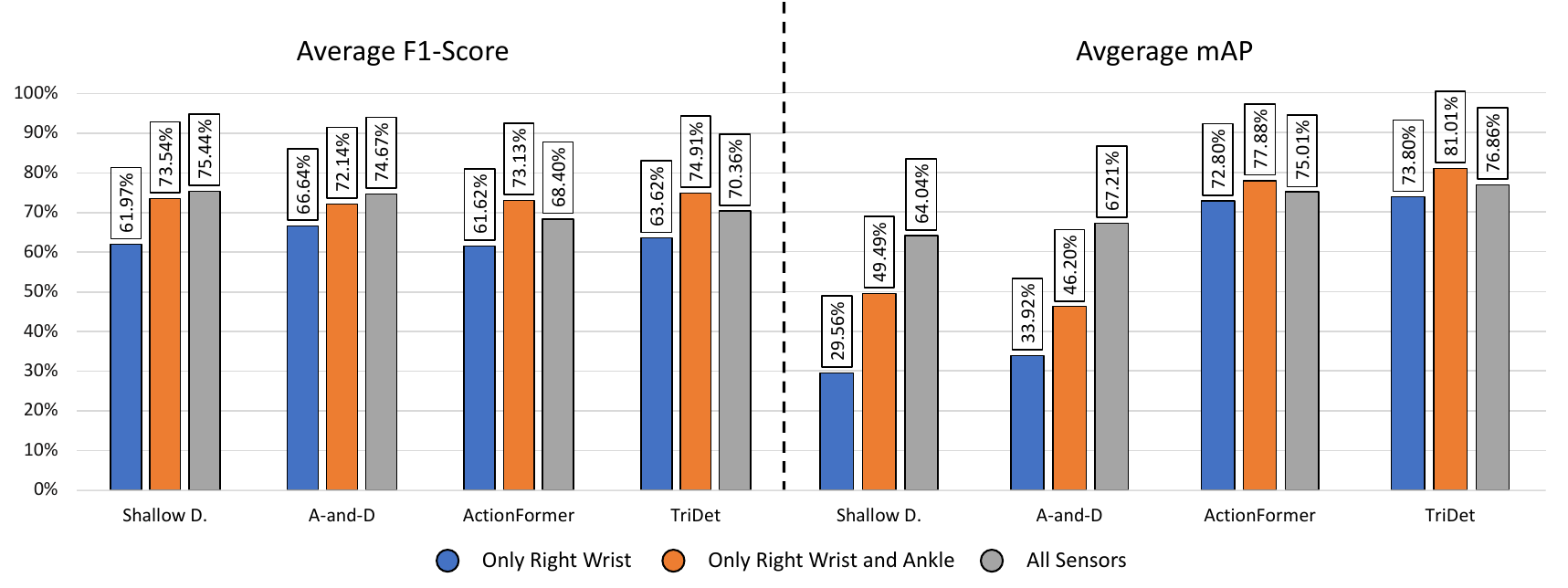}
\end{center}
   \caption{Results using only inertial sensors placed on the right wrist (RW) and right wrist + ankle (RW + RA) compared with using all sensors for a clip length of 1 second on our WEAR dataset evaluated in terms of F1-score and mean average precision (mAP) averaged across different temporal intersection over union (tIoU) thresholds (0.3:0.7:0.1). Using only wrist-worn data can see a clear overall decrease across all evaluation metrics.}
    \label{fig:sensor_selection_results}
 \Description{Visualization of the architectures being applied different subsets of inertial sensors. The figure shows two bar plots for F1-score and mAP respectively.}
\end{figure}

\subsection{Second execution of workout sessions}
\label{sec:appinertialsensors}
As mentioned in Chapter~\ref{subsec:datacollection}, as part of the test dataset provided along the WEAR dataset, we recorded all activities of two participants (sbj\_0 and sbj\_14) a second time in August. Both participants recording conditions significantly differed from their first recording, with temperatures being around 25 degrees Celsius with overall more sunny weather conditions. Further, as not all participants knew all activities beforehand (see Table 10 in the supplementary material), recording the same participants a second time would allow to analyse how a certain degree of familiarity with the recording setup can be seen in altered movements (e.g., via a smoother execution of activities) as well as participant-specific finetuning affects the overall recognition performance. Figure~\ref{fig:rerecording} compares validation results obtained on the first recording of sbj\_0 and sbj\_14 with their second execution of the workout plan in August. Unlike our prior experiments, each algorithm is trained using the validation data as reported in Chapter~\ref{subsec:single} and \ref{subsec:multi} except the participant which is to be analysed. All results are postprocessed as reported Chapter~\ref{subsec:single} and \ref{subsec:multi}. While, one can see improved results regarding sbj\_0, which only knew five of the workout activites prior to participating in the study, this trend does not apply to sbj\_14. More specifcially, improvements and decline rates between the two recordings lie within the expected standard deviation across participants (between 15\% to 20\%). Though being a small sample size of only two participants, the results suggest that in order to guarantee a reliable detection of activities, each participant would need to be recorded multiple times under different conditions. Nevertheless, in order to come up with reliable conclusions, future extensions of the WEAR dataset would need focus on re-recording more participants multiple times under varying conditions.

\begin{figure}
\begin{center}
   \includegraphics[width=1.\linewidth]{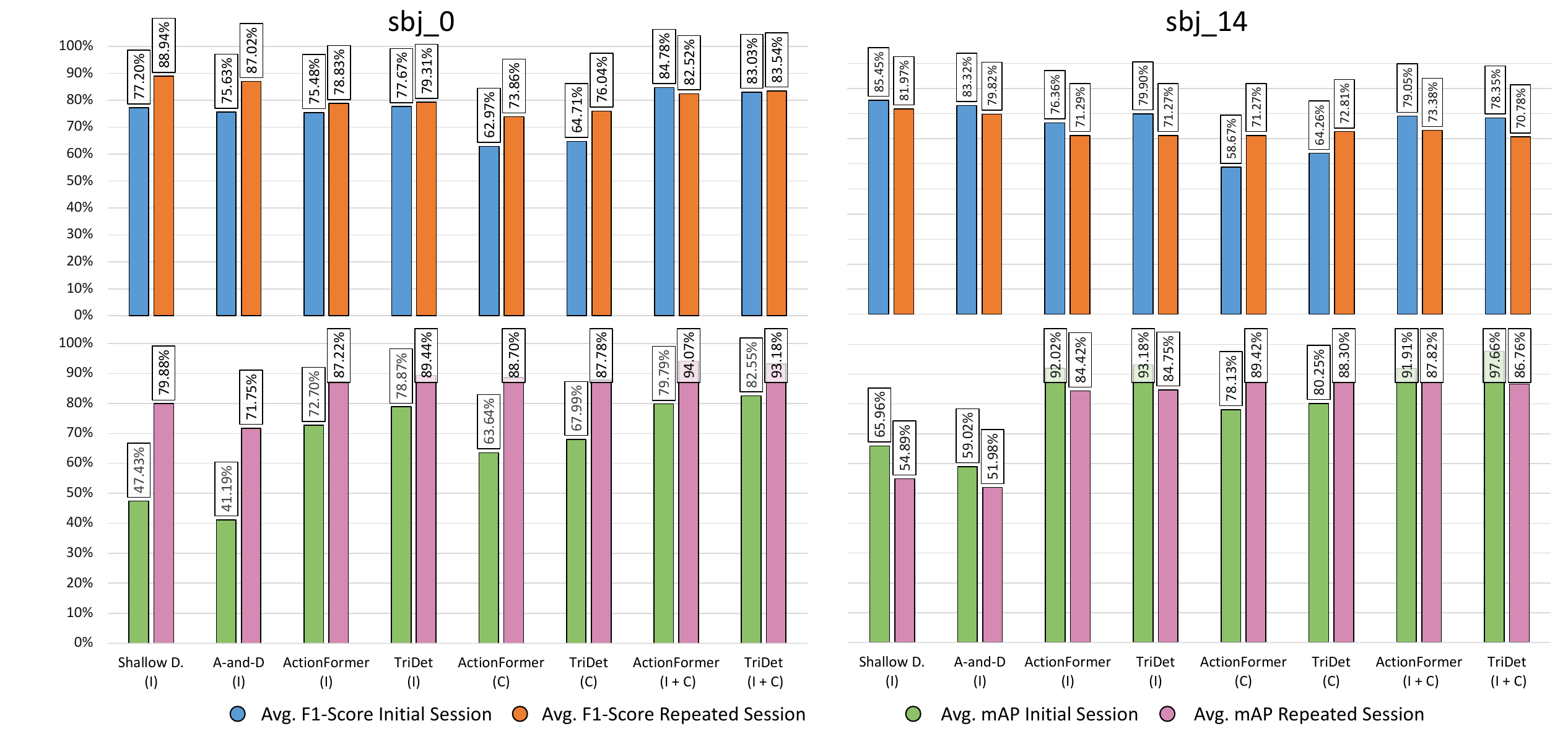}
\end{center}
   \caption{Comparison of obtained results of repeated sessions for participants sbj\_0 and sbj\_14 for a clip length of 1 second on our WEAR dataset evaluated in terms of F1-score and mean average precision (mAP). The two participants were invited to perform the recording plan a second time. While one can see that improved results regarding sbj\_0, suggesting potential learning effects of the correct execution of activities, this trend does not apply to sbj\_14. Note that weather conditions (temperature and sunlight) significantly differ amongst the recordings -- winter (first recording) compared to summer (2nd recording). These figures are, as in the earlier results, averaged across 3 runs using 3 different random seeds. Unlike our prior experiments, each algorithm is trained using the training data of all but the validation participants' recordings, ensuring the validation participants (sbj\_0 and sbj\_14) remain unseen during the training of each algorithm. All results are postprocessed as reported in the main paper.}
    \label{fig:rerecording}
 \Description{Visualization of the architectures being the first compared to the second recordings of sbj\_0 and sbj\_14. The figure shows bar two plots for F1-score and mAP respectively.}
\end{figure}

\section{Limitations}
Our dataset contributes a benchmark for human activity recognition classifiers, for the two leading wearable modalities of egocentric video and inertial data, using in particular a high variety of fitness exercises and outdoor scenes. With the current selection of participants, the WEAR dataset is biased towards young, healthy people. Given the ease of reproducibility, future extensions of the WEAR dataset could focus on featuring participants (1) of an older and/ or younger age, (2) with known physical impairments and (3) sessions recorded at new locations (outside of Germany) and at different times of the year (e.g. summer). As supplementary experiments already indicate, recording the same participants a second time would allow to analyse how a certain degree of familiarity with the recording setup can be seen in altered movements (e.g., via a smoother execution of activities) as well as give an intuition about robustness of learned approaches. Besides extending the amount of data recorded, further recordings could also involve other sensors such as higher-end commercial smartwatches to enable the study of increased sampling rates, the variability of the capturing devices, and the inclusion of additional modalities such as 3D gyroscopes, 3D magnetometers, or photoplethysmography (PPG) to obtain fitness-relevant information such as heart rate), as well as additional wearables, such as earables. Furthermore, as participants remain at the same location during their workout, including recordings of the same participants at different locations would allow to assess classifiers based on their ability to learn general cues to predict the workout activities rather than overfitting on location-specific ones. Lastly, as a camera observing the participant is needed for annotation of the collected data, a future extension of the WEAR dataset could include releasing the 3rd-person video data of the workouts along with the egocentric data, enabling the dataset to be also used e.g. for pose estimation.

\section{Discussion \& Conclusion}
\label{sec:conclusion}
In this paper, we introduced a benchmark dataset for both inertial- and vision-based Human Actvity Recognition (HAR), to explore the learning of HAR across these modalities. The dataset comprises data from 22 participants performing each 18 different sports activities with the two common types of wearable sensors delivering inertial (3D acceleration) and camera (egocentric video) data. Unlike previous egocentric datasets, the application-driven sensor setup of WEAR provides a challenging prediction scenario across both modalities marked by changing outdoor environments along with a small information overlap between the inertial and vision data, putting forward the necessity of exploring techniques to combine both modalities.

Benchmark results obtained using each modality separately show that each modality interestingly offers complementary strengths and weaknesses in their prediction performance. In light of the recent successes of single-stage TAL models following the architecture design as proposed by \citet{zhangActionFormerLocalizingMoments2022}, we demonstrate their versatility by applying them in a plain fashion without any pretraining using preextracted I3D features and raw inertial data as input. Using the inherent similarities of approaches of both communities, we suggest a simple, yet effective preprocessing technique of inertial data, which vectorizes sliding window such that they can be used to train TAL models. Vision-based TAL such as the ActionFormer \cite{zhangActionFormerLocalizingMoments2022} have thus far neither been explored in inertial nor in the combination of inertial- and vision-based human activity recognition. Results show that the vision-based models are not only able to produce competitive results using inertial data, but also can function as an architecture to fuse both modalities by means of simple concatenation with vision data. In experiments that combined raw inertial with extracted vision-based feature embeddings, the plain, vision-based TAL models were able to produce the highest average mAP and close-to-best F1-scores. Lastly, to give an intuition about a possible upper bound for future fusion-approaches, we evaluated an oracle-merged late fusion of the best inertial- and vision-based model predictions.

With WEAR, we provide both communities (inertial- and vision-based HAR) a common, challenging benchmark dataset to assess the applicability of combined human activity recognition approaches.

\begin{acks}
We gratefully acknowledge the DFG Project WASEDO (grant number 506589320 ) and the University of Siegen's OMNI cluster.
\end{acks}

\bibliographystyle{ACM-Reference-Format}
\bibliography{main}


\begin{thebibliography}{85}

%%% the Web address.

\ifx \showCODEN    \undefined \def \showCODEN     #1{\unskip}     \fi
\ifx \showDOI      \undefined \def \showDOI       #1{#1}\fi
\ifx \showISBNx    \undefined \def \showISBNx     #1{\unskip}     \fi
\ifx \showISBNxiii \undefined \def \showISBNxiii  #1{\unskip}     \fi
\ifx \showISSN     \undefined \def \showISSN      #1{\unskip}     \fi
\ifx \showLCCN     \undefined \def \showLCCN      #1{\unskip}     \fi
\ifx \shownote     \undefined \def \shownote      #1{#1}          \fi
\ifx \showarticletitle \undefined \def \showarticletitle #1{#1}   \fi
\ifx \showURL      \undefined \def \showURL       {\relax}        \fi
% invisible to TeX
\providecommand\bibfield[2]{#2}
\providecommand\bibinfo[2]{#2}
\providecommand\natexlab[1]{#1}
\providecommand\showeprint[2][]{arXiv:#2}

\bibitem[Abedin et~al\mbox{.}(2021)]%
        {abedinAttendDiscriminateStateoftheart2021}
\bibfield{author}{\bibinfo{person}{Alireza Abedin}, \bibinfo{person}{Mahsa
  Ehsanpour}, \bibinfo{person}{Qinfeng Shi}, \bibinfo{person}{Hamid
  Rezatofighi}, {and} \bibinfo{person}{Damith~C. Ranasinghe}.}
  \bibinfo{year}{2021}\natexlab{}.
\newblock \showarticletitle{Attend and {Discriminate}: {Beyond} the
  {State}-{Of}-{The}-{Art} for {Human} {Activity} {Recognition} {Using}
  {Wearable} {Sensors}}.
\newblock \bibinfo{journal}{\emph{ACM on Interactive, Mobile, Wearable and
  Ubiquitous Technologies}} \bibinfo{volume}{5}, \bibinfo{number}{1}
  (\bibinfo{year}{2021}), \bibinfo{pages}{1--22}.
\newblock
\urldef\tempurl%
\url{https://doi.org/10.1145/3448083}
\showURL{%
\tempurl}


\bibitem[Bai et~al\mbox{.}(2020)]%
        {baiBoundaryContentGraph2020}
\bibfield{author}{\bibinfo{person}{Yueran Bai}, \bibinfo{person}{Yingying
  Wang}, \bibinfo{person}{Yunhai Tong}, \bibinfo{person}{Yang Yang},
  \bibinfo{person}{Qiyue Liu}, {and} \bibinfo{person}{Junhui Liu}.}
  \bibinfo{year}{2020}\natexlab{}.
\newblock \showarticletitle{Boundary {Content} {Graph} {Neural} {Network} for
  {Temporal} {Action} {Proposal} {Generation}}. In
  \bibinfo{booktitle}{\emph{European {Conference} on {Computer} {Vision}}}.
\newblock
\urldef\tempurl%
\url{https://doi.org/10.1007/978-3-030-58604-1_8}
\showURL{%
\tempurl}


\bibitem[Bao and Intille(2004)]%
        {baoActivityRecognitionUserAnnotated2004}
\bibfield{author}{\bibinfo{person}{Ling Bao} {and} \bibinfo{person}{Stephen~S.
  Intille}.} \bibinfo{year}{2004}\natexlab{}.
\newblock \showarticletitle{Activity {Recognition} {From} {User}-{Annotated}
  {Acceleration} {Data}}.
\newblock \bibinfo{journal}{\emph{Pervasive Computing}} (\bibinfo{year}{2004}),
  \bibinfo{pages}{158--175}.
\newblock
\urldef\tempurl%
\url{https://doi.org/10.1007/978-3-540-24646-6_1}
\showURL{%
\tempurl}


\bibitem[Bin~Morshed et~al\mbox{.}(2022)]%
        {binmorshedPersonalizedApproachDeveloping2022}
\bibfield{author}{\bibinfo{person}{Mehrab Bin~Morshed},
  \bibinfo{person}{Harish~Kashyap Haresamudram}, \bibinfo{person}{Dheeraj
  Bandaru}, \bibinfo{person}{Gregory~D. Abowd}, {and} \bibinfo{person}{Thomas
  Ploetz}.} \bibinfo{year}{2022}\natexlab{}.
\newblock \showarticletitle{A {Personalized} {Approach} for {Developing} a
  {Snacking} {Detection} {System} using {Earbuds} in a {Semi}-{Naturalistic}
  {Setting}}. In \bibinfo{booktitle}{\emph{{ACM} {International} {Symposium} on
  {Wearable} {Computers}}}.
\newblock
\urldef\tempurl%
\url{https://doi.org/10.1145/3544794.3558469}
\showURL{%
\tempurl}


\bibitem[Bock et~al\mbox{.}(2021)]%
        {bockImprovingDeepLearning2021}
\bibfield{author}{\bibinfo{person}{Marius Bock}, \bibinfo{person}{Alexander
  Hoelzemann}, \bibinfo{person}{Michael Moeller}, {and}
  \bibinfo{person}{Kristof Van~Laerhoven}.} \bibinfo{year}{2021}\natexlab{}.
\newblock \showarticletitle{Improving {Deep} {Learning} for {HAR} {With}
  {Shallow} {Lstms}}. In \bibinfo{booktitle}{\emph{{ACM} {International}
  {Symposium} on {Wearable} {Computers}}}.
\newblock
\urldef\tempurl%
\url{https://doi.org/10.1145/3460421.3480419}
\showURL{%
\tempurl}


\bibitem[Bulling et~al\mbox{.}(2014)]%
        {bullingTutorialHumanActivity2014}
\bibfield{author}{\bibinfo{person}{Andreas Bulling}, \bibinfo{person}{Ulf
  Blanke}, {and} \bibinfo{person}{Bernt Schiele}.}
  \bibinfo{year}{2014}\natexlab{}.
\newblock \showarticletitle{A {Tutorial} on {Human} {Activity} {Recognition}
  {Using} {Body}-{Worn} {Inertial} {Sensors}}.
\newblock \bibinfo{journal}{\emph{Comput. Surveys}} \bibinfo{volume}{46},
  \bibinfo{number}{3} (\bibinfo{year}{2014}), \bibinfo{pages}{1--33}.
\newblock
\urldef\tempurl%
\url{https://doi.org/10.1145/2499621}
\showURL{%
\tempurl}


\bibitem[Carreira and Zisserman(2017)]%
        {carreiraQuoVadisAction2017}
\bibfield{author}{\bibinfo{person}{Joao Carreira} {and} \bibinfo{person}{Andrew
  Zisserman}.} \bibinfo{year}{2017}\natexlab{}.
\newblock \showarticletitle{Quo {Vadis}, {Action} {Recognition}? {A} {New}
  {Model} and the {Kinetics} {Dataset}}. In \bibinfo{booktitle}{\emph{{IEEE}
  {Conference} on {Computer} {Vision} and {Pattern} {Recognition}}}.
\newblock
\urldef\tempurl%
\url{https://doi.org/10.1109/cvpr.2017.502}
\showURL{%
\tempurl}


\bibitem[Chen et~al\mbox{.}(2016)]%
        {chenFusionDepthSkeleton2016}
\bibfield{author}{\bibinfo{person}{Chen Chen}, \bibinfo{person}{Roozbeh
  Jafari}, {and} \bibinfo{person}{Nasser Kehtarnavaz}.}
  \bibinfo{year}{2016}\natexlab{}.
\newblock \showarticletitle{Fusion of {Depth}, {Skeleton}, and {Inertial}
  {Data} for {Human} {Action} {Recognition}}. In
  \bibinfo{booktitle}{\emph{{IEEE} {International} {Conference} on {Acoustics},
  {Speech} and {Signal} {Processing}}}.
\newblock
\urldef\tempurl%
\url{https://doi.org/10.1109/ICASSP.2016.7472170}
\showURL{%
\tempurl}


\bibitem[Chen et~al\mbox{.}(2022)]%
        {chenDCANImprovingTemporal2022}
\bibfield{author}{\bibinfo{person}{Guo Chen}, \bibinfo{person}{Yin-Dong Zheng},
  \bibinfo{person}{Limin Wang}, {and} \bibinfo{person}{Tong Lu}.}
  \bibinfo{year}{2022}\natexlab{}.
\newblock \showarticletitle{{DCAN}: {Improving} {Temporal} {Action} {Detection}
  via {Dual} {Context} {Aggregation}}. In \bibinfo{booktitle}{\emph{{AAAI}
  {Conference} on {Artificial} {Intelligence}}}.
\newblock
\urldef\tempurl%
\url{https://doi.org/10.1609/aaai.v36i1.19900}
\showURL{%
\tempurl}


\bibitem[Cheng and Bertasius(2022)]%
        {chengTallFormerTemporalAction2022}
\bibfield{author}{\bibinfo{person}{Feng Cheng} {and} \bibinfo{person}{Gedas
  Bertasius}.} \bibinfo{year}{2022}\natexlab{}.
\newblock \showarticletitle{{TallFormer}: {Temporal} {Action} {Localization}
  {With} a {Long}-{Memory} {Transformer}}. In
  \bibinfo{booktitle}{\emph{European {Conference} on {Computer} {Vision}}}.
\newblock
\urldef\tempurl%
\url{https://doi.org/10.1007/978-3-031-19830-4_29}
\showURL{%
\tempurl}


\bibitem[Damen et~al\mbox{.}(2022)]%
        {damenRescalingEgocentricVision2022}
\bibfield{author}{\bibinfo{person}{Dima Damen}, \bibinfo{person}{Hazel
  Doughty}, \bibinfo{person}{Giovanni~Maria Farinella},
  \bibinfo{person}{Antonino Furnari}, \bibinfo{person}{Jian Ma},
  \bibinfo{person}{Evangelos Kazakos}, \bibinfo{person}{Davide Moltisanti},
  \bibinfo{person}{Jonathan Munro}, \bibinfo{person}{Toby Perrett},
  \bibinfo{person}{Will Price}, {and} \bibinfo{person}{Michael Wray}.}
  \bibinfo{year}{2022}\natexlab{}.
\newblock \showarticletitle{Rescaling {Egocentric} {Vision}: {Collection},
  {Pipeline} and {Challenges} for {EPIC}-{KITCHENS}-100}.
\newblock \bibinfo{journal}{\emph{International Journal of Computer Vision}}
  (\bibinfo{year}{2022}).
\newblock
\urldef\tempurl%
\url{https://doi.org/10.1007/s11263-021-01531-2}
\showURL{%
\tempurl}


\bibitem[{de la Torre, Fernando} et~al\mbox{.}(2009)]%
        {delatorrefernandoDetailedHumanData2009}
\bibfield{author}{\bibinfo{person}{{de la Torre, Fernando}},
  \bibinfo{person}{Jessica~K. Hodgins}, \bibinfo{person}{Javier Montano}, {and}
  \bibinfo{person}{Sergio Valcarcel}.} \bibinfo{year}{2009}\natexlab{}.
\newblock \showarticletitle{Detailed {Human} {Data} {Acquisition} of {Kitchen}
  {Activities}: {The} {CMU}-{Multimodal} {Activity} {Database} ({CMU}-{MMAC})}.
  In \bibinfo{booktitle}{\emph{Conference on {Human} {Factors} in {Computing}
  {Systems}}}.
\newblock
\urldef\tempurl%
\url{http://www.cs.cmu.edu/~ftorre/web_page/humansensing.cs.cmu.edu/projects/CMU-MMAC.html}
\showURL{%
\tempurl}


\bibitem[DelPreto et~al\mbox{.}(2022)]%
        {delpretoActionSenseMultimodalDataset2022}
\bibfield{author}{\bibinfo{person}{Joseph DelPreto}, \bibinfo{person}{Chao
  Liu}, \bibinfo{person}{Yiyue Luo}, \bibinfo{person}{Michael Foshey},
  \bibinfo{person}{Yunzhu Li}, \bibinfo{person}{Antonio Torralba},
  \bibinfo{person}{Wojciech Matusik}, {and} \bibinfo{person}{Daniela Rus}.}
  \bibinfo{year}{2022}\natexlab{}.
\newblock \showarticletitle{{ActionSense}: {A} {Multimodal} {Dataset} and
  {Recording} {Framework} for {Human} {Activities} {Using} {Wearable} {Sensors}
  in a {Kitchen} {Environment}}. In \bibinfo{booktitle}{\emph{Neural
  {Information} {Processing} {Systems} {Track} on {Datasets} and
  {Benchmarks}}}.
\newblock
\urldef\tempurl%
\url{https://action-sense.csail.mit.edu}
\showURL{%
\tempurl}


\bibitem[Devlin et~al\mbox{.}(2019)]%
        {devlinBERTPretrainingDeep2019}
\bibfield{author}{\bibinfo{person}{Jacob Devlin}, \bibinfo{person}{Ming-Wei
  Chang}, \bibinfo{person}{Kenton Lee}, {and} \bibinfo{person}{Kristina~N.
  Toutanova}.} \bibinfo{year}{2019}\natexlab{}.
\newblock \showarticletitle{{BERT}: {Pre}-training of {Deep} {Bidirectional}
  {Transformers} for {Language} {Understanding}}. In
  \bibinfo{booktitle}{\emph{Confernce of the {North} {American} {Chapter} of
  the {Association} for {Computational} {Linguistics}}}.
\newblock
\urldef\tempurl%
\url{https://arxiv.org/abs/1810.04805}
\showURL{%
\tempurl}


\bibitem[Diete and Stuckenschmidt(2019)]%
        {dieteFusingObjectInformation2019}
\bibfield{author}{\bibinfo{person}{Alexander Diete} {and}
  \bibinfo{person}{Heiner Stuckenschmidt}.} \bibinfo{year}{2019}\natexlab{}.
\newblock \showarticletitle{Fusing {Object} {Information} and {Inertial} {Data}
  for {Activity} {Recognition}}.
\newblock \bibinfo{journal}{\emph{MDPI Sensors}} \bibinfo{volume}{19},
  \bibinfo{number}{19} (\bibinfo{year}{2019}).
\newblock
\urldef\tempurl%
\url{https://doi.org/10.3390/s19194119}
\showURL{%
\tempurl}


\bibitem[Diete et~al\mbox{.}(2019)]%
        {dieteVisionAccelerationModalities2019}
\bibfield{author}{\bibinfo{person}{Alexander Diete}, \bibinfo{person}{Timo
  Sztyler}, {and} \bibinfo{person}{Heiner Stuckenschmidt}.}
  \bibinfo{year}{2019}\natexlab{}.
\newblock \showarticletitle{Vision and {Acceleration} {Modalities}: {Partners}
  for {Recognizing} {Complex} {Activities}}. In
  \bibinfo{booktitle}{\emph{{IEEE} {International} {Conference} on {Pervasive}
  {Computing} and {Communications} {Workshops}}}.
\newblock
\urldef\tempurl%
\url{https://doi.org/10.1109/PERCOMW.2019.8730690}
\showURL{%
\tempurl}


\bibitem[Diete et~al\mbox{.}(2018)]%
        {dieteImprovingMotionbasedActivity2018}
\bibfield{author}{\bibinfo{person}{Alexander Diete}, \bibinfo{person}{Timo
  Sztyler}, \bibinfo{person}{Lydia Weiland}, {and} \bibinfo{person}{Heiner
  Stuckenschmidt}.} \bibinfo{year}{2018}\natexlab{}.
\newblock \showarticletitle{Improving {Motion}-{Based} {Activity} {Recognition}
  {With} {Ego}-{Centric} {Vision}}. In \bibinfo{booktitle}{\emph{{IEEE}
  {International} {Conference} on {Pervasive} {Computing} and {Communications}
  {Workshops}}}.
\newblock
\urldef\tempurl%
\url{https://doi.org/10.1109/PERCOMW.2018.8480334}
\showURL{%
\tempurl}


\bibitem[Ehatisham-Ul-Haq et~al\mbox{.}(2019)]%
        {ehatisham-ul-haqRobustHumanActivity2019}
\bibfield{author}{\bibinfo{person}{Muhammad Ehatisham-Ul-Haq},
  \bibinfo{person}{Ali Javed}, \bibinfo{person}{Muhammad~Awais Azam},
  \bibinfo{person}{Hafiz M.~A. Malik}, \bibinfo{person}{Aun Irtaza},
  \bibinfo{person}{Ik~Hyun Lee}, {and} \bibinfo{person}{Muhammad~Tariq
  Mahmood}.} \bibinfo{year}{2019}\natexlab{}.
\newblock \showarticletitle{Robust {Human} {Activity} {Recognition} {Using}
  {Multimodal} {Feature}-{Level} {Fusion}}.
\newblock \bibinfo{journal}{\emph{IEEE Access: Practical innovations, Open
  Solutions}}  \bibinfo{volume}{7} (\bibinfo{year}{2019}),
  \bibinfo{pages}{60736--60751}.
\newblock
\urldef\tempurl%
\url{https://doi.org/10.1109/ACCESS.2019.2913393}
\showURL{%
\tempurl}


\bibitem[Gao et~al\mbox{.}(2023)]%
        {gaoMMTSAMultiModalTemporal2023}
\bibfield{author}{\bibinfo{person}{Ziqi Gao}, \bibinfo{person}{Yuntao Wang},
  \bibinfo{person}{Jianguo Chen}, \bibinfo{person}{Junliang Xing},
  \bibinfo{person}{Shwetak Patel}, \bibinfo{person}{Xin Liu}, {and}
  \bibinfo{person}{Yuanchun Shi}.} \bibinfo{year}{2023}\natexlab{}.
\newblock \showarticletitle{{MMTSA}: {Multi}-{Modal} {Temporal} {Segment}
  {Attention} {Network} for {Efficient} {Human} {Activity} {Recognition}}.
\newblock \bibinfo{journal}{\emph{ACM on Interactive, Mobile, Wearable and
  Ubiquitous Technologies}} \bibinfo{volume}{7}, \bibinfo{number}{3}
  (\bibinfo{year}{2023}), \bibinfo{pages}{1--26}.
\newblock
\urldef\tempurl%
\url{https://doi.org/10.1145/3610872}
\showURL{%
\tempurl}


\bibitem[Girdhar and Grauman(2021)]%
        {girdharAnticipativeVideoTransformer2021}
\bibfield{author}{\bibinfo{person}{Rohit Girdhar} {and}
  \bibinfo{person}{Kristen Grauman}.} \bibinfo{year}{2021}\natexlab{}.
\newblock \showarticletitle{Anticipative {Video} {Transformer}}. In
  \bibinfo{booktitle}{\emph{{IEEE}/{CVF} {International} {Conference} on
  {Computer} {Vision}}}.
\newblock
\urldef\tempurl%
\url{https://doi.org/10.1109/iccv48922.2021.01325}
\showURL{%
\tempurl}


\bibitem[Gong et~al\mbox{.}(2020)]%
        {gongScaleMattersTemporal2020}
\bibfield{author}{\bibinfo{person}{Guoqiang Gong}, \bibinfo{person}{Liangfeng
  Zheng}, {and} \bibinfo{person}{Yadong Mu}.} \bibinfo{year}{2020}\natexlab{}.
\newblock \showarticletitle{Scale matters: {Temporal} scale aggregation network
  for precise action localization in untrimmed videos}. In
  \bibinfo{booktitle}{\emph{{IEEE} {International} {Conference} on {Multimedia}
  and {Expo}}}.
\newblock
\urldef\tempurl%
\url{https://doi.org/10.1109/ICME46284.2020.9102850}
\showURL{%
\tempurl}


\bibitem[Grauman et~al\mbox{.}(2022)]%
        {graumanEgo4DWorld0002022}
\bibfield{author}{\bibinfo{person}{Kristen Grauman}, \bibinfo{person}{Andrew
  Westbury}, \bibinfo{person}{Eugene Byrne}, \bibinfo{person}{Zachary Chavis},
  \bibinfo{person}{Antonino Furnari}, \bibinfo{person}{Rohit Girdhar},
  \bibinfo{person}{Jackson Hamburger}, \bibinfo{person}{Hao Jiang},
  \bibinfo{person}{Miao Liu}, {and} \bibinfo{person}{Xingyu Liu}.}
  \bibinfo{year}{2022}\natexlab{}.
\newblock \showarticletitle{{Ego4D}: {Around} the {World} in 3,000 {Hours} of
  {Egocentric} {Video}}. In \bibinfo{booktitle}{\emph{{IEEE}/{CVF} {Conference}
  on {Computer} {Vision} and {Pattern} {Recognition}}}.
\newblock
\urldef\tempurl%
\url{https://doi.org/10.1109/CVPR52688.2022.01842}
\showURL{%
\tempurl}


\bibitem[Grauman et~al\mbox{.}(2024)]%
        {graumanEgoExo4DUnderstandingSkilled2024}
\bibfield{author}{\bibinfo{person}{Kristen Grauman}, \bibinfo{person}{Andrew
  Westbury}, \bibinfo{person}{Lorenzo Torresani}, \bibinfo{person}{Kris
  Kitani}, \bibinfo{person}{Jitendra Malik}, {and} \bibinfo{person}{et al.}}
  \bibinfo{year}{2024}\natexlab{}.
\newblock \showarticletitle{Ego-{Exo4D}: {Understanding} skilled human activity
  from first- and third-person perspectives}. In
  \bibinfo{booktitle}{\emph{{IEEE}/{CVF} {Conference} on {Computer} {Vision}
  and {Pattern} {Recognition}}}.
\newblock


\bibitem[Guan and Plötz(2017)]%
        {guanEnsemblesDeepLSTM2017}
\bibfield{author}{\bibinfo{person}{Yu Guan} {and} \bibinfo{person}{Thomas
  Plötz}.} \bibinfo{year}{2017}\natexlab{}.
\newblock \showarticletitle{Ensembles of {Deep} {LSTM} {Learners} for
  {Activity} {Recognition} {Using} {Wearables}}.
\newblock \bibinfo{journal}{\emph{ACM on Interactive, Mobile, Wearable and
  Ubiquitous Technologies}} \bibinfo{volume}{1}, \bibinfo{number}{2}
  (\bibinfo{year}{2017}), \bibinfo{pages}{1--28}.
\newblock
\urldef\tempurl%
\url{https://doi.org/10.1145/3090076}
\showURL{%
\tempurl}


\bibitem[Hammerla et~al\mbox{.}(2016)]%
        {hammerlaDeepConvolutionalRecurrent2016}
\bibfield{author}{\bibinfo{person}{Nils~Y. Hammerla}, \bibinfo{person}{Shane
  Halloran}, {and} \bibinfo{person}{Thomas Ploetz}.}
  \bibinfo{year}{2016}\natexlab{}.
\newblock \showarticletitle{Deep, {Convolutional}, and {Recurrent} {Models} for
  {Human} {Activity} {Recognition} using {Wearables}}. In
  \bibinfo{booktitle}{\emph{Twenty-{Fifth} {International} {Joint} {Conference}
  on {Artificial} {Intelligence}}}.
\newblock
\urldef\tempurl%
\url{https://dl.acm.org/doi/10.5555/3060832.3060835}
\showURL{%
\tempurl}


\bibitem[Heilbron et~al\mbox{.}(2015)]%
        {heilbronActivityNetLargescaleVideo2015}
\bibfield{author}{\bibinfo{person}{Fabian~Caba Heilbron},
  \bibinfo{person}{Juan~Carlos Niebles}, \bibinfo{person}{Victor Escorcia},
  {and} \bibinfo{person}{Bernard Ghanem}.} \bibinfo{year}{2015}\natexlab{}.
\newblock \showarticletitle{{ActivityNet}: {A} {Large}-{Scale} {Video}
  {Benchmark} for {Human} {Activity} {Understanding}}. In
  \bibinfo{booktitle}{\emph{{IEEE} {Conference} on {Computer} {Vision} and
  {Pattern} {Recognition}}}.
\newblock
\urldef\tempurl%
\url{https://doi.org/10.1109/cvpr.2015.7298698}
\showURL{%
\tempurl}


\bibitem[Hu et~al\mbox{.}(2023)]%
        {huMultimodalHumanActivity2023}
\bibfield{author}{\bibinfo{person}{Menghao Hu}, \bibinfo{person}{Mingxuan Luo},
  \bibinfo{person}{Menghua Huang}, \bibinfo{person}{Wenhua Meng},
  \bibinfo{person}{Baochen Xiong}, \bibinfo{person}{Xiaoshan Yang}, {and}
  \bibinfo{person}{Jitao Sang}.} \bibinfo{year}{2023}\natexlab{}.
\newblock \showarticletitle{Towards a {Multimodal} {Human} {Activity} {Dataset}
  for {Healthcare}}.
\newblock \bibinfo{journal}{\emph{Multimedia Systems}} \bibinfo{volume}{29},
  \bibinfo{number}{1} (\bibinfo{year}{2023}), \bibinfo{pages}{1--13}.
\newblock
\urldef\tempurl%
\url{https://doi.org/10.1007/s00530-021-00875-6}
\showURL{%
\tempurl}


\bibitem[Ijaz et~al\mbox{.}(2022)]%
        {ijazMultimodalTransformerNursing2022}
\bibfield{author}{\bibinfo{person}{Momal Ijaz}, \bibinfo{person}{Renato Diaz},
  {and} \bibinfo{person}{Chen Chen}.} \bibinfo{year}{2022}\natexlab{}.
\newblock \showarticletitle{Multimodal {Transformer} for {Nursing} {Activity}
  {Recognition}}. In \bibinfo{booktitle}{\emph{{IEEE}/{CVF} {Conference} on
  {Computer} {Vision} and {Pattern} {Recognition} {Workshops}}}.
\newblock
\urldef\tempurl%
\url{https://doi.org/10.1109/cvprw56347.2022.00224}
\showURL{%
\tempurl}


\bibitem[Imran and Raman(2020a)]%
        {imranEvaluatingFusionRGBD2020}
\bibfield{author}{\bibinfo{person}{Javed Imran} {and}
  \bibinfo{person}{Balasubramanian Raman}.} \bibinfo{year}{2020}\natexlab{a}.
\newblock \showarticletitle{Evaluating {Fusion} of {RGB}-{D} and {Inertial}
  {Sensors} for {Multimodal} {Human} {Action} {Recognition}}.
\newblock \bibinfo{journal}{\emph{Journal of Ambient Intelligence and Humanized
  Computing}} \bibinfo{volume}{11}, \bibinfo{number}{1} (\bibinfo{year}{2020}),
  \bibinfo{pages}{189--208}.
\newblock
\urldef\tempurl%
\url{https://doi.org/10.1007/s12652-019-01239-9}
\showURL{%
\tempurl}


\bibitem[Imran and Raman(2020b)]%
        {imranMultimodalEgocentricActivity2020}
\bibfield{author}{\bibinfo{person}{Javed Imran} {and}
  \bibinfo{person}{Balasubramanian Raman}.} \bibinfo{year}{2020}\natexlab{b}.
\newblock \showarticletitle{Multimodal {Egocentric} {Activity} {Recognition}
  {Using} {Multi}-{Stream} {CNN}}. In \bibinfo{booktitle}{\emph{11th {Indian}
  {Conference} on {Computer} {Vision}, {Graphics} and {Image} {Processing}}}.
\newblock
\urldef\tempurl%
\url{https://doi.org/10.1145/3293353.3293363}
\showURL{%
\tempurl}


\bibitem[Islam and Iqbal(2021)]%
        {islamMultiGATGraphicalAttentionbased2021}
\bibfield{author}{\bibinfo{person}{Md~Mofijul Islam} {and}
  \bibinfo{person}{Tariq Iqbal}.} \bibinfo{year}{2021}\natexlab{}.
\newblock \showarticletitle{Multi-{GAT}: {A} {Graphical} {Attention}-{Based}
  {Hierarchical} {Multimodal} {Representation} {Learning} {Approach} for
  {Human} {Activity} {Recognition}}.
\newblock \bibinfo{journal}{\emph{IEEE Robotics and Automation Letters}}
  \bibinfo{volume}{6}, \bibinfo{number}{2} (\bibinfo{year}{2021}),
  \bibinfo{pages}{1729--1736}.
\newblock
\urldef\tempurl%
\url{https://doi.org/10.1109/LRA.2021.3059624}
\showURL{%
\tempurl}


\bibitem[Islam and Iqbal(2022)]%
        {islamMuMuCooperativeMultitask2022}
\bibfield{author}{\bibinfo{person}{Md~Mofijul Islam} {and}
  \bibinfo{person}{Tariq Iqbal}.} \bibinfo{year}{2022}\natexlab{}.
\newblock \showarticletitle{{MuMu}: {Cooperative} {Multitask}
  {Learning}-{Based} {Guided} {Multimodal} {Fusion}}. In
  \bibinfo{booktitle}{\emph{{AAAI} {Conference} on {Artificial}
  {Intelligence}}}.
\newblock
\urldef\tempurl%
\url{https://doi.org/10.1609/aaai.v36i1.19988}
\showURL{%
\tempurl}


\bibitem[Jiang et~al\mbox{.}(2014)]%
        {jiangTHUMOSChallengeAction2014}
\bibfield{author}{\bibinfo{person}{Y.-G. Jiang}, \bibinfo{person}{J. Liu},
  \bibinfo{person}{A. Roshan~Zamir}, \bibinfo{person}{G. Toderici},
  \bibinfo{person}{I. Laptev}, \bibinfo{person}{M. Shah}, {and}
  \bibinfo{person}{R. Sukthankar}.} \bibinfo{year}{2014}\natexlab{}.
\newblock \bibinfo{title}{{THUMOS} challenge: {Action} {Recognition} {With} a
  {Large} {Number} of {Classes}}.
\newblock
\newblock
\urldef\tempurl%
\url{http://crcv.ucf.edu/THUMOS14/}
\showURL{%
\tempurl}


\bibitem[Karpathy et~al\mbox{.}(2015)]%
        {karpathyVisualizingUnderstandingRecurrent2015}
\bibfield{author}{\bibinfo{person}{Andrej Karpathy}, \bibinfo{person}{Justin
  Johnson}, {and} \bibinfo{person}{Fei-Fei Li}.}
  \bibinfo{year}{2015}\natexlab{}.
\newblock \showarticletitle{Visualizing and {Understanding} {Recurrent}
  {Networks}}.
\newblock \bibinfo{journal}{\emph{CoRR}}  \bibinfo{volume}{abs/1506.02078}
  (\bibinfo{year}{2015}).
\newblock
\urldef\tempurl%
\url{http://arxiv.org/abs/1506.02078}
\showURL{%
\tempurl}


\bibitem[Kay et~al\mbox{.}(2017)]%
        {kayKineticsHumanAction2017}
\bibfield{author}{\bibinfo{person}{Will Kay}, \bibinfo{person}{João Carreira},
  \bibinfo{person}{Karen Simonyan}, \bibinfo{person}{Brian Zhang},
  \bibinfo{person}{Chloe Hillier}, \bibinfo{person}{Sudheendra
  Vijayanarasimhan}, \bibinfo{person}{Fabio Viola}, \bibinfo{person}{Tim
  Green}, \bibinfo{person}{Trevor Back}, \bibinfo{person}{Paul Natsev},
  \bibinfo{person}{Mustafa Suleyman}, {and} \bibinfo{person}{Andrew
  Zisserman}.} \bibinfo{year}{2017}\natexlab{}.
\newblock \showarticletitle{The {Kinetics} {Human} {Action} {Video} {Dataset}}.
\newblock \bibinfo{journal}{\emph{CoRR}}  \bibinfo{volume}{abs/1705.06950}
  (\bibinfo{year}{2017}).
\newblock
\urldef\tempurl%
\url{http://arxiv.org/abs/1705.06950}
\showURL{%
\tempurl}


\bibitem[Kolesnikov et~al\mbox{.}(2021)]%
        {kolesnikovImageWorth16x162021}
\bibfield{author}{\bibinfo{person}{Alexander Kolesnikov},
  \bibinfo{person}{Alexey Dosovitskiy}, \bibinfo{person}{Dirk Weissenborn},
  \bibinfo{person}{Georg Heigold}, \bibinfo{person}{Jakob Uszkoreit},
  \bibinfo{person}{Lucas Beyer}, \bibinfo{person}{Matthias Minderer},
  \bibinfo{person}{Mostafa Dehghani}, \bibinfo{person}{Neil Houlsby},
  \bibinfo{person}{Sylvain Gelly}, \bibinfo{person}{Thomas Unterthiner}, {and}
  \bibinfo{person}{Xiaohua Zhai}.} \bibinfo{year}{2021}\natexlab{}.
\newblock \showarticletitle{An {Image} {Is} {Worth} 16x16 {Words}:
  {Transformers} for {Image} {Recognition} at {Scale}}. In
  \bibinfo{booktitle}{\emph{Ninth {International} {Conference} on {Learning}
  {Representations}}}.
\newblock
\urldef\tempurl%
\url{https://arxiv.org/abs/2010.11929}
\showURL{%
\tempurl}


\bibitem[Li et~al\mbox{.}(2022)]%
        {liMViTv2ImprovedMultiscale2022}
\bibfield{author}{\bibinfo{person}{Yanghao Li}, \bibinfo{person}{Chao-Yuan Wu},
  \bibinfo{person}{Haoqi Fan}, \bibinfo{person}{Karttikeya Mangalam},
  \bibinfo{person}{Bo Xiong}, \bibinfo{person}{Jitendra Malik}, {and}
  \bibinfo{person}{Christoph Feichtenhofer}.} \bibinfo{year}{2022}\natexlab{}.
\newblock \showarticletitle{{MViTv2}: {Improved} {Multiscale} {Vision}
  {Transformers} for {Classification} and {Detection}}. In
  \bibinfo{booktitle}{\emph{{IEEE}/ {CVF} {Computer} {Vision} and {Pattern}
  {Recognition}}}.
\newblock
\urldef\tempurl%
\url{https://doi.org/10.1109/cvpr52688.2022.00476}
\showURL{%
\tempurl}


\bibitem[Lin et~al\mbox{.}(2020)]%
        {linFastLearningTemporal2020}
\bibfield{author}{\bibinfo{person}{Chuming Lin}, \bibinfo{person}{Jian Li},
  \bibinfo{person}{Yabiao Wang}, \bibinfo{person}{Ying Tai},
  \bibinfo{person}{Donghao Luo}, \bibinfo{person}{Zhipeng Cui},
  \bibinfo{person}{Chengjie Wang}, \bibinfo{person}{Jilin Li},
  \bibinfo{person}{Feiyue Huang}, {and} \bibinfo{person}{Rongrong Ji}.}
  \bibinfo{year}{2020}\natexlab{}.
\newblock \showarticletitle{Fast {Learning} of {Temporal} {Action} {Proposal}
  via {Dense} {Boundary} {Generator}}. In \bibinfo{booktitle}{\emph{{IEEE}
  {Conference} on {Computer} {Vision} and {Pattern} {Recognition}}}.
\newblock
\urldef\tempurl%
\url{https://doi.org/10.1609/aaai.v34i07.6815}
\showURL{%
\tempurl}


\bibitem[Lin et~al\mbox{.}(2021)]%
        {linLearningSalientBoundary2021}
\bibfield{author}{\bibinfo{person}{Chuming Lin}, \bibinfo{person}{Chengming
  Xu}, \bibinfo{person}{Donghao Luo}, \bibinfo{person}{Yabiao Wang},
  \bibinfo{person}{Ying Tai}, \bibinfo{person}{Chengjie Wang},
  \bibinfo{person}{Jilin Li}, \bibinfo{person}{Feiyue Huang}, {and}
  \bibinfo{person}{Yanwei Fu}.} \bibinfo{year}{2021}\natexlab{}.
\newblock \showarticletitle{Learning {Salient} {Boundary} {Feature} for
  {Anchor}-{Free} {Temporal} {Action} {Localization}}. In
  \bibinfo{booktitle}{\emph{{IEEE}/{CVF} {Conference} on {Computer} {Vision}
  and {Pattern} {Recognition}}}.
\newblock
\urldef\tempurl%
\url{https://doi.org/10.1109/cvpr46437.2021.00333}
\showURL{%
\tempurl}


\bibitem[Lin et~al\mbox{.}(2019)]%
        {linBMNBoundarymatchingNetwork2019}
\bibfield{author}{\bibinfo{person}{Tianwei Lin}, \bibinfo{person}{Xiao Liu},
  \bibinfo{person}{Xin Li}, \bibinfo{person}{Errui Ding}, {and}
  \bibinfo{person}{Shilei Wen}.} \bibinfo{year}{2019}\natexlab{}.
\newblock \showarticletitle{{BMN}: {Boundary}-{Matching} {Network} for
  {Temporal} {Action} {Proposal} {Generation}}. In
  \bibinfo{booktitle}{\emph{{IEEE}/{CVF} {International} {Conference} on
  {Computer} {Vision}}}.
\newblock
\urldef\tempurl%
\url{https://doi.org/10.1109/iccv.2019.00399}
\showURL{%
\tempurl}


\bibitem[Liu and Wang(2020)]%
        {liuProgressiveBoundaryRefinement2020}
\bibfield{author}{\bibinfo{person}{Qinying Liu} {and} \bibinfo{person}{Zilei
  Wang}.} \bibinfo{year}{2020}\natexlab{}.
\newblock \showarticletitle{Progressive {Boundary} {Refinement} {Network} for
  {Temporal} {Action} {Detection}}. In \bibinfo{booktitle}{\emph{{AAAI}
  {Conference} on {Artificial} {Intelligence}}}.
\newblock
\urldef\tempurl%
\url{https://doi.org/10.1609/aaai.v34i07.6829}
\showURL{%
\tempurl}


\bibitem[Liu et~al\mbox{.}(2022a)]%
        {liuEmpiricalStudyEndtoend2022}
\bibfield{author}{\bibinfo{person}{Xiaolong Liu}, \bibinfo{person}{Song Bai},
  {and} \bibinfo{person}{Xiang Bai}.} \bibinfo{year}{2022}\natexlab{a}.
\newblock \showarticletitle{An {Empirical} {Study} of {End}-{To}-{End}
  {Temporal} {Action} {Detection}}. In \bibinfo{booktitle}{\emph{{IEEE}/{CVF}
  {Conference} on {Computer} {Vision} and {Pattern} {Recognition}}}.
\newblock
\urldef\tempurl%
\url{https://doi.org/10.1109/cvpr52688.2022.01938}
\showURL{%
\tempurl}


\bibitem[Liu et~al\mbox{.}(2021a)]%
        {liuMultishotTemporalEvent2021}
\bibfield{author}{\bibinfo{person}{Xiaolong Liu}, \bibinfo{person}{Yao Hu},
  \bibinfo{person}{Song Bai}, \bibinfo{person}{Fei Ding},
  \bibinfo{person}{Xiang Bai}, {and} \bibinfo{person}{Philip H.~S. Torr}.}
  \bibinfo{year}{2021}\natexlab{a}.
\newblock \showarticletitle{Multi-{Shot} {Temporal} {Event} {Localization}: {A}
  {Benchmark}}. In \bibinfo{booktitle}{\emph{{IEEE}/{CVF} {Conference} on
  {Computer} {Vision} and {Pattern} {Recognition}}}.
\newblock
\urldef\tempurl%
\url{https://doi.org/10.1109/cvpr46437.2021.01241}
\showURL{%
\tempurl}


\bibitem[Liu et~al\mbox{.}(2022b)]%
        {liuEndtoendTemporalAction2022}
\bibfield{author}{\bibinfo{person}{Xiaolong Liu}, \bibinfo{person}{Qimeng
  Wang}, \bibinfo{person}{Yao Hu}, \bibinfo{person}{Xu Tang},
  \bibinfo{person}{Shiwei Zhang}, \bibinfo{person}{Song Bai}, {and}
  \bibinfo{person}{Xiang Bai}.} \bibinfo{year}{2022}\natexlab{b}.
\newblock \showarticletitle{End-{To}-{End} {Temporal} {Action} {Detection}
  {With} {Transformer}}.
\newblock \bibinfo{journal}{\emph{IEEE Transactions on Image Processing}}
  \bibinfo{volume}{31} (\bibinfo{year}{2022}).
\newblock
\urldef\tempurl%
\url{https://doi.org/10.1109/TIP.2022.3195321}
\showURL{%
\tempurl}


\bibitem[Liu et~al\mbox{.}(2021b)]%
        {liuSwinTransformerHierarchical2021}
\bibfield{author}{\bibinfo{person}{Ze Liu}, \bibinfo{person}{Yutong Lin},
  \bibinfo{person}{Yue Cao}, \bibinfo{person}{Han Hu}, \bibinfo{person}{Yixuan
  Wei}, \bibinfo{person}{Zheng Zhang}, \bibinfo{person}{Stephen Lin}, {and}
  \bibinfo{person}{Baining Guo}.} \bibinfo{year}{2021}\natexlab{b}.
\newblock \showarticletitle{Swin {Transformer}: {Hierarchical} {Vision}
  {Transformer} using {Shifted} {Windows}}. In
  \bibinfo{booktitle}{\emph{{IEEE}/{CVF} {International} {Conference} on
  {Computer} {Vision}}}.
\newblock
\urldef\tempurl%
\url{https://doi.org/10.1109/iccv48922.2021.00986}
\showURL{%
\tempurl}


\bibitem[Long et~al\mbox{.}(2019)]%
        {longGaussianTemporalAwareness2019}
\bibfield{author}{\bibinfo{person}{Fuchen Long}, \bibinfo{person}{Ting Yao},
  \bibinfo{person}{Zhaofan Qiu}, \bibinfo{person}{Xinmei Tian},
  \bibinfo{person}{Jiebo Luo}, {and} \bibinfo{person}{Tao Mei}.}
  \bibinfo{year}{2019}\natexlab{}.
\newblock \showarticletitle{Gaussian {Temporal} {Awareness} {Networks} for
  {Action} {Localization}}. In \bibinfo{booktitle}{\emph{{IEEE}/{CVF}
  {Conference} on {Computer} {Vision} and {Pattern} {Recognition}}}.
\newblock
\urldef\tempurl%
\url{https://doi.org/10.1109/cvpr.2019.00043}
\showURL{%
\tempurl}


\bibitem[Lu and Velipasalar(2018)]%
        {luHumanActivityClassification2018}
\bibfield{author}{\bibinfo{person}{Yantao Lu} {and} \bibinfo{person}{Senem
  Velipasalar}.} \bibinfo{year}{2018}\natexlab{}.
\newblock \showarticletitle{Human {Activity} {Classification} {Incorporating}
  {Egocentric} {Video} and {Inertial} {Measurement} {Unit} {Data}}. In
  \bibinfo{booktitle}{\emph{{IEEE} {Global} {Conference} on {Signal} and
  {Information} {Processing}}}.
\newblock
\urldef\tempurl%
\url{https://doi.org/10.1109/GlobalSIP.2018.8646367}
\showURL{%
\tempurl}


\bibitem[Murahari and Plötz(2018)]%
        {murahariAttentionModelsHuman2018}
\bibfield{author}{\bibinfo{person}{Vishvak~S. Murahari} {and}
  \bibinfo{person}{Thomas Plötz}.} \bibinfo{year}{2018}\natexlab{}.
\newblock \showarticletitle{On {Attention} {Models} for {Human} {Activity}
  {Recognition}}. In \bibinfo{booktitle}{\emph{{ACM} {International}
  {Symposium} on {Wearable} {Computers}}}.
\newblock
\urldef\tempurl%
\url{https://doi.org/10.1145/3267242.3267287}
\showURL{%
\tempurl}


\bibitem[Nag et~al\mbox{.}(2022)]%
        {nagProposalfreeTemporalAction2022}
\bibfield{author}{\bibinfo{person}{Sauradip Nag}, \bibinfo{person}{Xiatian
  Zhu}, \bibinfo{person}{Yi-Zhe Song}, {and} \bibinfo{person}{Tao Xiang}.}
  \bibinfo{year}{2022}\natexlab{}.
\newblock \showarticletitle{Proposal-{Free} {Temporal} {Action} {Detection} via
  {Global} {Segmentation} {Mask} {Learning}}. In
  \bibinfo{booktitle}{\emph{European {Conferencee} on {Computer} {Vision}}}.
\newblock
\urldef\tempurl%
\url{https://doi.org/10.1007/978-3-031-20062-5_37}
\showURL{%
\tempurl}


\bibitem[Nakamura et~al\mbox{.}(2017)]%
        {nakamuraJointlyLearningEnergy2017}
\bibfield{author}{\bibinfo{person}{Katsuyuki Nakamura}, \bibinfo{person}{Serena
  Yeung}, \bibinfo{person}{Alexandre Alahi}, {and} \bibinfo{person}{Li
  Fei-Fei}.} \bibinfo{year}{2017}\natexlab{}.
\newblock \showarticletitle{Jointly {Learning} {Energy} {Expenditures} and
  {Activities} {Using} {Egocentric} {Multimodal} {Signals}}. In
  \bibinfo{booktitle}{\emph{{IEEE} {Conference} on {Computer} {Vision} and
  {Pattern} {Recognition}}}.
\newblock
\urldef\tempurl%
\url{https://doi.org/10.1109/CVPR.2017.721}
\showURL{%
\tempurl}


\bibitem[Ordóñez and Roggen(2016)]%
        {ordonezDeepConvolutionalLSTM2016}
\bibfield{author}{\bibinfo{person}{Francisco~Javier Ordóñez} {and}
  \bibinfo{person}{Daniel Roggen}.} \bibinfo{year}{2016}\natexlab{}.
\newblock \showarticletitle{Deep {Convolutional} and {LSTM} {Recurrent}
  {Neural} {Networks} for {Multimodal} {Wearable} {Activity} {Recognition}}.
\newblock \bibinfo{journal}{\emph{MDPI Sensors}} \bibinfo{volume}{16},
  \bibinfo{number}{1} (\bibinfo{year}{2016}).
\newblock
\urldef\tempurl%
\url{https://doi.org/10.3390/s16010115}
\showURL{%
\tempurl}


\bibitem[Patterson et~al\mbox{.}(2005)]%
        {pattersonFineGrainedActivityRecognition2005}
\bibfield{author}{\bibinfo{person}{Donald~J. Patterson},
  \bibinfo{person}{Dieter Fox}, \bibinfo{person}{Henry Kautz}, {and}
  \bibinfo{person}{Matthai Philipose}.} \bibinfo{year}{2005}\natexlab{}.
\newblock \showarticletitle{Fine-{Grained} {Activity} {Recognition} by
  {Aggregating} {Abstract} {Object} {Usage}}. In
  \bibinfo{booktitle}{\emph{Ninth {IEEE} {International} {Symposium} on
  {Wearable} {Computers}}}.
\newblock
\urldef\tempurl%
\url{https://doi.org/10.1109/ISWC.2005.22}
\showURL{%
\tempurl}


\bibitem[Possas et~al\mbox{.}(2018)]%
        {possasEgocentricActivityRecognition2018}
\bibfield{author}{\bibinfo{person}{Rafael Possas},
  \bibinfo{person}{Sheila~Pinto Caceres}, {and} \bibinfo{person}{Fabio Ramos}.}
  \bibinfo{year}{2018}\natexlab{}.
\newblock \showarticletitle{Egocentric {Activity} {Recognition} on a {Budget}}.
  In \bibinfo{booktitle}{\emph{{IEEE}/{CVF} {Conference} on {Computer} {Vision}
  and {Pattern} {Recognition}}}.
\newblock
\urldef\tempurl%
\url{https://doi.org/10.1109/CVPR.2018.00625}
\showURL{%
\tempurl}


\bibitem[Qing et~al\mbox{.}(2021)]%
        {qingTemporalContextAggregation2021}
\bibfield{author}{\bibinfo{person}{Zhiwu Qing}, \bibinfo{person}{Haisheng Su},
  \bibinfo{person}{Weihao Gan}, \bibinfo{person}{Dongliang Wang},
  \bibinfo{person}{Wei Wu}, \bibinfo{person}{Xiang Wang}, \bibinfo{person}{Yu
  Qiao}, \bibinfo{person}{Junjie Yan}, \bibinfo{person}{Changxin Gao}, {and}
  \bibinfo{person}{Nong Sang}.} \bibinfo{year}{2021}\natexlab{}.
\newblock \showarticletitle{Temporal {Context} {Aggregation} {Network} for
  {Temporal} {Action} {Proposal} {Refinement}}. In
  \bibinfo{booktitle}{\emph{{IEEE}/{CVF} {Conference} on {Computer} {Vision}
  and {Pattern} {Recognition}}}.
\newblock
\urldef\tempurl%
\url{https://doi.org/10.1109/cvpr46437.2021.00055}
\showURL{%
\tempurl}


\bibitem[Radu and Henne(2019)]%
        {raduVision2SensorKnowledgeTransfer2019}
\bibfield{author}{\bibinfo{person}{Valentin Radu} {and}
  \bibinfo{person}{Maximilian Henne}.} \bibinfo{year}{2019}\natexlab{}.
\newblock \showarticletitle{{Vision2Sensor}: {Knowledge} {Transfer} {Across}
  {Sensing} {Modalities} for {Human} {Activity} {Recognition}}.
\newblock \bibinfo{journal}{\emph{ACM on Interactive, Mobile, Wearable and
  Ubiquitous Technologies}} \bibinfo{volume}{3}, \bibinfo{number}{3}
  (\bibinfo{year}{2019}), \bibinfo{pages}{1--21}.
\newblock
\urldef\tempurl%
\url{https://doi.org/10.1145/3351242}
\showURL{%
\tempurl}


\bibitem[Roy and Fernando(2022)]%
        {royPredictingNextAction2022}
\bibfield{author}{\bibinfo{person}{Debaditya Roy} {and} \bibinfo{person}{Basura
  Fernando}.} \bibinfo{year}{2022}\natexlab{}.
\newblock \showarticletitle{Predicting the {Next} {Action} by {Modeling} the
  {Abstract} {Goal}}.
\newblock \bibinfo{journal}{\emph{CoRR}}  \bibinfo{volume}{abs/2209.05044}
  (\bibinfo{year}{2022}).
\newblock
\urldef\tempurl%
\url{https://doi.org/10.48550/arxiv.2209.05044}
\showURL{%
\tempurl}


\bibitem[Scholl et~al\mbox{.}(2019)]%
        {schollMultimediaExchangeFormat2019}
\bibfield{author}{\bibinfo{person}{Philipp~M. Scholl},
  \bibinfo{person}{Benjamin Völker}, \bibinfo{person}{Bernd Becker}, {and}
  \bibinfo{person}{Kristof~Van Laerhoven}.} \bibinfo{year}{2019}\natexlab{}.
\newblock \showarticletitle{A {Multi}-{Media} {Exchange} {Format} for
  {Time}-{Series} {Dataset} {Curation}}.
\newblock In \bibinfo{booktitle}{\emph{Human {Activity} {Sensing}}},
  \bibfield{editor}{\bibinfo{person}{Nobuo Kawaguchi},
  \bibinfo{person}{Nobuhiko Nishio}, \bibinfo{person}{Daniel Roggen},
  \bibinfo{person}{Sozo Inoue}, \bibinfo{person}{Susanna Pirttikangas}, {and}
  \bibinfo{person}{Kristof Van~Laerhoven}} (Eds.).
  \bibinfo{publisher}{Springer}, \bibinfo{pages}{111--119}.
\newblock
\urldef\tempurl%
\url{https://doi.org/10.1007/978-3-030-13001-5_8}
\showURL{%
\tempurl}


\bibitem[Shi et~al\mbox{.}(2023)]%
        {shiTriDetTemporalAction2023}
\bibfield{author}{\bibinfo{person}{Dingfeng Shi}, \bibinfo{person}{Yujie
  Zhong}, \bibinfo{person}{Qiong Cao}, \bibinfo{person}{Lin Ma},
  \bibinfo{person}{Jia Li}, {and} \bibinfo{person}{Dacheng Tao}.}
  \bibinfo{year}{2023}\natexlab{}.
\newblock \showarticletitle{{TriDet}: {Temporal} {Action} {Detection} {With}
  {Relative} {Boundary} {Modeling}}. In \bibinfo{booktitle}{\emph{{IEEE}/{CVF}
  {Conference} on {Computer} {Vision} and {Pattern} {Recognition}}}.
\newblock
\urldef\tempurl%
\url{https://doi.org/10.1109/cvpr52729.2023.01808}
\showURL{%
\tempurl}


\bibitem[Shi et~al\mbox{.}(2022)]%
        {shiReActTemporalAction2022}
\bibfield{author}{\bibinfo{person}{Dingfeng Shi}, \bibinfo{person}{Yujie
  Zhong}, \bibinfo{person}{Qiong Cao}, \bibinfo{person}{Jing Zhang},
  \bibinfo{person}{Lin Ma}, \bibinfo{person}{Jia Li}, {and}
  \bibinfo{person}{Dacheng Tao}.} \bibinfo{year}{2022}\natexlab{}.
\newblock \showarticletitle{{ReAct}: {Temporal} {Action} {Detection} {With}
  {Relational} {Queries}}. In \bibinfo{booktitle}{\emph{European {Conference}
  on {Computer} {Vision}}}.
\newblock
\urldef\tempurl%
\url{https://doi.org/10.1007/978-3-031-20080-9_7}
\showURL{%
\tempurl}


\bibitem[Song et~al\mbox{.}(2016a)]%
        {songMultimodalMultistreamDeep2016}
\bibfield{author}{\bibinfo{person}{Sibo Song}, \bibinfo{person}{Vijay
  Chandrasekhar}, \bibinfo{person}{Bappaditya Mandal}, \bibinfo{person}{Liyuan
  Li}, \bibinfo{person}{Joo-Hwee Lim}, \bibinfo{person}{Giduthuri~Sateesh
  Babu}, \bibinfo{person}{Phyo~Phyo San}, {and} \bibinfo{person}{Ngai-Man
  Cheung}.} \bibinfo{year}{2016}\natexlab{a}.
\newblock \showarticletitle{Multimodal {Multi}-{Stream} {Deep} {Learning} for
  {Egocentric} {Activity} {Recognition}}. In \bibinfo{booktitle}{\emph{{IEEE}
  conference on computer vision and pattern recognition workshops}}.
\newblock
\urldef\tempurl%
\url{https://doi.org/10.1109/CVPRW.2016.54}
\showURL{%
\tempurl}


\bibitem[Song et~al\mbox{.}(2016b)]%
        {songEgocentricActivityRecognition2016}
\bibfield{author}{\bibinfo{person}{Sibo Song}, \bibinfo{person}{Ngai-Man
  Cheung}, \bibinfo{person}{Vijay Chandrasekhar}, \bibinfo{person}{Bappaditya
  Mandal}, {and} \bibinfo{person}{Jie Lin}.} \bibinfo{year}{2016}\natexlab{b}.
\newblock \showarticletitle{Egocentric {Activity} {Recognition} {With}
  {Multimodal} {Fisher} {Vector}}. In \bibinfo{booktitle}{\emph{{IEEE}
  {International} {Conference} on {Acoustics}, {Speech} and {Signal}
  {Processing}}}.
\newblock
\urldef\tempurl%
\url{https://doi.org/10.1109/icassp.2016.7472171}
\showURL{%
\tempurl}


\bibitem[Spriggs et~al\mbox{.}(2009)]%
        {spriggsTemporalSegmentationActivity2009}
\bibfield{author}{\bibinfo{person}{Ekaterina~H. Spriggs},
  \bibinfo{person}{Fernando De~La~Torre}, {and} \bibinfo{person}{Martial
  Hebert}.} \bibinfo{year}{2009}\natexlab{}.
\newblock \showarticletitle{Temporal {Segmentation} and {Activity}
  {Classification} {From} {First}-{Person} {Sensing}}. In
  \bibinfo{booktitle}{\emph{{IEEE} {Computer} {Society} {Conference} on
  {Computer} {Vision} and {Pattern} {Recognition} {Workshops}}}.
\newblock
\urldef\tempurl%
\url{https://doi.org/10.1109/CVPRW.2009.5204354}
\showURL{%
\tempurl}


\bibitem[Sridhar et~al\mbox{.}(2021)]%
        {sridharClassSemanticsbasedAttention2021}
\bibfield{author}{\bibinfo{person}{Deepak Sridhar}, \bibinfo{person}{Niamul
  Quader}, \bibinfo{person}{Srikanth Muralidharan}, \bibinfo{person}{Yaoxin
  Li}, \bibinfo{person}{Peng Dai}, {and} \bibinfo{person}{Juwei Lu}.}
  \bibinfo{year}{2021}\natexlab{}.
\newblock \showarticletitle{Class {Semantics}-{Based} {Attention} for {Action}
  {Detection}}. In \bibinfo{booktitle}{\emph{{IEEE}/ {CVF} {International}
  {Conference} on {Computer} {Vision}}}.
\newblock
\urldef\tempurl%
\url{https://doi.org/10.1109/iccv48922.2021.01348}
\showURL{%
\tempurl}


\bibitem[Strömbäck et~al\mbox{.}(2020)]%
        {strombackMMFitMultimodalDeep2020}
\bibfield{author}{\bibinfo{person}{David Strömbäck}, \bibinfo{person}{Sangxia
  Huang}, {and} \bibinfo{person}{Valentin Radu}.}
  \bibinfo{year}{2020}\natexlab{}.
\newblock \showarticletitle{{MM}-{Fit}: {Multimodal} {Deep} {Learning} for
  {Automatic} {Exercise} {Logging} across {Sensing} {Devices}}.
\newblock \bibinfo{journal}{\emph{ACM on Interactive, Mobile, Wearable and
  Ubiquitous Technologies}} \bibinfo{volume}{4}, \bibinfo{number}{4}
  (\bibinfo{year}{2020}), \bibinfo{pages}{1--22}.
\newblock
\urldef\tempurl%
\url{https://doi.org/10.1145/3432701}
\showURL{%
\tempurl}


\bibitem[Tan et~al\mbox{.}(2021)]%
        {tanRelaxedTransformerDecoders2021}
\bibfield{author}{\bibinfo{person}{Jing Tan}, \bibinfo{person}{Jiaqi Tang},
  \bibinfo{person}{Limin Wang}, {and} \bibinfo{person}{Gangshan Wu}.}
  \bibinfo{year}{2021}\natexlab{}.
\newblock \showarticletitle{Relaxed {Transformer} {Decoders} for {Direct}
  {Action} {Proposal} {Generation}}. In \bibinfo{booktitle}{\emph{{IEEE}/{CVF}
  {International} {Conference} on {Computer} {Vision}}}.
\newblock
\urldef\tempurl%
\url{https://doi.org/10.1109/iccv48922.2021.01327}
\showURL{%
\tempurl}


\bibitem[Van~Laerhoven et~al\mbox{.}(2022)]%
        {vanlaerhovenValidationOpensourceAmbulatory2022}
\bibfield{author}{\bibinfo{person}{Kristof Van~Laerhoven},
  \bibinfo{person}{Alexander Hoelzemann}, \bibinfo{person}{Iris Pahmeier},
  \bibinfo{person}{Andrea Teti}, {and} \bibinfo{person}{Lars Gabrys}.}
  \bibinfo{year}{2022}\natexlab{}.
\newblock \showarticletitle{Validation of an {Open}-{Source} {Ambulatory}
  {Assessment} {System} in {Support} of {Replicable} {Activity} {Studies}}.
\newblock \bibinfo{journal}{\emph{German Journal of Exercise and Sport
  Research}} \bibinfo{volume}{52}, \bibinfo{number}{2} (\bibinfo{year}{2022}),
  \bibinfo{pages}{262--272}.
\newblock
\urldef\tempurl%
\url{https://doi.org/10.1007/s12662-022-00813-2}
\showURL{%
\tempurl}


\bibitem[Vaswani et~al\mbox{.}(2017)]%
        {vaswaniAttentionAllYou2017}
\bibfield{author}{\bibinfo{person}{Ashish Vaswani}, \bibinfo{person}{Noam
  Shazeer}, \bibinfo{person}{Niki Parmar}, \bibinfo{person}{Jakob Uszkoreit},
  \bibinfo{person}{Llion Jones}, \bibinfo{person}{Aidan~N Gomez},
  \bibinfo{person}{Łukasz Kaiser}, {and} \bibinfo{person}{Illia Polosukhin}.}
  \bibinfo{year}{2017}\natexlab{}.
\newblock \showarticletitle{Attention {Is} {All} {You} {Need}}. In
  \bibinfo{booktitle}{\emph{Advances in {Neural} {Information} {Processing}
  {Systems}}}.
\newblock
\urldef\tempurl%
\url{https://arxiv.org/abs/1706.03762}
\showURL{%
\tempurl}


\bibitem[Wang et~al\mbox{.}(2021)]%
        {wangActionCLIPNewParadigm2021}
\bibfield{author}{\bibinfo{person}{Mengmeng Wang}, \bibinfo{person}{Jiazheng
  Xing}, {and} \bibinfo{person}{Yong Liu}.} \bibinfo{year}{2021}\natexlab{}.
\newblock \showarticletitle{{ActionCLIP}: {A} {New} {Paradigm} for {Video}
  {Action} {Recognition}}.
\newblock \bibinfo{journal}{\emph{CoRR}}  \bibinfo{volume}{abs/2109.08472}
  (\bibinfo{year}{2021}).
\newblock
\urldef\tempurl%
\url{https://arxiv.org/abs/2109.08472}
\showURL{%
\tempurl}


\bibitem[Ward et~al\mbox{.}(2006)]%
        {wardActivityRecognitionAssembly2006}
\bibfield{author}{\bibinfo{person}{Jamie~A. Ward}, \bibinfo{person}{Paul
  Lukowicz}, \bibinfo{person}{Gerhard Tröster}, {and} \bibinfo{person}{Thad~E.
  Starner}.} \bibinfo{year}{2006}\natexlab{}.
\newblock \showarticletitle{Activity {Recognition} of {Assembly} {Tasks}
  {Using} {Body}-{Worn} {Microphones} and {Accelerometers}}.
\newblock \bibinfo{journal}{\emph{IEEE Transactions on Pattern Analysis and
  Machine Intelligence}} (\bibinfo{year}{2006}), \bibinfo{pages}{1553--1567}.
\newblock
\urldef\tempurl%
\url{https://doi.org/10.1109/TPAMI.2006.197}
\showURL{%
\tempurl}


\bibitem[Wei and Kehtarnavaz(2020)]%
        {weiSimultaneousUtilizationInertial2020}
\bibfield{author}{\bibinfo{person}{Haoran Wei} {and} \bibinfo{person}{Nasser
  Kehtarnavaz}.} \bibinfo{year}{2020}\natexlab{}.
\newblock \showarticletitle{Simultaneous {Utilization} of {Inertial} and
  {Video} {Sensing} for {Action} {Detection} and {Recognition} in {Continuous}
  {Action} {Streams}}.
\newblock \bibinfo{journal}{\emph{IEEE Sensors Journal}} \bibinfo{volume}{20},
  \bibinfo{number}{11} (\bibinfo{year}{2020}), \bibinfo{pages}{6055--6063}.
\newblock
\urldef\tempurl%
\url{https://doi.org/10.1109/JSEN.2020.2973361}
\showURL{%
\tempurl}


\bibitem[Wu et~al\mbox{.}(2017)]%
        {wuAnticipatingDailyIntention2017}
\bibfield{author}{\bibinfo{person}{Tz-Ying Wu}, \bibinfo{person}{Ting-An
  Chien}, \bibinfo{person}{Cheng-Sheng Chan}, \bibinfo{person}{Chan-Wei Hu},
  {and} \bibinfo{person}{Min Sun}.} \bibinfo{year}{2017}\natexlab{}.
\newblock \showarticletitle{Anticipating {Daily} {Intention} {Using}
  {On}-{Wrist} {Motion} {Triggered} {Sensing}}. In
  \bibinfo{booktitle}{\emph{{IEEE} {International} {Conference} on {Computer}
  {Vision}}}.
\newblock
\urldef\tempurl%
\url{https://doi.org/´}
\showDOI{\tempurl}


\bibitem[Xi et~al\mbox{.}(2018)]%
        {xiDeepDilatedConvolution2018}
\bibfield{author}{\bibinfo{person}{Rui Xi}, \bibinfo{person}{Mengshu Hou},
  \bibinfo{person}{Mingsheng Fu}, \bibinfo{person}{Hong Qu}, {and}
  \bibinfo{person}{Daibo Liu}.} \bibinfo{year}{2018}\natexlab{}.
\newblock \showarticletitle{Deep {Dilated} {Convolution} on {Multimodality}
  {Time} {Series} for {Human} {Activity} {Recognition}}. In
  \bibinfo{booktitle}{\emph{{IEEE} {International} {Joint} {Conference} on
  {Neural} {Networks}}}.
\newblock
\urldef\tempurl%
\url{https://doi.org/10.1109/IJCNN.2018.8489540}
\showURL{%
\tempurl}


\bibitem[Xu et~al\mbox{.}(2019)]%
        {xuInnoHARDeepNeural2019}
\bibfield{author}{\bibinfo{person}{Cheng Xu}, \bibinfo{person}{Duo Chai},
  \bibinfo{person}{Jie He}, \bibinfo{person}{Xiaotong Zhang}, {and}
  \bibinfo{person}{Shihong Duan}.} \bibinfo{year}{2019}\natexlab{}.
\newblock \showarticletitle{{InnoHAR}: {A} {Deep} {Neural} {Network} for
  {Complex} {Human} {Activity} {Recognition}}.
\newblock \bibinfo{journal}{\emph{IEEE Access}}  \bibinfo{volume}{7}
  (\bibinfo{year}{2019}).
\newblock
\urldef\tempurl%
\url{https://doi.org/10.1109/ACCESS.2018.2890675}
\showURL{%
\tempurl}


\bibitem[Xu et~al\mbox{.}(2023)]%
        {xuContinualEgocentricActivity2023}
\bibfield{author}{\bibinfo{person}{Linfeng Xu}, \bibinfo{person}{Qingbo Wu},
  \bibinfo{person}{Lili Pan}, \bibinfo{person}{Fanman Meng},
  \bibinfo{person}{Hongliang Li}, \bibinfo{person}{Chiyuan He},
  \bibinfo{person}{Hanxin Wang}, \bibinfo{person}{Shaoxu Cheng}, {and}
  \bibinfo{person}{Yu Dai}.} \bibinfo{year}{2023}\natexlab{}.
\newblock \showarticletitle{Towards continual egocentric activity recognition:
  {A} multi-modal egocentric activity dataset for continual learning}.
\newblock \bibinfo{journal}{\emph{CoRR}}  \bibinfo{volume}{abs/2301.10931}
  (\bibinfo{year}{2023}).
\newblock
\urldef\tempurl%
\url{https://arxiv.org/abs/2301.10931}
\showURL{%
\tempurl}


\bibitem[Xu et~al\mbox{.}(2020)]%
        {xuGTADSubgraphLocalization2020}
\bibfield{author}{\bibinfo{person}{Mengmeng Xu}, \bibinfo{person}{Chen Zhao},
  \bibinfo{person}{David~S. Rojas}, \bibinfo{person}{Ali Thabet}, {and}
  \bibinfo{person}{Bernard Ghanem}.} \bibinfo{year}{2020}\natexlab{}.
\newblock \showarticletitle{G-{TAD}: {Sub}-{Graph} {Localization} for
  {Temporal} {Action} {Detection}}. In \bibinfo{booktitle}{\emph{{IEEE}/{CVF}
  {Conference} on {Computer} {Vision} and {Pattern} {Recognition}}}.
\newblock
\urldef\tempurl%
\url{https://doi.org/10.1109/cvpr42600.2020.01017}
\showURL{%
\tempurl}


\bibitem[Yang et~al\mbox{.}(2023)]%
        {yangBasicTADAstoundingRGBOnly2023}
\bibfield{author}{\bibinfo{person}{Min Yang}, \bibinfo{person}{Guo Chen},
  \bibinfo{person}{Yin-Dong Zheng}, \bibinfo{person}{Tong Lu}, {and}
  \bibinfo{person}{Limin Wang}.} \bibinfo{year}{2023}\natexlab{}.
\newblock \showarticletitle{{BasicTAD}: {An} {Astounding} {RGB}-{Only}
  {Baseline} for {Temporal} {Action} {Detection}}.
\newblock \bibinfo{journal}{\emph{Computer Vision and Image Understanding}}
  \bibinfo{volume}{232} (\bibinfo{year}{2023}).
\newblock
\urldef\tempurl%
\url{https://doi.org/10.1016/j.cviu.2023.103692}
\showURL{%
\tempurl}


\bibitem[Yu et~al\mbox{.}(2019)]%
        {yuHierarchicalDeepFusion2019}
\bibfield{author}{\bibinfo{person}{Haibin Yu}, \bibinfo{person}{Guoxiong Pan},
  \bibinfo{person}{Mian Pan}, \bibinfo{person}{Chong Li},
  \bibinfo{person}{Wenyan Jia}, \bibinfo{person}{Li Zhang}, {and}
  \bibinfo{person}{Mingui Sun}.} \bibinfo{year}{2019}\natexlab{}.
\newblock \showarticletitle{A {Hierarchical} {Deep} {Fusion} {Framework} for
  {Egocentric} {Activity} {Recognition} {Using} a {Wearable} {Hybrid} {Sensor}
  {System}}.
\newblock \bibinfo{journal}{\emph{MDPI Sensors}} \bibinfo{volume}{19},
  \bibinfo{number}{3} (\bibinfo{year}{2019}).
\newblock
\urldef\tempurl%
\url{https://doi.org/10.3390/s19030546}
\showURL{%
\tempurl}


\bibitem[Yuan et~al\mbox{.}(2021)]%
        {yuanTokenstotokenViTTraining2021}
\bibfield{author}{\bibinfo{person}{Li Yuan}, \bibinfo{person}{Yunpeng Chen},
  \bibinfo{person}{Tao Wang}, \bibinfo{person}{Weihao Yu},
  \bibinfo{person}{Yujun Shi}, \bibinfo{person}{Zi-Hang Jiang},
  \bibinfo{person}{Francis~E.H. Tay}, \bibinfo{person}{Jiashi Feng}, {and}
  \bibinfo{person}{Shuicheng Yan}.} \bibinfo{year}{2021}\natexlab{}.
\newblock \showarticletitle{Tokens-{To}-{Token} {ViT}: {Training} {Vision}
  {Transformers} {From} {Scratch} on {Imagenet}}. In
  \bibinfo{booktitle}{\emph{{IEEE}/{CVF} {International} {Conference} on
  {Computer} {Vision}}}.
\newblock
\urldef\tempurl%
\url{https://doi.org/10.1109/iccv48922.2021.00060}
\showURL{%
\tempurl}


\bibitem[Yuki et~al\mbox{.}(2018)]%
        {yukiActivityRecognitionUsing2018}
\bibfield{author}{\bibinfo{person}{Yuta Yuki}, \bibinfo{person}{Junto Nozaki},
  \bibinfo{person}{Kei Hiroi}, \bibinfo{person}{Katsuhiko Kaji}, {and}
  \bibinfo{person}{Nobuo Kawaguchi}.} \bibinfo{year}{2018}\natexlab{}.
\newblock \showarticletitle{Activity {Recognition} {Using} {Dual}-{Convlstm}
  {Extracting} {Local} and {Global} {Features} for {Shl} {Recognition}
  {Challenge}}. In \bibinfo{booktitle}{\emph{{ACM} {International} {Joint}
  {Conference} and {International} {Symposium} on {Pervasive} and {Ubiquitous}
  {Computing} and {Wearable} {Computers}}}.
\newblock
\urldef\tempurl%
\url{https://doi.org/10.1145/3267305.3267533}
\showURL{%
\tempurl}


\bibitem[Zeng et~al\mbox{.}(2019)]%
        {zengGraphConvolutionalNetworks2019}
\bibfield{author}{\bibinfo{person}{Runhao Zeng}, \bibinfo{person}{Wenbing
  Huang}, \bibinfo{person}{Mingkui Tan}, \bibinfo{person}{Yu Rong},
  \bibinfo{person}{Peilin Zhao}, \bibinfo{person}{Junzhou Huang}, {and}
  \bibinfo{person}{Chuang Gan}.} \bibinfo{year}{2019}\natexlab{}.
\newblock \showarticletitle{Graph {Convolutional} {Networks} for {Temporal}
  {Action} {Localization}}. In \bibinfo{booktitle}{\emph{{IEEE}/{CVF}
  {International} {Conference} on {Computer} {Vision}}}.
\newblock
\urldef\tempurl%
\url{https://doi.org/10.1109/iccv.2019.00719}
\showURL{%
\tempurl}


\bibitem[Zhang et~al\mbox{.}(2022)]%
        {zhangActionFormerLocalizingMoments2022}
\bibfield{author}{\bibinfo{person}{Chen-Lin Zhang}, \bibinfo{person}{Jianxin
  Wu}, {and} \bibinfo{person}{Yin Li}.} \bibinfo{year}{2022}\natexlab{}.
\newblock \showarticletitle{Actionformer: {Localizing} {Moments} of {Actions}
  {With} {Transformers}}. In \bibinfo{booktitle}{\emph{European {Conference} on
  {Computer} {Vision}}}.
\newblock
\urldef\tempurl%
\url{https://doi.org/10.1007/978-3-031-19772-7_29}
\showURL{%
\tempurl}


\bibitem[Zhao et~al\mbox{.}(2021)]%
        {zhaoVideoSelfstitchingGraph2021}
\bibfield{author}{\bibinfo{person}{Chen Zhao}, \bibinfo{person}{Ali~K. Thabet},
  {and} \bibinfo{person}{Bernard Ghanem}.} \bibinfo{year}{2021}\natexlab{}.
\newblock \showarticletitle{Video {Self}-{Stitching} {Graph} {Network} for
  {Temporal} {Action} {Localization}}. In
  \bibinfo{booktitle}{\emph{{IEEE}/{CVF} {International} {Conference} on
  {Computer} {Vision}}}.
\newblock
\urldef\tempurl%
\url{https://doi.org/10.1109/iccv48922.2021.01340}
\showURL{%
\tempurl}


\bibitem[Zhao et~al\mbox{.}(2020)]%
        {zhaoBottomupTemporalAction2020}
\bibfield{author}{\bibinfo{person}{Peisen Zhao}, \bibinfo{person}{Lingxi Xie},
  \bibinfo{person}{Chen Ju}, \bibinfo{person}{Ya Zhang},
  \bibinfo{person}{Yanfeng Wang}, {and} \bibinfo{person}{Qi Tian}.}
  \bibinfo{year}{2020}\natexlab{}.
\newblock \showarticletitle{Bottom-up {Temporal} {Action} {Localization} {With}
  {Mutual} {Regularization}}. In \bibinfo{booktitle}{\emph{European
  {Conference} on {Computer} {Vision}}}.
\newblock
\urldef\tempurl%
\url{https://doi.org/10.1007/978-3-030-58598-3_32}
\showURL{%
\tempurl}


\bibitem[Zhou et~al\mbox{.}(2022)]%
        {zhouTinyHARLightweightDeep2022}
\bibfield{author}{\bibinfo{person}{Yexu Zhou}, \bibinfo{person}{Haibin Zhao},
  \bibinfo{person}{Yiran Huang}, \bibinfo{person}{Till Riedel},
  \bibinfo{person}{Michael Hefenbrock}, {and} \bibinfo{person}{Michael Beigl}.}
  \bibinfo{year}{2022}\natexlab{}.
\newblock \showarticletitle{{TinyHAR}: {A} {Lightweight} {Deep} {Learning}
  {Model} {Designed} for {Human} {Activity} {Recognition}}. In
  \bibinfo{booktitle}{\emph{{ACM} {International} {Symposium} on {Wearable}
  {Computers}}}.
\newblock
\urldef\tempurl%
\url{https://doi.org/10.1145/3544794.3558467}
\showURL{%
\tempurl}


\bibitem[Zhu et~al\mbox{.}(2021)]%
        {zhuEnrichingLocalGlobal2021}
\bibfield{author}{\bibinfo{person}{Zixin Zhu}, \bibinfo{person}{Wei Tang},
  \bibinfo{person}{Le Wang}, \bibinfo{person}{Nanning Zheng}, {and}
  \bibinfo{person}{Gang Hua}.} \bibinfo{year}{2021}\natexlab{}.
\newblock \showarticletitle{Enriching {Local} and {Global} {Contexts} for
  {Temporal} {Action} {Localization}}. In
  \bibinfo{booktitle}{\emph{{IEEE}/{CVF} {International} {Conference} on
  {Computer} {Vision}}}.
\newblock
\urldef\tempurl%
\url{https://doi.org/10.1109/iccv48922.2021.01326}
\showURL{%
\tempurl}


\end{thebibliography}

\end{document}

\maketitle
\appendix

\section{Dataset Overview and Contents}

The outdoor sports dataset WEAR features data of 18 participants performing each a total of 18 different workout activities with untrimmed inertial (acceleration) and camera (egocentric video) data recorded at 10 different outside locations. It provides a challenging prediction scenario marked by purposely introduced activity variations and an overall small information overlap across modalities. Figure~\ref{fig:keyfacts} provides a dataset nutrition label inspired by \citet{hollandDatasetNutritionLabel2018} in a table-like manner. 

\begin{figure}[h]
\begin{center}
   \includegraphics[width=.55\columnwidth]{figures/key_facts.pdf}
\end{center}
   \caption{Dataset nutrition label of the WEAR dataset. The dataset nutrition label was originally proposed by \citet{hollandDatasetNutritionLabel2018}. Our adaptation is inspired by \citet{delpretoActionSenseMultimodalDataset2022}.}
\label{fig:keyfacts}
\end{figure}

\subsection{Intended Uses and Ethical Considerations}

Before participating in the study, participants were notified that by nature the data they provide can only be pseudonymised. This means that, though requiring a substantial amount of effort, the identity of a person can be reconstructed. Although participants agreed to include their egocentric videos in a public dataset, it is essential to refrain from actively identifying the individuals featured in the WEAR dataset. If other researchers decide to contribute to the WEAR dataset by recording additional participants, societal and ethical implications should be considered. As with the participants part of the original release of the WEAR dataset, all participants must be briefed before their first recording, making them aware of all necessary information and implications that come with providing to the WEAR dataset. Recording locations should only be chosen if video recordings are allowed at said location and participants are given enough space to perform each activity safely. If the recording location involves pedestrians walking within close proximity, pedestrians should be notified that they are being recorded and, if applicable, captured faces should be blurred during postprocessing.

The WEAR dataset and associated code are made public for research purposes. With the accurate detection of physical activities that we perform in our daily lives having been identified as valuable information, the WEAR dataset focuses on one of the most popular application scenarios of wearable smartwatches and action cameras, i.e. self-tracking of workout activities. With the ease of reproducability we hope to make WEAR a collaborative, expanding dataset which researchers from different locations and backgrounds can contribute to. For example, as the current selection of participants is biased towards healthy, young people, we hope to overcome said limitation by including people from more diverse backgrounds and age groups in future iterations of the dataset. 

Lastly, the authors took great care of avoiding any infringement of rights during the data
collection process. Yet, in case of conflicts, they are of course committed to taking appropriate actions, such as promptly removing data associated with such concerns.
%Lastly, the authors take full responsibility for any potential infringement of rights during the data collection process or any other related work. They are committed to taking appropriate actions, such as promptly removing data associated with such concerns.

\subsection{Data Availability and Licensing}

WEAR and all associated files are offered under a Creative Commons Attribution-NonCommercial-ShareAlike 4.0 International License. The dataset is hosted via the cloud-storage platform [Anonymized], which is a service hosted by [anonymized] (\url{https://www.anonymous.edu/anon}). It is a non-commercial cloud storage service for research, studying and teaching and is provided to participating institutions exclusively. With locations exclusively in [anonymized], [anonymized] is subject to the strict [anonymized] directives on data protection and data security. The complete dataset can be downloaded via [anonyimized] (\url{https://www.anonymous.edu/anon}). The dataset download is structured into the (1) '.json'-formatted annotations, (2) raw, synchronized inertial and vision data and (3) precomputed feature embeddings as mentioned in the main paper. Third party data-hosting services will be explored once the dataset paper is published and in a non-changing state. We will involve the ethics council of [anonymized] during our decision process to ensure a each selected hosting platform is inline with our data privacy standards.

The source code that was used to conduct all experiments is available via [Anonymized]~(\url{https://www.anonymous.edu/anon}). A snapshot of the code is provided as part of the supplementary material download. The repository is written in such a way that other architectures (both inertial- and vision-based) can be added in the future. The repository provides Readme files which give details on the overall structure of the repository, how collect additional data and how to set up an Anaconda environment with the needed packages to run experiments. Experiments are defined via '.json'-format configuration files which allow for easy sharing of used hyperparameter settings.

\section{Experimental Protocol}

Table~\ref{tab:classes} gives an overview of the 18 activity classes featured in the WEAR dataset and provides number of coherent sequences as well as total duration per workout activity class. In order to properly explain participants the activities they needed to perform and give insights on the overall study design a recording plan (see Section~\ref{sec:recordingplan}) was provided to participants prior to their first session. The recording plan details all necessary materials and is written in such a way that the can easily be reproduced by persons other than the authors. The plan further outlines the study protocol as well informs about any risks of harm, data collection, usage, anonymisation and publication, as well as how to revoke data usage rights at any point in the future. Besides a written description of each activity, the original document provides short video-clips of each activity, showing the correct execution of exercises. To avoid any misunderstandings, the participants further received a one-on-one session with the researchers being able to ask their questions about the plan and activities in it. Other than the used sensors for video and acceleration recording, the exercises only require a yoga mat and a chair (or similar items). Sessions can be recorded at any location outside as long as the privacy and safety of the participants as well as pedestrians is ensured.

\begin{table}
  \scriptsize
  \centering
  \caption{Overview of the activity classes featured in the WEAR dataset. Each activity is categorized into as either a running (R), flexibility~(F) or strength (S) exercise. The total duration of each class is provided in minutes and averaged across all activities. The total duration of the null class, i.e. samples not belonging to any of the classes of interest, is provided. A detailed description of each activity can be found in the recording plan attached at the end of the supplementary material.}
 \label{tab:classes}
  \begin{tabular}{@{}llccc@{}}
    Label ID & Activity Class & Category & Action Segments & Total duration \\
     & & & & (in min) \\ 
    \bottomrule
    1  & jogging                       & R & 28 & 32:30 \\
    2  & jogging (rotating arms)       & R & 22 & 29:34 \\
    3  & jogging (skipping)            & R & 34 & 29:11 \\
    4  & jogging (sidesteps)           & R & 30 & 33:33 \\
    5  & jogging (butt-kicks)          & R & 37 & 28:43 \\
    6  & stretching (triceps)          & F & 25 & 29:13 \\
    7  & stretching (lunging)          & F & 23 & 31:09 \\
    8  & stretching (shoulders)        & F & 22 & 30:04 \\
    9  & stretching (hamstrings)       & F & 23 & 30:50 \\
    10 & stretching (lumbar rotation)  & F & 27 & 31:36 \\
    12 & push-ups                      & S & 57 & 27:33 \\
    13 & push-ups (complex)            & S & 41 & 29:14 \\
    14 & sit-ups                       & S & 43 & 30:50 \\
    15 & sit-ups (complex)             & S & 32 & 31:02 \\
    16 & burpees                       & S & 49 & 31:25 \\
    17 & lunges                        & S & 31 & 31:54 \\
    18 & lunges (complex)              & S & 35 & 33:19 \\
    19 & bench-dips                    & S & 56 & 28:38 \\
    \bottomrule
    0 & null                           & - & 592 & 358:29 \\
  \end{tabular}
\end{table}

\subsection{Participant and Session Information}
\label{subsec:participantsessioninfo}
The location and the time of day at which the sessions were performed were not fixed and thus vary across subjects. As participants were allowed to split activities across more (or less) than two sessions, session counts vary across subjects. Table~\ref{tab:locations} provides information on all 10 recording locations that are part of the WEAR dataset. The table details general information such as surface conditions of the location as well as which direction the static camera seen in videos is facing. Table~\ref{tab:sessions} provides supplementary information on all separate sessions contained in the dataset. For each session, we detail its overall length in minutes, the number of distinct activities performed by the participant, the location it was recorded at, the month and time of day it was recorded, as well as the overall weather conditions during the duration of the session.

\begin{table}
  \centering
  \scriptsize
  \caption{Description of the 10 locations featured in the WEAR dataset. For each location we provide information on surface conditions, overall surroundings and direction the static camera is facing.}
  \label{tab:locations}
  \begin{tabular}{@{}ll@{}}
    Location ID & Description \\
    \toprule
    1       & \begin{tabular}[c]{@{}l@{}}Meadow in proximity to a larger building. \\Area is surrounded by trees with from November on-wards, fallen leaves laying on the ground. \\Static camera faces North-West. \end{tabular}  \\
    \midrule

    2        & \begin{tabular}[c]{@{}l@{}}Parking lot in proximity to building. Concrete surface.\\ Static camera faces West. \end{tabular}  \\
    \midrule
    3        & \begin{tabular}[c]{@{}l@{}} Small square with concrete surface. Surrounded by bushes and buildings.\\ Static camera faces West. \end{tabular} \\
    \midrule
    4        & \begin{tabular}[c]{@{}l@{}} Meadow enclosed by bungalow-style living quarters. \\ Static camera faces North-East \end{tabular} \\
    \midrule
    5        &  \begin{tabular}[c]{@{}l@{}} Covered walkway next to a building. Concrete surface. \\ Walkway enclosed by building and bushes. \\ Static camera faces North. \end{tabular} \\
    \midrule
    6        &  \begin{tabular}[c]{@{}l@{}} Football field with ash surface build behind a supermarket next a road and crop-fields. \\ Long side of the football field is surrounded by bushes. \\Static camera faces mostly North-West. \end{tabular} \\
    \midrule
    7        &  \begin{tabular}[c]{@{}l@{}} Backyard in an urban-village with both concrete and grass surface. \\ Terrace has a garden table and chairs standing around. \\ Static camera faces mostly West.\end{tabular} \\
    \midrule    
    8        &  \begin{tabular}[c]{@{}l@{}} Parking lot next to allotments in a city-area. \\ Static camera faces mostly North-East. \end{tabular} \\
    \midrule
    9        &  \begin{tabular}[c]{@{}l@{}}Meadow next to a building. \\ Static camera faces South. \end{tabular} \\
    \midrule
    10       &  \begin{tabular}[c]{@{}l@{}} City park in a metropolitan area behind a city mall. \\Park is surrounded by buildings, a playing ground, football and basketball fields. \\ Static camera faces mostly North. \end{tabular} \\
      \bottomrule

  \end{tabular}
\end{table}

\begin{table}
  \centering
  \scriptsize
  \caption{Per-session meta-information. We provide the individual session count, duration of each session, number of activities performed during the session, location ID (LID) the session was performed at, approximate time of the year and day and weather conditions during recording time. More detailed information on each location can be found in Table~\ref{tab:locations} using the location ID.}
  \label{tab:sessions}
  \begin{tabular}{@{}lllcllll@{}}
    Subject & Session & Duration & \# Activities & Month & Time-of-day & LID & Weather conditions \\
    \toprule
    sbj\_0        & 1 & 16:33:30 & 7  & mid-Oct. & morning & 1 & sunny, $\approx$10$^{\circ}$C \\
    sbj\_0        & 2 & 11:55:00 & 6  & mid-Oct. & afternoon & 1 & partly-cloudy, $\approx$10$^{\circ}$C \\
    sbj\_0        & 3 & 18:06:00 & 7  & late-Oct. & afternoon & 1 & partly-cloudy, $\approx$20$^{\circ}$C \\
    sbj\_1        & 1 & 20:20:00 & 9  & late-Oct. & afternoon & 1 & sunny, $\approx$15$^{\circ}$C \\
    sbj\_1        & 2 & 25:58:00 & 9  & early-Nov. & afternoon & 1 & sunny, $\approx$10$^{\circ}$C \\
    sbj\_2        & 1 & 32:24:00 & 9  & early-Nov. & morning & 1 & sunny, $\approx$10$^{\circ}$C \\
    sbj\_2        & 2 & 25:08:00 & 9  & mid-Jan. & afternoon & 2 & cloudy, after rain, $\approx$0$^{\circ}$C \\
    sbj\_2        & 3 & 01:52:00 & 1  & mid-Feb. & afternoon & 3 & sunny, $\approx$5$^{\circ}$C \\
    sbj\_3        & 1 & 33:34:00 & 10 & mid-Nov. & afternoon & 4 & sunny, $\approx$5$^{\circ}$C \\
    sbj\_3        & 2 & 25:52:00 & 6  & mid-Nov. & afternoon & 4 & partly-cloudy, $\approx$10$^{\circ}$C \\
    sbj\_3        & 3 & 06:24:00 & 2  & mid-Nov. & afternoon & 4 & sunny, $\approx$10$^{\circ}$C \\
    sbj\_3        & 4 & 03:41:00 & 2  & late-Jan. & afternoon & 5 & cloudy, snowy, $\approx$-5$^{\circ}$C \\
    sbj\_4        & 1 & 24:07:30 & 9  & mid-Nov. & midday & 1 & foggy, cloudy, windy, $\approx$5$^{\circ}$C\\
    sbj\_4        & 2 & 29:04:00 & 9  & late-Nov. & afternoon & 1 & partly-cloudy, $\approx$5$^{\circ}$C \\
    sbj\_5        & 1 & 19:48:30 & 9  & mid-Nov. & afternoon & 1 & sunny, $\approx$10$^{\circ}$C\\
    sbj\_5        & 2 & 16:02:00 & 9  & end-Nov. & afternoon & 1 & cloudy, $\approx$5$^{\circ}$C\\
    sbj\_6        & 1 & 23:52:00 & 10 & end-Nov. & afternoon & 1 & foggy, $\approx$5$^{\circ}$C\\
    sbj\_6        & 2 & 17:51:30 & 8  & end-Jan. & morning & 5 & cloudy, snowy, $\approx$-5$^{\circ}$C \\
    sbj\_7        & 1 & 22:48:00 & 9  & late-Dec. & morning & 6 & partly-sunny, $\approx$10$^{\circ}$C\\
    sbj\_7        & 2 & 24:45:00 & 9  & late-Dec. & midday & 6 & partly-sunny, $\approx$10$^{\circ}$C\\
    sbj\_8        & 1 & 20:00:00 & 9  & late-Dec. & midday & 6 & partly-cloudy, $\approx$10$^{\circ}$C\\
    sbj\_8        & 2 & 21:35:00 & 9  & late-Jan. & afternoon & 7 & cloudy, $\approx$0$^{\circ}$C\\
    sbj\_9        & 1 & 18:50:00 & 9  & early-Jan. & afternoon & 8 & cloudy, $\approx$10$^{\circ}$C\\
    sbj\_9        & 2 & 17:16:00 & 9  & early-Jan. & afternoon & 8 & cloudy, $\approx$10$^{\circ}$C\\
    sbj\_10       & 1 & 21:42:00 & 9  & mid-Jan. & afternoon & 5 & rainy, windy, $\approx$5$^{\circ}$C\\
    sbj\_10       & 2 & 21:04:00 & 9  & early-Feb. & afternoon & 5 & rainy, windy, $\approx$5$^{\circ}$C \\
    sbj\_10       & 3 & 23:39:00 & 9  & mid-Feb. & afternoon & 9, 3 & sunny, cloudy, windy, $\approx$5$^{\circ}$C \\
    sbj\_11       & 1 & 17:41:00 & 9  & mid-Jan. & morning & 5 & cloudy, rainy, $\approx$5$^{\circ}$C \\
    sbj\_11       & 2 & 19:21:00 & 9  & mid-Jan. & midday & 5 & cloudy, rainy, $\approx$5$^{\circ}$C\\
    sbj\_12       & 1 & 27:08:00 & 9  & mid-Jan. & afternoon & 5 & cloudy, windy, $\approx$0$^{\circ}$C\\
    sbj\_12       & 2 & 27:22:00 & 9  & late-Feb. & afternoon & 5 & partly-sunny, windy, $\approx$0$^{\circ}$C \\
    sbj\_13       & 1 & 30:08:00 & 9  & mid-Jan. & afternoon & 5, 3 & sunny, $\approx$0$^{\circ}$C\\
    sbj\_13       & 2 & 36:10:00 & 9  & mid-Jan. & afternoon & 5, 3 & sunny, $\approx$0$^{\circ}$C\\
    sbj\_14       & 1 & 22:18:00 & 9  & mid-Jan. & afternoon & 5, 3 & sunny, $\approx$-5$^{\circ}$C \\
    sbj\_14       & 2 & 31:03:00 & 9  & mid-Jan. & afternoon & 5, 3 & cloudy, $\approx$-5$^{\circ}$C\\
    sbj\_15       & 1 & 23:17:00 & 9  & late-Jan. & afternoon & 5, 3 & cloudy, $\approx$0$^{\circ}$C\\
    sbj\_15       & 2 & 20:06:00 & 9  & late-Jan. & afternoon & 5, 3 & cloudy, $\approx$0$^{\circ}$C\\
    sbj\_16       & 1 & 26:34:00 & 9  & early-Feb. & midday & 10 & partly-sunny, $\approx$10$^{\circ}$C\\
    sbj\_16       & 2 & 31:56:00 & 9  & early-Feb. & midday & 10 & partly-sunny, $\approx$10$^{\circ}$C\\
    sbj\_17       & 1 & 23:16:00 & 9  & early-Feb. & afternoon & 1 & sunny, $\approx$0$^{\circ}$C\\
    sbj\_17       & 2 & 28:15:00 & 9  & early-Feb. & afternoon & 3 & sunny, $\approx$0$^{\circ}$C\\
      
      \bottomrule

  \end{tabular}
\end{table}

After having completed all sessions, participants were asked to take part in a questionnaire which was used to gather vital information (gender, age, height and weight) as well as workout-specific questions, aiming towards assessing the overall fitness level and experience with the activities detailed in the study protocol. The workout-specific questions were:

\begin{enumerate}
    \item How many workouts (longer than 15 min) do you usually do per week?
    \item Which kind of workout do you usually do (cycling, team sport, gym, cardio, yoga etc.)?
    \item How many activities that are part of the workout plan did you know in advance? 
    \item How many activities that are part of the workout plan do you perform regularly yourself as part of your own workouts?
\end{enumerate}

Table~\ref{tab:questionnaire} shows the answers to the questionnaire items for each participant. Note that, to protect the privacy of our study participants, we only asked for age, height and weight in ranges instead of exact values, and always provided the option to not answer the questions if preferred.

\begin{table}
  \centering
  \scriptsize
  \caption{Per subject answers to the questionnaire handed to participants after having completed all sessions. The questionnaire collected vital information (gender (G), left- or righthanded (L/R), age, height and weight) as well as workout-specific questions, i.e. frequency and type of private workouts and number of activities, part of the WEAR dataset, which were known in advance and regularly conducted in private workouts.}
  \label{tab:questionnaire}
  \begin{tabular}{@{}lllllcccccc@{}}
    Subject & G & L/R & Age & Height & Weight & \multicolumn{2}{c}{Private Workouts} & & \multicolumn{2}{c}{Activities} \\ \cmidrule{7-8} \cmidrule{10-11} 
    &  &  &  &  &  & Frequency & Type & & Known & Regularly\\ 
    \toprule
    sbj\_0        & M & R & $\geq$40 & 180-189 & 70-79 & 5 & Cycling & & 5 & 0 \\
    sbj\_1        & M & R & 25-29 & 170-179 & 60-69 & 3 & Hiking & & 11 & 0 \\
    sbj\_2        & M & R & 25-29 & 180-189 & 80-89 & 5 & Gym, Cardio & & 18 & 9 \\
    sbj\_3        & M & R & 35-39 & 170-179 & 70-79 & 4-5 & Gym, Basketball, Cardio & & 18 & 9 \\
    sbj\_4        & M & R & 25-29 & 180-189 & 60-69 & 0 & Table-tennis & & 18 & 0 \\
    sbj\_5        & F & R & 30-34 & 160 -169 & N/A & 2-3 & Freeletics & & 16 & 9 \\
    sbj\_6        & F & R & 25-29 & 150-159 & 50-59 & 1 & Gym & & 9 & 0 \\
    sbj\_7        & M & R & 30-34 & 180-189 & 80-89 & 5 & Gym, Cardio & & 18 & 5 \\
    sbj\_8        & F & R & 25-29 & 170-179 & 60-69 & 2-3 & Volleyball, Yoga & & 15 & 7 \\
    sbj\_9        & F & R & 25-29 & 150-159 & 50-59 & 7 & Gym, Bicycling, Cardio, Ballet & & 18 & 7 \\
    sbj\_10       & F & R & 20-24 & 160-169 & 50-59 & 5 & Gym, Dancing, Yoga & & 15 & 7 \\
    sbj\_11       & F & R & 25-29 & 160-169 & 50-59 & 3 & Volleyball, Cardio, Yoga & & 18 & 11 \\
    sbj\_12       & F & R & 20-24 & 170-179 & 60-69 & 4 & Gym & & 17 & 8 \\
    sbj\_13       & M & R & 20-24 & $\geq$190 & 90-99 & 2 & Gym, Cardio & & 16 & 8 \\
    sbj\_14       & M & R & 30-34 & 170-179 & 80-89 & 0 & N/A & & 11 & 2 \\
    sbj\_15       & F & L & 25-29 & 180-189 & 60-69 & 8 & Rowing, Gym, Cycling, Cardio & & 18 & 9 \\
    sbj\_16       & M & R & 20-24 & 180-189 & 60-69 & 2-3 & Gym & & 15 & 3 \\
    sbj\_17       & M & R & 25-29 & 180-189 & 70-79 & 4 & Badminton, Bouldering, Hiking & & 15 & 5 \\
      \bottomrule

  \end{tabular}
\end{table}

\subsection{Hardware Overview}

In order to capture the accelerometer data, four open-source Bangle.js Version 1 smartwatches running a custom, open-source firmware \citep{vanlaerhovenValidationOpensourceAmbulatory2022} were used. The Bangle.js Version 1 comes with a Nordic 64MHz nRF52832 ARM Cortex-M4 processor with Bluetooth LE, 64kB RAM, 512kB on-chip flash, 4MB external flash, a heart rate monitor, a 3D accelerometer and a 3D magnetometer. The raw 3D acceleration was captured at 50 Hz with a sensitivity of $\pm$ 8g. As outlined in the recording plan (see Section~\ref{sec:recordingplan}), watches were placed by the researchers before each session in a predetermined orientation on the participants' limbs and ankles. Egocentric video data was captured using a GoPro Hero 8 action camera. The camera was mounted using a headstrap with the camera tilted downwards in a roughly 45 degree angle. The GoPro was set to record at 1080p with 60 frames using a \emph{SuperView} FOV with \emph{Hypersmooth 2.0} electronic image stabilization and \emph{Auto Low-Light} correction turned on. As the recorded egocentric video of participants makes accurate ground truth annotations more difficult (due to e.g. participants not looking at the actions they perform), a second camera was placed on a tripod in the proximity to the participants. Using again a large FOV setting, the second camera was placed in a way such that as much area as possible was captured. To allow for even more freedom of movement, participants were allowed to move out of the FOV of the second camera, but were asked to start and end their activities within the camera's FOV. This allowed participants, especially during running exercises, to run straight distances and overall commence activities in a more natural way. To preserve the privacy of our participants, the second camera's video stream and all audio streams captured during the experiments are not part of the WEAR dataset.

\subsection{Postprocessing and Annotation Process}

The open-source firmware \citep{vanlaerhovenValidationOpensourceAmbulatory2022} running on each Bangle.js smartwatch stores the lossless, delta-compressed inertial data in separate files on the internal memory of each watch. During post-processing, said compressed files were extracted, uncompressed and concatenated to a single CSV file per session. Being a common issue with accelerometers sampling at a high sampling rate, the Bangle.js smartwatch is not able to maintain an exact sampling rate of 50 Hz, with the true sampling rate being closer to 48 Hz with fluctuations ranging between $\pm$ 1 Hz. The firmware \citep{vanlaerhovenValidationOpensourceAmbulatory2022} provides for each file a timestamp that was set by the on-board real-time clock, which allows correcting individual times of all delta-compressed samples. Therefore, in order to obtain the true sampling rate and correct the timestamps of the concatenated CSV-file, synchronisation jumps were performed by each participant at the start and end of each session. The synchronization jumps involved participants move in front of the tripod-mounted camera, stand still for approximately 10 seconds, jump three times while raising the arms while jumping and stand still for another 10 seconds. This allowed to map peaks in the inertial sensor streams to be mapped to points in the video stream and thus obtain a start and end point within both modality data streams. Lastly, assuming recorded inertial data records are equidistant, all records within the span of the start and end-point were evenly distributed across the experiment's duration and, as a final step, resampled to have a sampling rate of 50 Hz via linear interpolation. Similar to the inertial data, the video data recorded by the head-mounted GoPro was not recording at a true frame rate of 60 FPS, but slightly deviated from that (i.e. 59.94 FPS). We therefore also resampled the egocentric videos to be of a frame rate of 60 FPS. 

\begin{figure}
\begin{center}
   \includegraphics[width=1\linewidth]{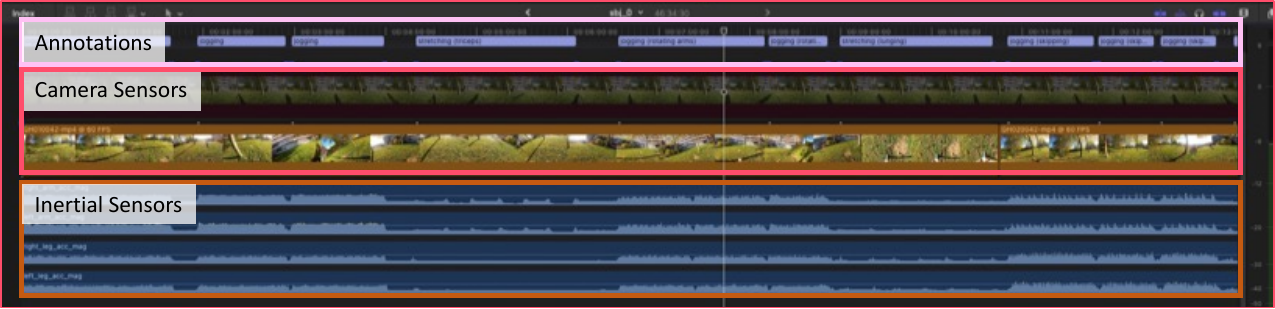}
\end{center}
   \caption{Snapshot along with descriptions of the annotation process using Final Cut Pro. Importing the converted video and inertial data (as '.wav'-files) allowed for an easy validation of the synchronization process. Labels were added via subtitles, exported as '.srt'-files and converted such that they can be appended to the respective '.csv'-files.}
\label{fig:finalcut}
\end{figure}

In order to validate our synchronization process we made use of the similarities between sensor and audio data and converted each axis of the 3D accelerometer as well as their combined magnitude to four separate WAV-files. This approach is inspired by the works of \citet{schollMultimediaExchangeFormat2019} and \citet{morshedPersonalizedApproachDeveloping2022}. We calculated the magnitude as the summed norm of each individual inertial sensor channels, i.e. $\sqrt{x^2 + y^2 + z^2}$ with $x$, $y$ and $z$ being the x-, y- and z-axis of the 3D accelerometer data. Having converted the CSV data to WAV files allowed us to import both video data and inertial data into a standard video editing tool, in our case we used Final Cut Pro (see Figure~\ref{fig:finalcut}). The user interface of Final Cut offers to see previews of sound files being in our case equivalent to a graph-like visualization of the acceleration data. This feature enabled us to have a visualised data stream of all modalities simultaneously while annotating. On average, the combined magnitude proved to be most useful when verifying the correctness of our synchronization across time. Labels of the activities were added by a single expert annotator as subtitles in SRT-format. A final script then converted the exported SRT-file to CSV-format, filling gaps within the subtitles with a \textit{NULL} label and appended this to the respective final inertial sensor data CSV-file.

\section{Supplementary Experiments and Figures}

\subsection{Attend-and-Discriminate Improvements}

Instead of employing a plain Attend-and-Discriminate model as proposed by \citet{abedinAttendDiscriminateStateoftheart2021}, we incorporate architecture improvements suggested by \citet{bockImprovingDeepLearning2021}. Said architecture improvements are (1) using one instead of two recurrent layers, (2) increasing the amount of hidden units in the recurrent layer from 128 to 1024 and (3) scaling the convolutional kernel by the same factor the window size increases or decreases. Table~\ref{tab:aanddmods} shows performance difference gained from employing the improved Attend-and-Discriminate architecture by comparing it to the original architecture. Note that results were obtained using longer training times along with a learning rate schedule (see Section~\ref{sec:traintime} for more details) and are reported without having applied any postprocessing.

\begin{table}
  \centering
  \small
  \caption{Results demonstrating the effectiveness of made modifications to the Attend-and-Discriminate model \citep{abedinAttendDiscriminateStateoftheart2021}. We compare the plain original model with an optimised version (1-layered LSTM with 1024 hidden units and an adjusted convolutional kernel sizes). We report results on the three employed window sizes (0.5, 1.0 and 2.0 seconds) each with a 50\% overlap. Note that results are reported with no postprocessing applied.}
  \label{tab:aanddmods}
  \begin{tabular}{@{}llccccccccc@{}}
    & Model  & P & R & F1 & \multicolumn{6}{c}{mAP} \\ \cmidrule{6-11} 
    & & & & & 0.3 & 0.4 & 0.5 & 0.6 & 0.7 & Avg \\
    \toprule
    \multirow{2}{*}{\rotatebox[origin=c]{90}{0.5s}}
    & Original A-and-D  & 71.87 & 72.78 & 71.63 & 1.86 & 1.54 & 1.35 & 1.07 & 0.87 & 1.34 \\
    & Optimised A-and-D & 76.29 & 69.08 & 71.60 & 1.69 & 1.14 & 0.83 & 0.63 & 0.48 & 0.96 \\
    \toprule
    \multirow{2}{*}{\rotatebox[origin=c]{90}{1.0s}}
    & Original A-and-D  & 72.37 & 72.38 & 71.60 & 3.07 & 2.46 & 1.96 & 1.49 & 1.31 & 2.06 \\
    & Optimised A-and-D & 78.90 & 73.25 & 75.22 & 4.35 & 3.38 & 2.76 & 2.22 & 1.76 & 2.90 \\
    \toprule
    \multirow{2}{*}{\rotatebox[origin=c]{90}{2.0s}}
    & Original A-and-D  & 74.48 & 73.99 & 73.26 & 8.75 & 7.1 & 5.94 & 4.85 & 3.95 & 6.12 \\
    & Optimised A-and-D & 81.13 & 76.47 & 77.90 & 11.13 & 9.35 & 7.42 & 6.04 & 5.17 & 7.82 \\
      \bottomrule

  \end{tabular}
\end{table}

\subsection{Longer vs. Shorter Training Runs}
\label{sec:traintime}
As mentioned in the main paper, all inertial-based architectures are trained for 300 epochs as compared to 30 epochs reported in \cite{bockImprovingDeepLearning2021}. These longer training times are inspired by the training reported in \citet{abedinAttendDiscriminateStateoftheart2021}. To compensate for longer training times we employ a step-wise learning rate schedule as seen in \citet{abedinAttendDiscriminateStateoftheart2021} with a step size of 10 epochs and a decay rate of 0.9. Table~\ref{tab:longvsshort} shows the improvement gained from using such longer training times by comparing it to a shorter training time of 30 epochs.

\begin{table}
  \centering
  \small
  \caption{Results demonstrating the effectiveness of longer training times on the inertial-based models. Compared are the shallow DeepConvLSTM \citep{bockImprovingDeepLearning2021} and improved Attend-and-Discriminate \citep{abedinAttendDiscriminateStateoftheart2021} model using either a short training time (30 epochs and no step-wise learning rate schedule (LRS)) or long training time (300 epochs and LRS). We report results on the three employed window sizes (0.5, 1.0 and 2.0 seconds) each with a 50\% overlap. Note that results are reported with no postprocessing applied.}
  \label{tab:longvsshort}
  \begin{tabular}{@{}llllccccccccc@{}}
    & Model & Epochs & LRS  & P & R & F1 & \multicolumn{6}{c}{mAP} \\ \cmidrule{8-13} 
    & & & & & & & 0.3 & 0.4 & 0.5 & 0.6 & 0.7 & Avg \\
    \toprule
    \multirow{4}{*}{\scriptsize\rotatebox[origin=c]{90}{0.5s}}
    & Shallow D.                     & 30  &              & 70.51 & 72.92 & 70.71 & 2.13 & 1.82 & 1.55 & 1.33 & 1.22 & 1.61 \\
    & Shallow D.                     & 300 & \checkmark   & 77.29 & 69.13 & 71.91 & 2.50 & 1.97 & 1.65 & 1.37 & 1.16 & 1.73 \\
    & A-and-D                        & 30  &              & 72.15 & 71.87 & 71.24 & 1.97 & 1.61 & 1.33 & 1.04 & 0.84 & 1.36 \\
    & A-and-D                        & 300 &\checkmark    & 76.29 & 69.08 & 71.60 & 1.69 & 1.14 & 0.83 & 0.63 & 0.48 & 0.96 \\
    \midrule
    \multirow{4}{*}{\scriptsize\rotatebox[origin=c]{90}{1.0s}}
    & Shallow D.                     & 30  &              & 73.35 & 76.25 & 73.78 & 4.83 & 4.01 & 3.38 & 2.81 & 2.32 & 3.47 \\
    & Shallow D.                     & 300 & \checkmark   & 81.09 & 72.07 & 75.29 & 5.71 & 4.50 & 3.66 & 2.77 & 2.50 & 3.83 \\
    & A-and-D                        & 30  &              & 74.00 & 74.96 & 73.70 & 4.48 & 3.47 & 3.00 & 2.35 & 2.01 & 3.06 \\
    & A-and-D                        & 300 & \checkmark   & 78.90 & 73.25 & 75.22 & 4.35 & 3.38 & 2.76 & 2.22 & 1.76 & 2.90 \\
    \midrule
    \multirow{4}{*}{\scriptsize\rotatebox[origin=c]{90}{2.0s}}
    & Shallow D.                    & 30  &               & 74.97 & 78.21 & 75.63 & 11.68 & 10.44 & 8.71 & 7.8 & 6.42 & 9.01 \\
    & Shallow D.                    & 300 & \checkmark    & 82.95 & 74.63 & 77.60 & 13.24 & 11.1 & 8.79 & 7.77 & 6.77 & 9.53 \\
    & A-and-D                       & 30  &               & 77.04 & 79.01 & 77.29 & 10.55 & 8.74 & 7.21 & 6.17 & 5.08 & 7.55 \\
    & A-and-D                       & 300 & \checkmark    & 81.13 & 76.47 & 77.90 & 11.13 & 9.35 & 7.42 & 6.04 & 5.17 & 7.82 \\
      \bottomrule

  \end{tabular}
\end{table}

\subsection{Ablation Study on Postprocessing}

The following details ablation experiments conducted to demonstrate the effectiveness and validity of the applied postprocessing described in the experiments section of the main paper.

Figure~\ref{fig:postinertial} illustrates the effect the majority vote filter has on the prediction stream of the inertial-based models. One can see that without applying a majority vote filter, inertial-based architectures produce a large amount of non-coherent segments. This is due to the fact that during training, inertial models such as \citet{bockImprovingDeepLearning2021} and \citet{abedinAttendDiscriminateStateoftheart2021} are not explicitly trained to predict coherent segments, but rather predict a continuous stream of windowed data. The models therefore tend to show a lot of intermediate switches in-between activity labels which causes mAP scores of inertial-based architectures to be substantially lower than scores of vision-based models. We therefore make use of a majority vote filter to erase short activity-label switches. Table~\ref{tab:deepconvpostprocessing} and \ref{tab:aanddpostprocessing} shows experimental results of applying different-sized majority vote filters (5, 10, 15, 20 and 25 seconds) compared to applying no filter. Interestingly, results (see Table~\ref{tab:deepconvpostprocessing} and \ref{tab:aanddpostprocessing}) not only demonstrate the effectiveness of the majority vote filter through a substantial increase in mAP scores, yet also show that said increase does not come at the cost of a decreased F1-score, but rather an increase. Table~\ref{tab:deepconvpostprocessing} and \ref{tab:aanddpostprocessing} further show a majority vote filter of 15 seconds being most effective resulting in the highest F1-score.

\begin{figure}
\begin{center}
   \includegraphics[width=1\linewidth]{figures/post_inertial.pdf}
\end{center}
   \caption{Color-coded comparison of the ground truth data (top row) with the raw and postprocessed (15 sec majority vote filter) activity streams of the shallow DeepConvLSTM \citep{bockImprovingDeepLearning2021} and Attend-and-Discriminate \citep{abedinAttendDiscriminateStateoftheart2021} (A-and-D) model. The illustrated activity stream is of a sample subject having trained using inertial data which is windowed using a 1 second sliding window with 50\% overlap.}
\label{fig:postinertial}
\end{figure}

\begin{table}
  \centering
  \small
  \caption{Ablation experiments on the effect of different-sized majority vote (MV) filters (5, 10, 15, 20 and 25 seconds) on the raw prediction results (0 seconds) of the shallow DeepConvLSTM model \citep{bockImprovingDeepLearning2021}. We report results on the three employed window sizes (0.5, 1.0 and 2.0 seconds) each with a 50\% overlap. Best results per clip-length are in \textbf{bold}.}
  \label{tab:deepconvpostprocessing}
  \begin{tabular}{@{}lllcccccccc@{}}
    & MV filter & P & R & F1 & \multicolumn{6}{c}{mAP} \\ \cmidrule{6-11} 
    & & & & & 0.3 & 0.4 & 0.5 & 0.6 & 0.7 & Avg \\
    \toprule
    \multirow{5}{*}{\rotatebox[origin=c]{90}{0.5 window}}
    &  0 sec      & 77.29 & 69.13 & 71.91 &  2.50 &  1.97 &  1.65 &  1.37 &  1.16 &  1.73 \\
    &  5 sec      & 85.87 & 75.63 & 79.04 & 37.83 & 36.02 & 34.30 & 32.48 & 29.89 & 34.10 \\
    & 10 sec      & 86.63 & \textbf{75.81} & \textbf{79.38} & 50.43 & 47.92 & 46.17 & 43.86 & 41.94 & 46.06 \\
    & 15 sec      & \textbf{86.77} & 75.42 & 79.18 & 54.36 & 51.67 & 49.42 & 47.40 & 44.70 & 49.51 \\
    & 20 sec      & 86.72 & 74.71 & 78.66 & 56.90 & 53.97 & 51.65 & 49.06 & \textbf{46.25} & 51.57 \\
    & 25 sec      & 86.56 & 73.89 & 78.02 & \textbf{57.22} & \textbf{54.16} & \textbf{52.20} & \textbf{49.56} & 45.89 & \textbf{51.81} \\
    \toprule
    \multirow{5}{*}{\rotatebox[origin=c]{90}{1.0 window}}
    &  0 sec      & 81.09 & 72.07 & 75.29 &  5.71 &  4.50 &  3.66 &  2.77 &  2.50 &  3.83 \\
    &  5 sec      & 87.27 & 77.21 & 80.73 & 42.92 & 40.25 & 38.10 & 35.38 & 32.51 & 37.83 \\
    & 10 sec      & 87.87 & \textbf{77.35} & \textbf{81.05} & 52.02 & 49.72 & 47.21 & 44.42 & 41.94 & 47.06 \\
    & 15 sec      & \textbf{88.02} & 77.03 & 80.86 & 57.09 & 55.32 & 53.61 & 50.59 & 47.85 & 52.89 \\
    & 20 sec      & 87.98 & 76.44 & 80.44 & 59.24 & 57.28 & 55.49 & 52.17 & 50.07 & 54.85 \\
    & 25 sec      & 87.74 & 75.81 & 79.93 & \textbf{61.50} & \textbf{59.63} & \textbf{57.41} & \textbf{53.88} & \textbf{51.13} & \textbf{56.71} \\
    \toprule
    \multirow{5}{*}{\rotatebox[origin=c]{90}{2.0 window}}
    &  0 sec      & 82.95 & 74.63 & 77.60 & 13.24 & 11.10 &  8.79 &  7.77 &  6.77 &  9.53 \\
    &  5 sec      & 86.92 & 77.88 & 81.08 & 42.44 & 40.53 & 37.92 & 35.18 & 32.81 & 37.78 \\
    & 10 sec      & 87.80 & \textbf{78.37} & \textbf{81.71} & 55.51 & 52.62 & 49.75 & 47.09 & 44.87 & 49.97 \\
    & 15 sec      & \textbf{87.92} & 78.16 & 81.60 & 59.89 & 57.00 & 54.69 & 51.77 & 48.99 & 54.47 \\
    & 20 sec      & 87.90 & 77.74 & 81.32 & 61.04 & 58.99 & 57.05 & 53.31 & 50.49 & 56.18 \\
    & 25 sec      & 87.70 & 77.22 & 80.89 & \textbf{62.35} & \textbf{60.18} & \textbf{57.95} & \textbf{54.64} & \textbf{50.97} & \textbf{57.22} \\
      \bottomrule

  \end{tabular}
\end{table}

\begin{table}
  \centering
  \small
  \caption{Ablation experiments on the effect of different-sized majority vote (MV) filters (5, 10, 15, 20 and 25 seconds) on the raw prediction results (0 seconds) of the improved Attend-and-Discriminate model \citep{abedinAttendDiscriminateStateoftheart2021}. We report results on the three employed window sizes (0.5, 1.0 and 2.0 seconds) each with a 50\% overlap. Best results per clip length are in \textbf{bold}.}
  \label{tab:aanddpostprocessing}
  \begin{tabular}{@{}llccccccccc@{}}
    & MV filter & P & R & F1 & \multicolumn{6}{c}{mAP} \\ \cmidrule{6-11} 
    & & & & & 0.3 & 0.4 & 0.5 & 0.6 & 0.7 & Avg \\
    \toprule
    \multirow{6}{*}{\rotatebox[origin=c]{90}{0.5 window}}
    &  0 sec      & 76.29 & 69.08 & 71.60 &  1.69 &  1.14 &  0.83 &  0.63 &  0.48 &  0.96 \\
    &  5 sec      & 86.18 & 76.16 & 79.40 & 36.38 & 34.07 & 31.09 & 28.14 & 26.25 & 31.19 \\
    & 10 sec      & 87.25 & \textbf{76.31} & \textbf{79.78} & 49.15 & 46.28 & 43.86 & 41.41 & 39.61 & 44.06 \\
    & 15 sec      & 87.54 & 75.98 & 79.59 & 53.57 & 51.08 & 48.51 & 45.82 & 42.87 & 48.37 \\
    & 20 sec      & \textbf{87.61} & 75.38 & 79.20 & 56.13 & 53.42 & 50.90 & 47.50 & 44.76 & 50.54 \\
    & 25 sec      & 87.40 & 74.53 & 78.52 & \textbf{59.00} & \textbf{55.82} & \textbf{53.45} & \textbf{49.49} & \textbf{45.90} & \textbf{52.73} \\
    \toprule
    \multirow{6}{*}{\rotatebox[origin=c]{90}{1.0 window}}
    &  0 sec      & 78.90 & 73.25 & 75.22 &  4.35 &  3.38 &  2.76 &  2.22 &  1.76 &  2.90 \\
    &  5 sec      & 86.58 & 78.95 & 81.56 & 40.43 & 37.88 & 35.03 & 32.22 & 29.61 & 35.03 \\
    & 10 sec      & 87.61 & \textbf{79.24} & \textbf{82.09} & 51.81 & 49.59 & 47.73 & 44.88 & 41.55 & 47.11 \\
    & 15 sec      & 87.87 & 79.02 & 82.01 & 56.38 & 54.47 & 52.28 & 50.07 & 46.92 & 52.03 \\
    & 20 sec      & \textbf{87.94} & 78.59 & 81.74 & 57.80 & 57.80 & 55.88 & 52.95 & 49.58 & 55.19 \\
    & 25 sec      & 87.82 & 77.92 & 81.23 & \textbf{61.65 }& \textbf{59.71} & \textbf{58.10} & \textbf{54.85} & \textbf{51.44} & \textbf{57.15} \\
    \toprule
    \multirow{6}{*}{\rotatebox[origin=c]{90}{2.0 window}}
    &  0 sec      & 81.13 & 76.47 & 77.90 & 11.13 &  9.35 &  7.42 &  6.04 &  5.17 &  7.82 \\
    &  5 sec      & 86.57 & 80.10 & 82.22 & 38.81 & 36.58 & 33.69 & 31.05 & 28.98 & 33.82 \\
    & 10 sec      & 87.91 & \textbf{80.71} & 83.06 & 52.89 & 50.98 & 48.32 & 45.34 & 42.49 & 48.00 \\
    & 15 sec      & 88.24 & 80.55 & \textbf{83.08} & 58.32 & 56.68 & 54.44 & 51.58 & 48.34 & 53.87 \\
    & 20 sec      & \textbf{88.37} & 80.22 & 82.89 & 61.18 & 59.97 & 57.99 & 54.69 & 51.07 & 56.98 \\
    & 25 sec      & 88.24 & 79.76 & 82.51 & \textbf{62.83} & \textbf{61.06} & \textbf{58.96} & \textbf{56.20} & \textbf{52.87} & \textbf{58.38} \\
        \bottomrule

  \end{tabular}
\end{table}

Temporal action localization models such as the ActionFormer \citep{zhangActionFormerLocalizingMoments2022} and TriDet architecture \citep{shiTriDetTemporalAction2023} are not trained on an explicitly modelled NULL-class. This means, that unlike \cite{bockImprovingDeepLearning2021} and \citet{abedinAttendDiscriminateStateoftheart2021}, models are only able to predict segments with activity labels other than the NULL-class. With both models being set to predict up to 2000 action segments per video, the unprocessed prediction results resulted in activity streams such as illustrated in Figure~\ref{fig:postvision}. One can see that almost all samples have been assigned an activity label, leaving only a few data to be predicted as NULL, ultimately resulting in a substantially lower NULL-class accuracy than compared to inertial-based models mentioned in this paper. We therefore increased the score-threshold of both the ActionFormer and TriDet model, eliminating low-scoring segments and replacing them with NULL (see Figure~\ref{fig:postvision}). This improved classification performance of the ActionFormer (see Table~\ref{tab:actionformerpostprocessing}) and TriDet model (see Table~\ref{tab:tridetpostprocessing}) significantly across all experiments (i.e. using inertial, vision and a combined setup as input data), while only marginally decreasing mAP scores. Table~\ref{tab:tridetpostprocessing} further shows 0.2 being the most effective identified threshold of our ablation study, resulting in the highest F1-score of the temporal action localization models.

\begin{figure}
\begin{center}
   \includegraphics[width=0.9\linewidth]{figures/post_video.pdf}
\end{center}
   \caption{Color-coded comparison of the ground truth data (top row) with the raw and score-thresholded (0.2) activity streams of the TriDet \citep{shiTriDetTemporalAction2023} model. The illustrated activity stream is of sample subject having trained the model using both inertial and vision data which is windowed using a 1 second sliding window with 50\% overlap.}
\label{fig:postvision}
\end{figure}

\begin{table}
  \centering
  \scriptsize
  \caption{ActionFormer score thresholding results \citep{zhangActionFormerLocalizingMoments2022} ablation experiments on the effect of different score thresholds (0.05, 0.1, 0.15, 0.2 and 0.25) on the raw prediction results (0.0 threshold) of experiments involving the ActionFormer model. We report results on the ActionFormer being applied to only inertial, camera and a combined (inertial + camera) features using three clip length window sizes (0.5, 1.0 and 2.0 seconds) each with a 50\% overlap. Best results per modality are in \textbf{bold}.}
  \label{tab:actionformerpostprocessing}
  \begin{tabular}{@{}llllccccccccc@{}}
    & Threshold & CL & P & R & F1 & \multicolumn{6}{c}{mAP} \\ \cmidrule{7-12} 
    & & & & & & 0.3 & 0.4 & 0.5 & 0.6 & 0.7 & Avg \\
    \toprule
    \multirow{18}{*}{\rotatebox[origin=c]{90}{Inertial}}
    &  0.0       & 0.5s        & 55.65 & 77.50 & 61.41 & 73.55 & 70.70 & 62.51 & 48.14 & 32.02 & 57.38 \\
    &  0.05      & 0.5s        & 65.11 & 78.53 & 67.78 & 72.61 & 69.73 & 61.60 & 47.24 & 31.25 & 56.49 \\
    &  0.1       & 0.5s        & 71.15 & 77.77 & 72.15 & 70.28 & 67.59 & 59.74 & 45.25 & 29.67 & 54.51 \\
    &  0.15      & 0.5s        & 76.27 & 75.18 & 73.96 & 67.45 & 64.89 & 57.02 & 42.55 & 28.10 & 52.00 \\
    &  0.2       & 0.5s        & 78.73 & 70.50 & 72.51 & 63.71 & 61.28 & 53.90 & 39.81 & 26.40 & 49.02 \\
    &  0.25      & 0.5s        & 81.76 & 64.25 & 69.46 & 59.09 & 56.93 & 49.66 & 36.17 & 24.36 & 45.24 \\ \cmidrule{2-12}
    &  0.0       & 1.0s        & 58.46 & 78.94 & 61.91 & \textbf{80.02} & \textbf{78.14} & \textbf{74.28} & \textbf{69.19} & \textbf{61.32} & \textbf{72.59} \\
    &  0.05      & 1.0s        & 67.40 & \textbf{80.21} & 70.60 & 79.24 & 77.40 & 73.55 & 68.45 & 60.59 & 71.85 \\
    &  0.1       & 1.0s        & 74.00 & 79.14 & 74.72 & 77.63 & 75.80 & 72.05 & 67.14 & 59.34 & 70.39 \\
    &  0.15      & 1.0s        & 78.82 & 77.21 & 76.41 & 75.15 & 73.46 & 70.03 & 65.55 & 57.90 & 68.42 \\
    &  0.2       & 1.0s        & 81.69 & 75.37 & \textbf{76.86} & 72.90 & 71.30 & 68.28 & 64.14 & 56.65 & 66.65 \\
    &  0.25      & 1.0s        & \textbf{84.12} & 73.38 & \textbf{76.86} & 70.25 & 69.01 & 66.15 & 62.49 & 55.35 & 64.65 \\ \cmidrule{2-12} 
    &  0.0       & 2.0s        & 54.47 & 74.61 & 57.84 & 74.85 & 71.16 & 67.88 & 63.67 & 56.53 & 66.82 \\
    &  0.05      & 2.0s        & 61.67 & 75.41 & 64.98 & 73.92 & 70.13 & 66.81 & 62.62 & 55.69 & 65.84 \\
    &  0.1       & 2.0s        & 68.67 & 73.72 & 68.95 & 71.70 & 68.20 & 65.00 & 61.01 & 54.27 & 64.04 \\
    &  0.15      & 2.0s        & 74.59 & 71.78 & 71.00 & 69.22 & 66.05 & 63.02 & 59.22 & 52.51 & 62.00 \\
    &  0.2       & 2.0s        & 78.18 & 69.15 & 71.15 & 66.43 & 63.30 & 60.47 & 56.66 & 50.26 & 59.43 \\
    &  0.25      & 2.0s        & 80.99 & 66.93 & 70.90 & 64.11 & 61.06 & 58.31 & 54.73 & 48.47 & 57.34 \\
    \midrule
    \multirow{18}{*}{\rotatebox[origin=c]{90}{Camera}}
    &  0.0       & 0.5s            & 49.81 & 70.46 & 54.24 & 67.44 & 65.10 & 59.96 & 47.89 & 31.61 & 54.40 \\
    &  0.05      & 0.5s            & 60.74 & 71.84 & 61.94 & 65.69 & 63.30 & 58.37 & 46.42 & 30.48 & 52.85 \\
    &  0.1       & 0.5s            & 64.66 & 69.02 & 63.90 & 61.28 & 59.04 & 54.71 & 43.57 & 28.40 & 49.40 \\
    &  0.15      & 0.5s            & 66.39 & 63.73 & 62.19 & 56.29 & 54.18 & 50.10 & 39.76 & 25.88 & 45.24 \\
    &  0.2       & 0.5s            & 68.06 & 57.68 & 58.47 & 51.27 & 49.45 & 45.74 & 36.10 & 23.38 & 41.19 \\
    &  0.25      & 0.5s            & 66.90 & 51.37 & 53.93 & 45.81 & 44.29 & 40.92 & 32.29 & 20.96 & 36.85 \\ \cmidrule{2-12}
    &  0.0       & 1.0s            & 55.09 & 71.87 & 55.96 & \textbf{74.07} & \textbf{72.05} & \textbf{69.54} & \textbf{65.81} & \textbf{59.28} & \textbf{68.15} \\
    &  0.05      & 1.0s            & 65.74 & \textbf{73.82} & 65.65 & 72.63 & 70.60 & 68.14 & 64.44 & 58.04 & 66.77 \\
    &  0.1       & 1.0s            & 69.02 & 72.32 & 66.98 & 69.87 & 67.99 & 65.71 & 62.29 & 56.17 & 64.41 \\
    &  0.15      & 1.0s            & 71.61 & 70.33 & 67.18 & 66.59 & 64.76 & 62.83 & 59.75 & 54.23 & 61.63 \\
    &  0.2       & 1.0s            & 72.63 & 68.87 & \textbf{67.26} & 63.99 & 62.32 & 60.62 & 57.88 & 52.79 & 59.52 \\
    &  0.25      & 1.0s            & \textbf{73.27} & 66.99 & 66.84 & 61.76 & 60.27 & 58.78 & 56.31 & 51.42 & 57.71 \\ \cmidrule{2-12}
    &  0.0       & 2.0s            & 53.31 & 68.90 & 53.53 & 71.61 & 68.95 & 65.86 & 63.05 & 56.53 & 65.20 \\
    &  0.05      & 2.0s            & 59.24 & 69.98 & 59.74 & 70.45 & 67.70 & 64.52 & 61.81 & 55.38 & 63.97 \\
    &  0.1       & 2.0s            & 64.35 & 69.29 & 62.97 & 67.74 & 65.15 & 62.23 & 59.79 & 53.64 & 61.71 \\
    &  0.15      & 2.0s            & 66.97 & 67.45 & 63.83 & 64.14 & 62.31 & 59.9 & 57.64 & 51.75 & 59.15 \\
    &  0.2       & 2.0s            & 69.67 & 65.79 & 64.15 & 61.32 & 59.92 & 57.96 & 55.91 & 50.39 & 57.10 \\
    &  0.25      & 2.0s            & 69.90 & 63.15 & 63.00 & 58.07 & 56.88 & 55.16 & 53.31 & 48.22 & 54.33 \\
    \midrule
    \multirow{18}{*}{\rotatebox[origin=c]{90}{Inertial + Camera}}
    &  0.0       & 0.5s        & 58.49 & 81.20 & 64.60 & 76.95 & 75.25 & 69.60 & 54.99 & 38.62 & 63.08 \\
    &  0.05      & 0.5s        & 70.50 & 82.93 & 73.57 & 75.67 & 73.92 & 68.28 & 53.51 & 37.45 & 61.76 \\
    &  0.1       & 0.5s        & 75.70 & 80.32 & 76.13 & 72.35 & 70.82 & 65.52 & 50.69 & 35.30 & 58.94 \\
    &  0.15      & 0.5s        & 79.23 & 75.95 & 75.91 & 68.58 & 67.28 & 62.02 & 47.49 & 33.52 & 55.78 \\
    &  0.2       & 0.5s        & 82.40 & 70.96 & 73.76 & 64.95 & 63.89 & 58.49 & 44.67 & 31.77 & 52.75 \\
    &  0.25      & 0.5s        & 83.87 & 64.68 & 70.06 & 60.10 & 59.29 & 53.92 & 40.80 & 29.37 & 48.70 \\ \cmidrule{2-12}
    &  0.0       & 1.0s        & 60.91 & 82.08 & 64.96 & \textbf{84.41} & \textbf{82.67} & \textbf{79.73} & \textbf{76.01} & \textbf{68.01} & \textbf{78.16}  \\
    &  0.05      & 1.0s        & 72.45 & \textbf{83.75} & 75.61 & 83.50 & 81.77 & 78.76 & 75.07 & 67.02 & 77.22 \\
    &  0.1       & 1.0s        & 77.00 & 82.96 & 78.46 & 81.63 & 79.83 & 76.97 & 73.38 & 65.52 & 75.46 \\
    &  0.15      & 1.0s        & 79.84 & 81.61 & 79.43 & 79.70 & 77.92 & 75.01 & 71.70 & 64.22 & 73.71 \\
    &  0.2       & 1.0s        & 82.38 & 80.30 & 80.15 & 77.63 & 75.97 & 73.28 & 70.31 & 63.04 & 72.05 \\
    &  0.25      & 1.0s        & \textbf{84.48} & 78.66 & \textbf{80.20} & 75.58 & 74.02 & 71.52 & 68.65 & 61.80 & 70.31 \\ \cmidrule{2-12} 
    &  0.0       & 2.0s        & 56.73 & 77.66 & 60.37 & 78.90 & 75.83 & 72.84 & 69.29 & 63.15 & 72.00 \\
    &  0.05      & 2.0s        & 64.75 & 78.75 & 68.25 & 77.56 & 74.55 & 71.65 & 68.11 & 62.09 & 70.80 \\
    &  0.1       & 2.0s        & 71.04 & 77.83 & 72.35 & 75.64 & 72.87 & 70.07 & 66.46 & 60.54 & 69.12 \\
    &  0.15      & 2.0s        & 75.27 & 75.80 & 73.75 & 73.23 & 70.66 & 68.04 & 64.52 & 58.82 & 67.06 \\
    &  0.2       & 2.0s        & 79.19 & 73.88 & 74.52 & 71.10 & 68.79 & 66.38 & 63.00 & 57.54 & 65.36 \\
    &  0.25      & 2.0s        & 81.26 & 72.13 & 74.26 & 69.17 & 66.79 & 64.40 & 61.18 & 56.14 & 63.53 \\ 
        \bottomrule

  \end{tabular}
\end{table}

\begin{table}
  \centering
  \scriptsize
  \caption{TriDet score thresholding results \citep{shiTriDetTemporalAction2023} ablation experiments on the effect of different score thresholds (0.05, 0.1, 0.15, 0.2 and 0.25) on the raw prediction results (0.0 threshold) of experiments involving the TriDet model. We report results on the TriDet being applied to only inertial, camera and a combined (inertial + camera) features using three clip length sizes (0.5, 1.0 and 2.0 seconds) each with a 50\% overlap. Best results per modality are in \textbf{bold}.}
  \label{tab:tridetpostprocessing}
  \begin{tabular}{@{}llllccccccccc@{}}
    & Threshold & CL & P & R & F1 & \multicolumn{6}{c}{mAP} \\ \cmidrule{7-12} 
    & & & & & & 0.3 & 0.4 & 0.5 & 0.6 & 0.7 & Avg \\
    \toprule
    \multirow{18}{*}{\rotatebox[origin=c]{90}{Inertial}}
    &  0.0       & 0.5s        & 54.94 & 77.88 & 61.53 & 76.30 & 73.57 & 67.90 & 59.18 & 49.16 & 65.22 \\
    &  0.05      & 0.5s        & 68.51 & \textbf{79.26} & 70.92 & 75.39 & 72.72 & 67.04 & 58.34 & 48.35 & 64.37 \\
    &  0.1       & 0.5s        & 77.48 & 78.04 & 76.26 & 73.34 & 70.84 & 65.04 & 56.14 & 46.43 & 62.36 \\
    &  0.15      & 0.5s        & 82.56 & 75.02 & \textbf{77.19} & 70.28 & 67.93 & 62.10 & 53.15 & 43.92 & 59.48 \\
    &  0.2       & 0.5s        & 86.06 & 70.10 & 75.25 & 66.01 & 63.71 & 57.70 & 49.30 & 41.09 & 55.56 \\
    &  0.25      & 0.5s        & \textbf{87.78} & 63.97 & 71.09 & 60.73 & 58.58 & 52.94 & 45.06 & 37.85 & 51.03 \\ \cmidrule{2-12}
    &  0.0       & 1.0s        & 55.34 & 78.01 & 60.87 & b & \textbf{78.45} & \textbf{76.11} & \textbf{72.94} & \textbf{67.48} & \textbf{75.03} \\
    &  0.05      & 1.0s        & 66.81 & 79.22 & 70.05 & 79.42 & 77.66 & 75.28 & 72.22 & 66.73 & 74.26 \\
    &  0.1       & 1.0s        & 75.83 & 77.89 & 75.28 & 77.92 & 76.26 & 73.95 & 70.86 & 65.47 & 72.89 \\
    &  0.15      & 1.0s        & 80.70 & 75.94 & 76.91 & 75.97 & 74.38 & 72.16 & 69.09 & 64.04 & 71.13 \\
    &  0.2       & 1.0s        & 83.85 & 73.76 & 77.12 & 73.27 & 71.66 & 69.83 & 66.79 & 62.25 & 68.76 \\
    &  0.25      & 1.0s        & 85.73 & 71.77 & 76.59 & 70.96 & 69.39 & 67.51 & 64.72 & 60.43 & 66.60 \\ \cmidrule{2-12} 
    &  0.0       & 2.0s        & 50.57 & 75.56 & 58.19 & 74.94 & 72.67 & 70.35 & 67.05 & 61.67 & 69.33 \\
    &  0.05      & 2.0s        & 63.35 & 75.93 & 66.64 & 73.77 & 71.50 & 69.14 & 66.04 & 60.82 & 68.26 \\
    &  0.1       & 2.0s        & 71.97 & 74.06 & 71.04 & 71.23 & 68.99 & 66.91 & 63.87 & 59.06 & 66.01 \\
    &  0.15      & 2.0s        & 77.71 & 71.69 & 72.47 & 68.04 & 65.97 & 64.04 & 61.08 & 56.62 & 63.15 \\
    &  0.2       & 2.0s        & 81.72 & 69.37 & 72.53 & 65.57 & 63.65 & 61.86 & 59.07 & 54.82 & 60.99 \\
    &  0.25      & 2.0s        & 84.13 & 67.14 & 71.99 & 63.01 & 61.13 & 59.28 & 56.78 & 52.98 & 58.64 \\
    \midrule
    \multirow{18}{*}{\rotatebox[origin=c]{90}{Camera}}
    &  0.0       & 0.5s            & 49.81 & 70.46 & 54.24 & 67.44 & 65.10 & 59.96 & 47.89 & 31.61 & 54.40 \\
    &  0.05      & 0.5s            & 60.74 & 71.84 & 61.94 & 65.69 & 63.30 & 58.37 & 46.42 & 30.48 & 52.85 \\
    &  0.1       & 0.5s            & 64.66 & 69.02 & 63.90 & 61.28 & 59.04 & 54.71 & 43.57 & 28.40 & 49.40 \\
    &  0.15      & 0.5s            & 66.39 & 63.73 & 62.19 & 56.29 & 54.18 & 50.10 & 39.76 & 25.88 & 45.24 \\
    &  0.2       & 0.5s            & 68.06 & 57.68 & 58.47 & 51.27 & 49.45 & 45.74 & 36.10 & 23.38 & 41.19 \\
    &  0.25      & 0.5s            & 66.90 & 51.37 & 53.93 & 45.81 & 44.29 & 40.92 & 32.29 & 20.96 & 36.85 \\ \cmidrule{2-12}
    &  0.0       & 1.0s            & 55.09 & 71.87 & 55.96 & \textbf{74.07} & \textbf{72.05} & \textbf{69.54} & \textbf{65.81} & \textbf{59.28} & \textbf{68.15} \\
    &  0.05      & 1.0s            & 65.74 & \textbf{73.82} & 65.65 & 72.63 & 70.60 & 68.14 & 64.44 & 58.04 & 66.77 \\
    &  0.1       & 1.0s            & 69.02 & 72.32 & 66.98 & 69.87 & 67.99 & 65.71 & 62.29 & 56.17 & 64.41 \\
    &  0.15      & 1.0s            & 71.61 & 70.33 & 67.18 & 66.59 & 64.76 & 62.83 & 59.75 & 54.23 & 61.63 \\
    &  0.2       & 1.0s            & 72.63 & 68.87 & \textbf{67.26} & 63.99 & 62.32 & 60.62 & 57.88 & 52.79 & 59.52 \\
    &  0.25      & 1.0s            & \textbf{73.27} & 66.99 & 66.84 & 61.76 & 60.27 & 58.78 & 56.31 & 51.42 & 57.71 \\ \cmidrule{2-12}
    &  0.0       & 2.0s            & 53.31 & 68.90 & 53.53 & 71.61 & 68.95 & 65.86 & 63.05 & 56.53 & 65.20 \\
    &  0.05      & 2.0s            & 59.24 & 69.98 & 59.74 & 70.45 & 67.70 & 64.52 & 61.81 & 55.38 & 63.97 \\
    &  0.1       & 2.0s            & 64.35 & 69.29 & 62.97 & 67.74 & 65.15 & 62.23 & 59.79 & 53.64 & 61.71 \\
    &  0.15      & 2.0s            & 66.97 & 67.45 & 63.83 & 64.14 & 62.31 & 59.9 & 57.64 & 51.75 & 59.15 \\
    &  0.2       & 2.0s            & 69.67 & 65.79 & 64.15 & 61.32 & 59.92 & 57.96 & 55.91 & 50.39 & 57.10 \\
    &  0.25      & 2.0s            & 69.90 & 63.15 & 63.00 & 58.07 & 56.88 & 55.16 & 53.31 & 48.22 & 54.33 \\
    \midrule
    \multirow{18}{*}{\rotatebox[origin=c]{90}{Inertial + Camera}}
    &  0.0       & 0.5s        & 58.91 & 80.98 & 64.74 & 80.30 & 78.52 & 74.52 & 67.53 & 56.76 & 71.53 \\
    &  0.05      & 0.5s        & 73.96 & 82.69 & 75.74 & 78.95 & 77.12 & 73.10 & 66.16 & 55.44 & 70.15 \\
    &  0.1       & 0.5s        & 81.07 & 79.82 & 78.94 & 75.31 & 73.67 & 69.70 & 62.86 & 52.32 & 66.77 \\
    &  0.15      & 0.5s        & 84.88 & 75.43 & 78.36 & 71.95 & 70.36 & 66.35 & 59.14 & 49.20 & 63.40 \\
    &  0.2       & 0.5s        & 87.85 & 70.34 & 75.90 & 67.65 & 66.05 & 62.22 & 55.55 & 46.12 & 59.52 \\
    &  0.25      & 0.5s        & \textbf{88.98} & 63.95 & 71.24 & 62.04 & 60.46 & 56.70 & 50.70 & 42.00 & 54.38 \\ \cmidrule{2-12}
    &  0.0       & 1.0s        & 58.69 & 81.51 & 64.16 & \textbf{84.95} & \textbf{83.77} & \textbf{82.05} & \textbf{79.49} & \textbf{74.19} & \textbf{80.89} \\
    &  0.05      & 1.0s        & 71.84 & \textbf{83.39} & 75.19 & 84.03 & 82.83 & 81.13 & 78.55 & 73.17 & 79.94 \\
    &  0.1       & 1.0s        & 78.61 & 82.79 & 79.37 & 82.69 & 81.48 & 79.76 & 77.15 & 71.88 & 78.59 \\
    &  0.15      & 1.0s        & 82.64 & 81.42 & 80.93 & 80.99 & 79.76 & 78.09 & 75.60 & 70.57 & 77.00 \\
    &  0.2       & 1.0s        & 84.99 & 79.55 & \textbf{81.08} & 78.64 & 77.45 & 75.74 & 73.40 & 68.79 & 74.81 \\
    &  0.25      & 1.0s        & 86.81 & 77.11 & 80.38 & 75.60 & 74.39 & 72.76 & 70.40 & 66.20 & 71.87 \\ \cmidrule{2-12} 
    &  0.0       & 2.0s        & 51.17 & 78.44 & 60.46 & 79.51 & 77.74 & 75.56 & 72.54 & 68.28 & 74.73 \\
    &  0.05      & 2.0s        & 66.62 & 79.44 & 70.01 & 78.08 & 76.28 & 74.17 & 71.23 & 67.13 & 73.38 \\
    &  0.1       & 2.0s        & 74.65 & 78.03 & 74.43 & 75.36 & 73.74 & 71.85 & 69.13 & 65.12 & 71.04 \\
    &  0.15      & 2.0s        & 79.86 & 76.49 & 76.35 & 73.39 & 71.72 & 69.95 & 67.56 & 63.76 & 69.28 \\
    &  0.2       & 2.0s        & 83.10 & 74.55 & 76.72 & 71.20 & 69.69 & 67.88 & 65.49 & 61.77 & 67.20 \\
    &  0.25      & 2.0s        & 84.29 & 72.39 & 76.03 & 68.64 & 67.07 & 65.30 & 63.02 & 59.49 & 64.70 \\
    \bottomrule
  \end{tabular}
\end{table}

\clearpage
\subsection{Single-Stage Temporal Action Localization for Inertial Data}

\begin{figure}[H]
\begin{center}
   \includegraphics[width=0.8\linewidth]{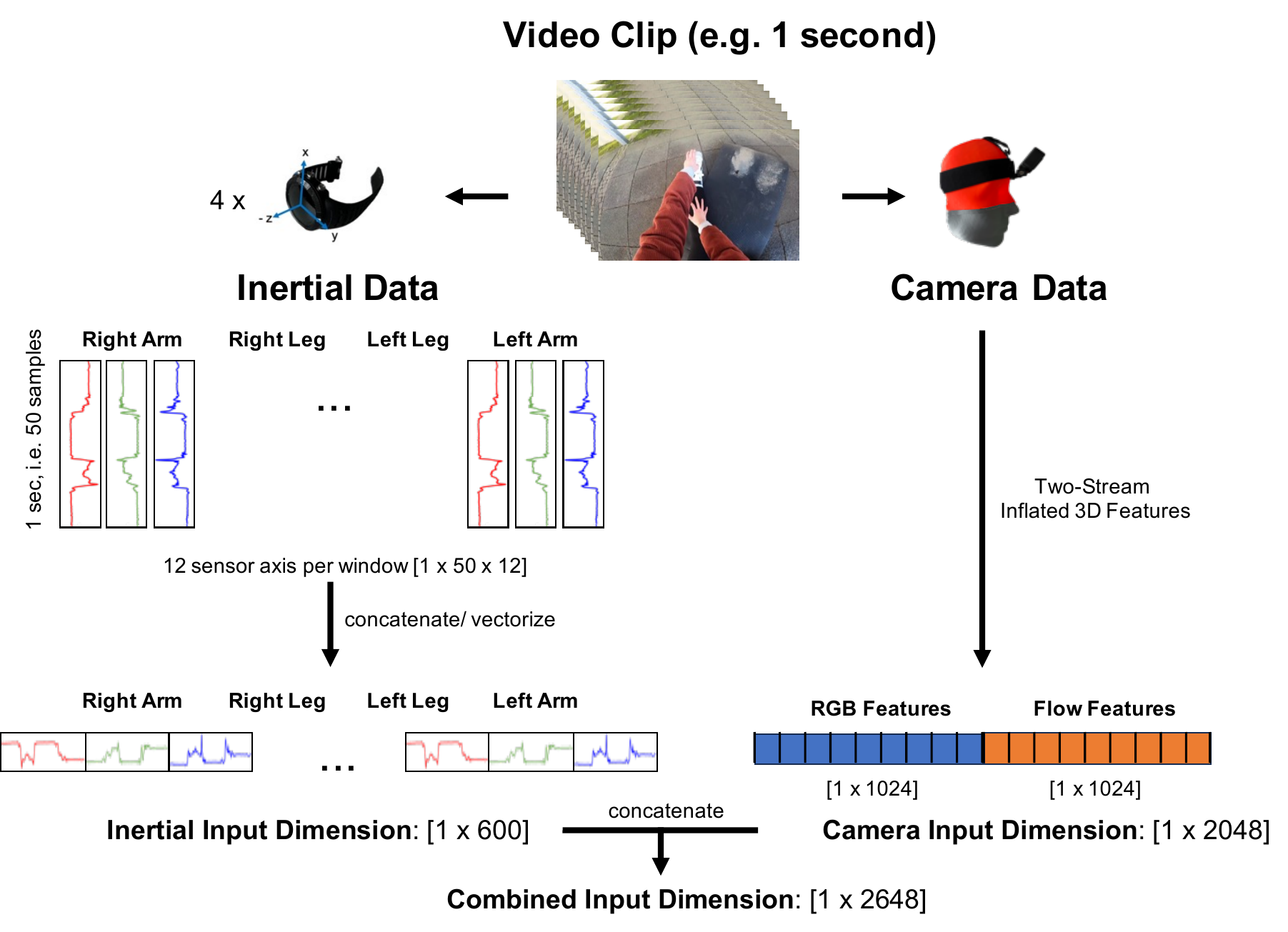}
\end{center}
   \caption{Visualization of the applied preprocessing on inertial and camera data in order to make to create a feature embedding which can be used to train the TriDet \citet{shiTriDetTemporalAction2023} and ActionFormer \citet{zhangActionFormerLocalizingMoments2022} network. }
\label{fig:preprocessing}
\end{figure}

In this paper we demonstrated the applicability of vision-based single-stage temporal action localization models on a previously unexplored modality, i.e. inertial data. As the investigated architectures, namely the TriDet \citep{shiTriDetTemporalAction2023} and ActionFormer \citep{zhangActionFormerLocalizingMoments2022}, both require clip-based, one-dimensional feature embeddings as input, data of both camera and inertial sensors had to be preprocessed. Figure~\ref{fig:preprocessing} summarizes the applied preprocessing on both modalities. First step for both modalities included windowing the data streams using a predefined clip length and overlap. In total three different clip lengths were tested (0.5, 1 and 2 seconds). Having windowed the inertial data left us with a 3-dimensional feature array, i.e. $[\textit{no. windows}, \textit{window length}, \textit{no. sensor axis}]$. In order to obtain a vectorized feature embedding per sliding window, individual sensor axis were concatenated. Depending on the window length this left us with a one-dimensional feature vector of size 300 (0.5 second), 600 (1 second) and 1200 (2 seconds) per video clip, i.e. sliding window. Contrarily, as also applied in \citet{shiTriDetTemporalAction2023}, we extracted two-stream I3D feature embeddings \citep{carreiraQuoVadisAction2017} pretrained on Kinetics-400 \citep{kayKineticsHumanAction2017} from the raw video stream, resulting in a vision-based embedding of size 2048 per video clip. Having vectorized both modalities we were able to train both temporal action localization architectures on either (1) inertial, (2) camera or (3) a concatenation of the two (inertial + camera).
Even though our concatenation approach results in varying input dimensions, said change does not come at increased computational costs. More specifically, while amount of learnable parameters marginally increases (not more than 10\%) with an increased input dimension, unlike other approaches, no additional embedding needs to be extracted from the inertial data and raw data streams can directly be used.

% \begin{table}
%   \centering
%   \small
%   \begin{tabular}{@{}lccccc@{}}
%     & Clip-L. & Shallow D. & A-and-D & ActionFormer & TriDet \\
%     \toprule
%     \multirow{3}{*}{\rotatebox[origin=c]{90}{I}}
%     & 0.5s & 7.429M & 5.605M & 26.564M & 13.288M \\
%     & 1.0s & 7.503M & 5.679M & 27.024M & 13.749M \\
%     & 2.0s & 7.626M & 5.803M & 27.946M & 14.671M \\
%     \midrule
%     \multirow{3}{*}{\rotatebox[origin=c]{90}{I + C}}
%     & 0.5s & - & - & 29.709M & 18.271M \\
%     & 1.0s & - & - & 30.170M & 18.732M \\
%     & 2.0s & - & - & 31.092M & 19.653M \\
%   \end{tabular}
%   \caption{}
%   \label{tab:parameters}
% \end{table}

\subsection{Ablation Study on Influence of Frequency of Inputs}

With the frequencies both the camera (60 FPS) and inertial sensors (50 HZ) being set fairly high, the WEAR dataset allows to explore lower frequency experiments and their effect fewer datapoints per second might have on the predictive quality of the trained models. Table \ref{tab:freq} summarizes experiments conducted using only 50\% and 20\% of the available frequency for both types of sensors. Note that a clip length of 0.5 seconds was not explored during experiments as it was not possible anymore to extract two-stream I3D feature embeddings \citep{carreiraQuoVadisAction2017} as the amount of frames was lower than the required minimum input frames. Looking at results presented in Table \ref{tab:freq} one can see that all models trained using only inertial data suffered from lower frequency inputs with both classification and mAP scores decreasing. Contrarily, models trained using camera-based improved when using features extracted from videos with a lower FPS, which might be caused by Kinetics-400 \citep{kayKineticsHumanAction2017}, which was used for pretraining the I3D extraction method, on consisting of videos with a lower FPS than the WEAR dataset.

\begin{table*}
  \centering
  \scriptsize
  \caption{Results of evaluating different frequencies (Freq.) as input for different clip lengths (CL) on our WEAR dataset. Both inertial- and camera-based features were downsampled to be only 50\% (i.e. 30 FPS and 25 Hz) and 20\% (i.e. 12 FPS and 10 Hz) of the original frequency input (i.e. 60 FPS and 50 Hz). One can see that the predictive performance of inertial models  decreases with a lower frequency input. Interestingly camera and combined models increase in performance when lower frequency inputs with I3D being calculated on 12 FPS videos resulting in the highest classification and mAP scores during camera-based experiments. Experiments are evaluated in terms of precision (P), recall (R), F1-score and mean average precision (mAP) for different temporal intersection over union (tIoU) thresholds. Best results per modality are in \textbf{bold}.}
  \label{tab:freq}
  \begin{tabular}{@{}llccccccccccc@{}}
    & Threshold & CL & P & R & F1 & \multicolumn{6}{c}{mAP} \\ \cmidrule{7-12} 
    & & & & & & 0.3 & 0.4 & 0.5 & 0.6 & 0.7 & Avg \\
    \toprule
    \multirow{24}{*}{\rotatebox[origin=c]{90}{Inertial}}
    & Shallow D.              & Orig & 1s                  & 88.02 & 77.03 & 80.86 & 57.09 & 55.32 & 53.61 & 50.59 & 47.85 & 52.89 \\
    & Shallow D.              & 50\% & 1s                  & 87.02 & 76.51 & 80.10 & 55.33 & 52.70 & 51.02 & 48.30 & 45.67 & 50.61 \\
    & Shallow D.              & 20\% & 1s                  & 86.38 & 76.10 & 79.59 & 53.94 & 51.95 & 50.05 & 47.60 & 45.19 & 49.75 \\ \cmidrule{2-13}
    & A-and-D                 & Orig & 1s                  & 87.87 & 79.02 & \textbf{82.01} & 56.38 & 54.47 & 52.28 & 50.07 & 46.92 & 52.03 \\
    & A-and-D                 & 50\% & 1s                  & 87.57 & 78.25 & 81.23 & 55.88 & 53.51 & 51.33 & 48.16 & 44.78 & 50.73 \\
    & A-and-D                 & 20\% & 1s                  & 86.10 & 79.88 & 81.67 & 56.76 & 54.87 & 53.03 & 49.85 & 47.39 & 52.38 \\ \cmidrule{2-13}
    & ActionFormer            & Orig & 1s                  & 81.69 & 75.37 & 76.86 & 72.90 & 71.30 & 68.28 & 64.14 & 56.65 & 66.65 \\
    & ActionFormer            & 50\% & 1s                  & 80.93 & 73.66 & 75.62 & 71.40 & 69.69 & 66.77 & 63.09 & 56.01 & 65.39 \\
    & ActionFormer            & 20\% & 1s                  & 80.73 & 72.43 & 74.51 & 70.07 & 68.34 & 65.44 & 60.83 & 54.45 & 63.82 \\ \cmidrule{2-13}
    & TriDet                  & Orig & 1s                  & 83.85 & 73.76 & 77.12 & \textbf{73.27} & \textbf{71.66} & \textbf{69.83} & \textbf{66.79} & \textbf{62.25} & \textbf{68.76} \\
    & TriDet                  & 50\% & 1s                  & 84.52 & 72.82 & 76.67 & 72.01 & 70.62 & 68.86 & 65.13 & 60.32 & 67.39 \\
    & TriDet                  & 20\% & 1s                  & 84.30 & 71.72 & 75.68 & 70.60 & 69.38 & 67.34 & 63.75 & 57.61 & 65.74 \\ \cmidrule{2-13}
    & Shallow D.              & Orig & 2s                  & 87.92 & 78.16 & 81.60 & 59.89 & 57.00 & 54.69 & 51.77 & 48.99 & 54.47 \\ 
    & Shallow D.              & 50\% & 2s                  & 85.08 & 76.13 & 79.09 & 53.85 & 51.57 & 49.30 & 46.50 & 43.79 & 49.00 \\
    & Shallow D.              & 20\% & 2s                  & 84.59 & 75.39 & 78.33 & 53.21 & 50.98 & 48.66 & 45.69 & 43.08 & 48.32 \\ \cmidrule{2-13}
    & A-and-D                 & Orig & 2s                  & \textbf{88.24} & \textbf{80.55} & 83.08 & 58.32 & 56.68 & 54.44 & 51.58 & 48.34 & 53.87 \\
    & A-and-D                 & 50\% & 2s                  & 87.24 & 78.05 & 80.88 & 53.69 & 51.36 & 48.58 & 45.99 & 42.66 & 48.46\\
    & A-and-D                 & 20\% & 2s                  & 86.98 & 77.94 & 80.83 & 55.63 & 53.32 & 49.76 & 46.45 & 43.74 & 49.78 \\ \cmidrule{2-13}
    & ActionFormer            & Orig & 2s                  & 78.18 & 69.15 & 71.15 & 66.43 & 63.30 & 60.47 & 56.66 & 50.26 & 59.43 \\
    & ActionFormer            & 50\% & 2s                  & 77.85 & 67.46 & 70.24 & 64.88 & 62.47 & 59.26 & 55.65 & 49.35 & 58.32 \\ 
    & ActionFormer            & 20\% & 2s                  & 76.84 & 65.71 & 68.69 & 62.51 & 59.90 & 56.87 & 52.64 & 45.50 & 55.48 \\ \cmidrule{2-13}
    & TriDet                  & Orig & 2s                  & 81.72 & 69.37 & 72.53 & 65.57 & 63.65 & 61.86 & 59.07 & 54.82 & 60.99 \\
    & TriDet                  & 50\% & 2s                  & 80.61 & 65.51 & 69.77 & 62.32 & 60.60 & 58.20 & 55.90 & 51.50 & 57.71 \\
    & TriDet                  & 20\% & 2s                  & 78.59 & 63.55 & 67.79 & 59.87 & 58.24 & 56.09 & 53.33 & 47.92 & 55.09 \\
    \midrule
    \multirow{12}{*}{\rotatebox[origin=c]{90}{Camera}}
    & ActionFormer            & Orig & 1s                  & 72.63 & \textbf{68.87} & 67.26 & 63.99 & 62.32 & 60.62 & 57.88 & 52.79 & 59.52 \\
    & ActionFormer            & 50\% & 1s                  & 74.62 & 68.58 & 67.81 & 64.61 & 63.12 & 61.28 & 58.56 & 53.77 & 60.27 \\
    & ActionFormer            & 20\% & 1s                  & 74.36 & 67.82 & 67.92 & 65.92 & 64.36 & 62.98 & 59.99 & 55.31 & 61.71 \\ \cmidrule{2-13}
    & TriDet                  & Orig & 1s                  & 75.32 & 68.07 & 67.95 & 64.36 & 63.30 & 61.38 & 59.13 & 54.64 & 60.56 \\
    & TriDet                  & 50\% & 1s                  & 77.21 & 68.41 & 68.82 & 66.01 & 65.06 & 63.46 & 61.53 & 57.56 & 62.72 \\
    & TriDet                  & 20\% & 1s                  & \textbf{75.95} & 68.41 & \textbf{69.10} & \textbf{66.61} & \textbf{65.71} & \textbf{63.72} & \textbf{61.85} & \textbf{57.66} & \textbf{63.11} \\ \cmidrule{2-13}
    & ActionFormer            & Orig & 2s                  & 69.67 & 65.79 & 64.15 & 61.32 & 59.92 & 57.96 & 55.91 & 50.39 & 57.10 \\
    & ActionFormer            & 50\% & 2s                  & 72.64 & 67.53 & 66.43 & 64.22 & 62.25 & 60.65 & 57.71 & 52.93 & 59.55 \\ 
    & ActionFormer            & 20\% & 2s                  & 71.94 & 65.39 & 65.62 & 61.94 & 59.93 & 58.10 & 54.62 & 49.88 & 56.89 \\ \cmidrule{2-13}
    & TriDet                  & Orig & 2s                  & 73.85 & 64.09 & 64.25 & 60.95 & 60.03 & 57.75 & 55.55 & 52.19 & 57.30 \\
    & TriDet                  & 50\% & 2s                  & 75.08 & 66.27 & 67.10 & 64.18 & 62.98 & 61.37 & 59.95 & 56.11 & 60.92 \\
    & TriDet                  & 20\% & 2s                  & 74.04 & 63.20 & 64.98 & 59.48 & 58.27 & 56.82 & 55.70 & 52.05 & 56.47 \\
    \midrule
    \multirow{12}{*}{\rotatebox[origin=c]{90}{Inertial + Camera}}
    & ActionFormer            & Orig & 1s                  & 82.38 & \textbf{80.30} & 80.15 & 77.63 & 75.97 & 73.28 & 70.31 & 63.04 & 72.05 \\
    & ActionFormer            & 50\% & 1s                  & 82.04 & 80.24 & 79.84 & 76.98 & 75.34 & 73.35 & 69.60 & 63.07 & 71.67 \\
    & ActionFormer            & 20\% & 1s                  & 81.89 & 79.33 & 79.24 & 76.24 & 74.92 & 72.95 & 70.43 & 63.08 & 71.52 \\ \cmidrule{2-13}
    & TriDet                  & Orig & 1s                  & 84.99 & 79.55 & 81.08 & \textbf{78.64} & \textbf{77.45} & \textbf{75.74} & \textbf{73.40} & 68.79 & \textbf{74.81} \\
    & TriDet                  & 50\% & 1s                  & \textbf{86.25} & 79.48 & \textbf{81.46} & 77.68 & 77.07 & 75.26 & 73.05 & \textbf{68.94} & 74.40 \\
    & TriDet                  & 20\% & 1s                  & 84.79 & 79.13 & 80.45 & 77.55 & 76.83 & 75.05 & 71.94 & 68.19 & 73.91 \\ \cmidrule{2-13}
    & ActionFormer            & Orig & 2s                  & 79.19 & 73.88 & 74.52 & 71.10 & 68.79 & 66.38 & 63.00 & 57.54 & 65.36 \\
    & ActionFormer            & 50\% & 2s                  & 79.24 & 75.14 & 75.55 & 70.84 & 68.13 & 65.71 & 62.77 & 57.33 & 64.96 \\
    & ActionFormer            & 20\% & 2s                  & 78.47 & 73.78 & 74.15 & 68.60 & 66.69 & 63.24 & 60.14 & 55.78 & 62.89 \\ \cmidrule{2-13}
    & TriDet                  & Orig & 2s                  & 83.10 & 74.55 & 76.72 & 71.20 & 69.69 & 67.88 & 65.49 & 61.77 & 67.20 \\
    & TriDet                  & 50\% & 2s                  & 81.33 & 73.58 & 75.69 & 70.52 & 69.06 & 67.31 & 64.91 & 61.32 & 66.62 \\
    & TriDet                  & 20\% & 2s                  & 82.51 & 72.91 & 75.48 & 69.09 & 67.08 & 64.87 & 62.59 & 59.41 & 64.61 \\
    \bottomrule
  \end{tabular}
\end{table*}

\subsection{Ablation Study on Inertial Sensor Selection}

As reported experimental results are based on acceleration recordings of all limbs, the following experiments investigate how the predictive performance of each algorithm is affected by using only (1) acceleration recorded from the right wrist and (2) acceleration recorded from both the right wrist and right ankle. Results in Table \ref{tab:right_wrist} show that using only acceleration data obtained from the right wrist significantly decreases predictive performance across all algorithms across all metrics. Moreover, Table \ref{tab:right_wrist_and_ankle} clearly underlines the value of additionally measuring acceleration at the ankles' of participants, as results again significantly increase, being mostly on par compared to using all four inertial sensor locations. Interestingly, unlike the inertial-based architectures, results of vision-based models improve when excluding data captured by the left-wrist and left-ankle inertial sensors, which could be caused by the dataset being biased towards right-handed participants (see Table \ref{tab:questionnaire}) and dominant hand movement might being overall more consistent. Figure \ref{fig:sensor_selection} shows the per-class results of the TriDet model \citep{shiTriDetTemporalAction2023} being trained using (1) data obtained from the right wrist inertial sensor, (2) right wrist and right ankle inertial sensor and (3) all four inertial sensors (right and left wrists and ankles).

\begin{table*}
  \centering
  \small
  \caption{Results of using only inertial features captured by the sensor placed on the right wrist for different clip lengths (CL) on our WEAR dataset evaluated in terms of precision (P), recall (R), F1-score and mean average precision (mAP) for different temporal intersection over union (tIoU) thresholds. One can see a clear overall decrease across all evaluation metrics. Best results per modality are in \textbf{bold}.}
  \label{tab:right_wrist}
  \begin{tabular}{@{}llcccccccccc@{}}
    & Model & CL & P & R & F1 & \multicolumn{6}{c}{mAP} \\ \cmidrule{7-12} 
    & & & & & & 0.3 & 0.4 & 0.5 & 0.6 & 0.7 & Avg \\
    \toprule
    \multirow{12}{*}{\rotatebox[origin=c]{90}{Inertial}}
    & Shallow D.              & 0.5s                & 64.54 & 68.08 & 64.23 & 23.26 & 21.57 & 19.27 & 17.31 & 16.00 & 19.48 \\
    & A-and-D                 & 0.5s                & 75.34 & 64.09 & 66.93 & 27.08 & 25.33 & 22.55 & 20.53 & 18.94 & 22.89 \\
    & ActionFormer            & 0.5s                & 72.96 & 63.59 & 65.30 & 54.45 & 52.42 & 45.70 & 34.82 & 22.11 & 41.90 \\
    & TriDet                  & 0.5s                & \textbf{79.48} & 62.89 & 66.98 & 54.32 & 52.10 & 47.57 & 40.39 & 30.38 & 44.95 \\
    & Shallow D.              & 1s                  & 66.98 & 68.81 & 66.19 & 25.53 & 23.62 & 22.11 & 19.56 & 18.28 & 21.82 \\
    & A-and-D                 & 1s                  & 75.56 & 64.31 & 67.21 & 29.18 & 26.39 & 23.52 & 21.60 & 19.57 & 24.05 \\
    & ActionFormer            & 1s                  & 73.07 & 65.51 & 66.91 & 61.00 & 58.05 & 52.69 & 47.32 & 39.82 & 51.78 \\
    & TriDet                  & 1s                  & 78.04 & \textbf{67.88} & \textbf{70.44} & \textbf{63.08} & \textbf{62.09} & \textbf{60.07} & \textbf{57.07} & \textbf{50.36} & \textbf{58.54} \\
    & Shallow D.              & 2s                  & 66.79 & 67.68 & 65.34 & 28.34 & 26.46 & 24.05 & 21.58 & 19.40 & 23.97 \\
    & A-and-D                 & 2s                  & 76.71 & 65.87 & 68.63 & 31.93 & 28.51 & 25.86 & 23.46 & 21.31 & 26.21 \\
    & ActionFormer            & 2s                  & 69.44 & 61.33 & 63.06 & 55.42 & 53.22 & 51.32 & 47.34 & 39.90 & 49.44 \\
    & TriDet                  & 2s                  & 70.73 & 58.22 & 61.08 & 52.06 & 50.54 & 48.51 & 46.05 & 40.93 & 47.62 \\
    \midrule
    \multirow{6}{*}{\rotatebox[origin=c]{90}{I + C}}
    & ActionFormer            & 0.5s                & 76.91 & 65.69 & 67.69 & 58.60 & 57.36 & 51.30 & 38.48 & 26.15 & 46.38 \\
    & TriDet                  & 0.5s                & \textbf{81.87} & 65.19 & 69.57 & 60.83 & 59.12 & 55.57 & 48.84 & 40.71 & 53.01 \\
    & ActionFormer            & 1s                  & 79.62 & \textbf{77.00} & \textbf{76.56} & 72.06 & 70.65 & 68.94 & 66.26 & 60.49 & 67.68 \\
    & TriDet                  & 1s                  & 79.96 & 76.26 & 76.45 & \textbf{74.39} & \textbf{73.55} & \textbf{71.84} & \textbf{69.52} & \textbf{65.88} & \textbf{71.03} \\
    & ActionFormer            & 2s                  & 74.43 & 73.48 & 72.01 & 68.87 & 66.86 & 64.51 & 60.95 & 55.89 & 63.42 \\
    & TriDet                  & 2s                  & 77.07 & 71.70 & 72.43 & 67.87 & 66.93 & 64.93 & 62.12 & 58.30 & 64.03 \\
    \bottomrule
  \end{tabular}
\end{table*}

\begin{table*}
  \centering
  \small
  \caption{Results using only inertial features captured by the sensor placed on the right wrist and right ankle for different clip lengths (CL) on our WEAR dataset evaluated in terms of precision (P), recall (R), F1-score and mean average precision (mAP) for different temporal intersection over union (tIoU) thresholds. Comparing results to \ref{tab:right_wrist} one can see the increase in performance one can achieve when tracking acceleration measured at the ankle in addition to a wrist-worn inertial sensor. Best results per modality are in \textbf{bold}.}
  \label{tab:right_wrist_and_ankle}
  \begin{tabular}{@{}llcccccccccc@{}}
    & Model & CL & P & R & F1 & \multicolumn{6}{c}{mAP} \\ \cmidrule{7-12} 
    & & & & & & 0.3 & 0.4 & 0.5 & 0.6 & 0.7 & Avg \\
    \toprule
    \multirow{12}{*}{\rotatebox[origin=c]{90}{Inertial}}
    & Shallow D.              & 0.5s                & 78.73 & 74.71 & 75.24 & 42.19 & 40.40 & 37.77 & 34.94 & 32.07 & 37.47 \\
    & A-and-D                 & 0.5s                & 81.88 & 69.35 & 73.02 & 41.76 & 39.42 & 35.94 & 32.48 & 30.29 & 35.98 \\
    & ActionFormer            & 0.5s                & 78.62 & 74.39 & 74.47 & 68.28 & 64.52 & 54.70 & 39.06 & 25.95 & 50.50 \\
    & TriDet                  & 0.5s                & \textbf{84.83} & 73.00 & 76.38 & 68.73 & 65.80 & 60.60 & 50.99 & 40.43 & 57.31 \\
    & Shallow D.              & 1s                  & 80.63 & 74.87 & 76.32 & 44.49 & 42.87 & 40.70 & 37.12 & 34.95 & 40.03 \\
    & A-and-D                 & 1s                  & 82.83 & 72.72 & 75.77 & 43.75 & 41.17 & 38.26 & 34.65 & 32.20 & 38.00 \\
    & ActionFormer            & 1s                  & 81.33 & \textbf{78.60} & 78.64 & 76.64 & 74.60 & 70.97 & 65.88 & 58.13 & 69.24 \\
    & TriDet                  & 1s                  & 84.03 & 78.16 & \textbf{79.75} & \textbf{77.84} & \textbf{75.93} & \textbf{73.69} & \textbf{70.80} & \textbf{64.39} & \textbf{72.53} \\
    & Shallow D.              & 2s                  & 80.41 & 75.62 & 76.76 & 46.07 & 44.36 & 41.52 & 38.50 & 35.63 & 41.21 \\
    & A-and-D                 & 2s                  & 84.56 & 76.65 & 79.07 & 50.25 & 47.27 & 43.27 & 40.19 & 36.96 & 43.59 \\
    & ActionFormer            & 2s                  & 77.72 & 73.34 & 73.63 & 70.06 & 67.29 & 64.45 & 60.00 & 52.62 & 62.88 \\
    & TriDet                  & 2s                  & 79.75 & 72.70 & 74.49 & 68.19 & 66.31 & 64.34 & 61.33 & 57.12 & 63.46 \\
    \midrule
    \multirow{6}{*}{\rotatebox[origin=c]{90}{I + C}}
    & ActionFormer            & 0.5s                & 81.20 & 73.51 & 75.19 & 66.99 & 65.65 & 60.32 & 44.59 & 30.61 & 53.63 \\
    & TriDet                  & 0.5s                & 86.97 & 71.16 & 75.78 & 67.41 & 65.68 & 61.54 & 53.14 & 43.45 & 58.24 \\
    & ActionFormer            & 1s                  & 83.01 & \textbf{82.35} & 81.47 & 79.17 & 77.84 & 75.34 & 71.12 & 65.54 & 73.80 \\
    & TriDet                  & 1s                  & \textbf{85.39} & 81.59 & \textbf{82.47} & \textbf{80.22} & \textbf{79.12} & \textbf{77.01} & \textbf{73.81} & \textbf{71.07} & \textbf{76.25} \\
    & ActionFormer            & 2s                  & 78.22 & 77.84 & 76.53 & 73.87 & 71.83 & 69.07 & 64.91 & 59.47 & 67.83 \\
    & TriDet                  & 2s                  & 80.93 & 77.67 & 77.90 & 73.47 & 72.01 & 70.12 & 68.21 & 64.43 & 69.65 \\
    \bottomrule
  \end{tabular}
\end{table*}

\begin{figure}[H]
\begin{center}
   \includegraphics[width=1.\linewidth]{figures/sensor_selection.pdf}
\end{center}
   \caption{Confusion matrices of the TriDet model \citep{shiTriDetTemporalAction2023} being applied using only inertial obtained from the (1) right wrist, (2) right wrist and ankle and (3) right and left wrists and ankles with a one second sliding window and 50\% overlap.}
\label{fig:sensor_selection}
\end{figure}

\subsection{Ablation study on second execution of workout sessions}

In order to explore the robustness of obtained results, we recorded all activities of two participants (sbj\_0 and sbj\_14) a second time in August. Both participants recording conditions significantly differed from their first recording, with temperatures being around 25 degrees Celsius with overall more sunny weather conditions. Further, as not all participants knew all activities beforehand (see Table \ref{tab:questionnaire}), recording the same participants a second time would allow to analyse how a certain degree of familiarity with the recording setup can be seen in altered movements (e.g., via a smoother execution of activities) as well as subject-specific finetuning affects the overall recognition performance. Table \ref{tab:rerecording} compares validation results obtained on the original, first recording of sbj\_0 and sbj\_14 with their second execution of the workout plan. Unlike our prior experiments, each algorithm is trained using the data of all but the validation subjects' recordings, ensuring the validation subjects (sbj\_0 and sbj\_14) remain unseen during the training of each algorithm. All results are postprocessed as reported in the main paper. While, one can see improved results regarding sbj\_0, this trend does not apply to sbj\_14. More specifcially, improvements and decline rates between the two recordings lie within the expected standard deviation across participants (between 15\% to 20\%). Though being a small sample size of only two participants, the results suggest that in order to guarantee a reliable detection of activities, each participant would need to be recorded multiple times under different conditions. Nevertheless, in order to come up with reliable conclusions, future extensions of the WEAR dataset would need focus on re-recording more participants multiple times under varying conditions.

\begin{table*}
  \centering
  \scriptsize
  \caption{Comparison of obtained results of repeated sessions for participants sbj\_0 and sbj\_14 for different clip lengths (CL) on our WEAR dataset evaluated in terms of F1-score and mean average precision (mAP). The two participants were invited to perform the recording plan a second time. While one can see that improved results regarding sbj\_0, suggesting potential learning effects of the correct execution of activities, this trend does not apply to sbj\_14. Note that weather conditions (temperature and sunlight) significantly differ amongst the recordings -- winter (first recording) compared to summer (2nd recording). These figures are, as in the earlier results, averaged across 3 runs using 3 different random seeds. For the first recording, both subjects' best results per modality are in \underline{underlined}. For the second recording, both subjects' best results per modality are in \textbf{bold}. Unlike our prior experiments, each algorithm is trained using the data of all but the validation subjects' recordings, ensuring the validation subjects (sbj\_0 and sbj\_14) remain unseen during the training of each algorithm. All results are postprocessed as reported in the main paper.}
  \label{tab:rerecording}
  \begin{tabular}{@{}llcccccccccc@{}}
    & Model & CL & \multicolumn{4}{c}{sbj\_0} & & \multicolumn{4}{c}{sbj\_14} \\ \cmidrule{4-7} \cmidrule{9-12} 
    &       &         & \multicolumn{2}{c}{1st Recording} & \multicolumn{2}{c}{2nd Recording} & & \multicolumn{2}{c}{1st Recording} & \multicolumn{2}{c}{2nd Recording} \\
    & & & F1 & mAP & F1 & mAP & & F1 & mAP & F1 & mAP \\
    \toprule
    \multirow{12}{*}{\rotatebox[origin=c]{90}{Inertial}}
    & Shallow D.              & 0.5s                & 69.75 & 42.18 & 85.15 & 75.20 & & 77.52 & 65.18 & 77.60 & 62.40 \\
    & A-and-D                 & 0.5s                & 73.51 & 38.02 & 84.10 & 70.60 & & 79.09 & 59.26 & 75.88 & 61.70 \\
    & ActionFormer            & 0.5s                & 76.05 & 69.17 & 81.98 & 72.68 & & 79.02 & 78.01 & 69.09 & 62.24 \\
    & TriDet                  & 0.5s                & 74.84 & 67.57 & 80.07 & 74.97 & & 81.59 & 85.67 & 72.29 & 67.90 \\
    & Shallow D.              & 1s                  & 73.72 & 49.62 & \textbf{85.46} & \textbf{76.47} & & 82.77 & 69.30 & 77.15 & 62.19 \\
    & A-and-D                 & 1s                  & 78.08 & 48.78 & 84.52 & 71.40 & & 79.75 & 62.44 & 76.24 & 63.27 \\
    & ActionFormer            & 1s                  & 77.98 & 75.22 & 75.37 & 83.92 & & 84.01 & 91.39 & 75.67 & 80.49 \\
    & TriDet                  & 1s                  & 76.54 & \underline{75.68} & 72.88 & 81.31 & & \underline{84.92} & \underline{93.81} & 74.75 & 80.08 \\
    & Shallow D.              & 2s                  & 69.75 & 42.18 & 85.01 & 77.84 & & 84.60 & 73.36 & 79.72 & 67.69 \\
    & A-and-D                 & 2s                  & \underline{78.33} & 50.22 & 84.78 & 73.66 & & 82.12 & 65.43 & \textbf{85.40} & 71.87 \\
    & ActionFormer            & 2s                  & 61.20 & 59.68 & 68.56 & 70.85 & & 78.72 & 87.25 & 72.91 & \textbf{81.73} \\
    & TriDet                  & 2s                  & 68.30 & 62.51 & 69.88 & 74.12 & & 82.09 & 91.70 & 74.04 & 79.84 \\
    \midrule
    \multirow{6}{*}{\rotatebox[origin=c]{90}{Camera}}
    & ActionFormer            & 0.5s                & 48.97 & 42.98 & 68.39 & 62.88 & & 58.60 & 53.70 & 70.43 & 73.19 \\
    & TriDet                  & 0.5s                & 51.25 & 50.85 & 70.96 & 69.08 & & 60.98 & 55.52 & 68.52 & 66.87 \\
    & ActionFormer            & 1s                  & \underline{64.29} & \underline{62.62} & \textbf{77.37} & 87.25 & & \underline{74.11} & 78.20 & 63.00 & 82.72 \\
    & TriDet                  & 1s                  & 60.74 & 62.18 & 76.84 & 87.84 & & 66.60 & 73.17 & 62.31 & \textbf{84.26} \\
    & ActionFormer            & 2s                  & 59.88 & 57.27 & 76.76 & \textbf{90.20} & & 73.65 & \underline{82.32} & \textbf{71.78} & 78.15 \\
    & TriDet                  & 2s                  & 55.54 & 58.63 & 76.39 & 84.31 & & 66.91 & 78.75 & 61.31 & 75.83 \\
    \midrule
    \multirow{6}{*}{\rotatebox[origin=c]{90}{I + C}}
    & ActionFormer            & 0.5s                & 79.81 & 69.93 & 80.94 & 74.64 & & 81.65 & 84.12 & 71.43 & 75.35 \\
    & TriDet                  & 0.5s                & 79.98 & 71.20 & 74.35 & 69.08 & & 84.83 & 85.85 & 77.16 & 80.15 \\
    & ActionFormer            & 1s                  & \underline{83.55} & 80.74 & \textbf{87.30} & \textbf{94.12} & & 85.71 & 94.71 & 75.98 & 82.18 \\
    & TriDet                  & 1s                  & 81.75 & \underline{83.90} & 86.51 & 92.22 & & \underline{88.27} & \underline{97.60} & 75.16 & 80.27 \\
    & ActionFormer            & 2s                  & 67.90 & 69.20 & 80.19 & 91.18 & & 82.72 & 94.50 & 77.35 & 88.00 \\
    & TriDet                  & 2s                  & 70.16 & 72.56 & 78.35 & 87.84 & & 83.30 & 94.62 & 76.97 & 85.00 \\
    \bottomrule
  \end{tabular}
\end{table*}

\subsection{Additional Visualizations of Final Runs}

In addition to the visualisations supplied in the main paper, the following provides supplementary visualizations for further analysis. All models mentioned in this section were trained using a clip length of 1.0 second with a 50\% overlap. Predictions made by the temporal action localization models \citep{zhangActionFormerLocalizingMoments2022, shiTriDetTemporalAction2023} were filtered using a score threshold of 0.2 and predictions made by inertial-based architectures \citep{bockImprovingDeepLearning2021, abedinAttendDiscriminateStateoftheart2021} were filtered using a majority vote filter of 15 seconds. Figure~\ref{fig:actionformer_confmats} provides confusion matrices of the ActionFormer \citep{zhangActionFormerLocalizingMoments2022} being applied using inertial, camera and combined (inertial + camera) features. Figure~\ref{fig:add_confmats} provides Confusion matrices of the shallow DeepConvLSTM \citep{bockImprovingDeepLearning2021} and improved Attend-and-Discriminate \citep{abedinAttendDiscriminateStateoftheart2021} applied on inertial data. Figure~\ref{fig:vizall} shows a color-coded visualisation of predictions streams of all models mentioned in the results table of the main paper. Figure~\ref{fig:oracleconfmats} delivers a side-by-side comparison of the confusion matrices of all models involved in the \emph{Oracle}-late-fusion-approach analysis mentioned in the main paper. Figure~\ref{fig:oracleconfmats} shows that a joint learning of both modalities particularly improves differentiation between the NULL-class and the activity classes resulting in better action boundaries, i.e. higher mAP scores, and classification scores.

\begin{figure}
\begin{center}
   \includegraphics[width=1\linewidth]{figures/actionformer_confmats.pdf}
\end{center}
   \caption{Confusion matrices of the ActionFormer \citep{zhangActionFormerLocalizingMoments2022} being applied using only inertial, vision (camera) and both combined (inertial + camera).}
\label{fig:actionformer_confmats}
\end{figure}

\begin{figure}
\begin{center}
   \includegraphics[width=.7\linewidth]{figures/inertial_confmats.pdf}
\end{center}
   \caption{Confusion matrices of the shallow DeepConvLSTM \citep{bockImprovingDeepLearning2021} and improved Attend-and-Discriminate \citep{abedinAttendDiscriminateStateoftheart2021}.}
\label{fig:add_confmats}
\end{figure}

\begin{figure}
\begin{center}
   \includegraphics[width=0.9\linewidth]{figures/activity_stream_all.pdf}
\end{center}
   \caption{Color-coded comparison of the ground truth data (top row) with the shallow DeepConvLSTM \citep{bockImprovingDeepLearning2021}, improved Attend-and-Discriminate \citep{abedinAttendDiscriminateStateoftheart2021}, ActionFormer \citep{zhangActionFormerLocalizingMoments2022} and TriDet \citep{shiTriDetTemporalAction2023} model on varying input modalities.}
\label{fig:vizall}
\end{figure}

\begin{figure}
\begin{center}
   \includegraphics[width=1.\linewidth]{figures/oracle_confmats.pdf}
\end{center}
   \caption{Confusion matrices of the best (1) inertial model (Attend-and-Dicriminate \citep{abedinAttendDiscriminateStateoftheart2021}), (2) vision model (TriDet \citep{shiTriDetTemporalAction2023}) and (3) combined model (vision + inertial) compared with (4) an \emph{Oracle}-combination of the inertial and camera as well (5) \emph{Oracle}-combination of the previous oracle with the combined approach.}
\label{fig:oracleconfmats}
\end{figure}

\clearpage

\section{Full-Text Recording Plan}
\label{sec:recordingplan}

\begin{figure}[H]
    \begin{center}
        \fbox{\includegraphics[page=1, height=1.2\linewidth]{figures/recording_plan.pdf}}
    \end{center}
    \caption{First page of the recording plan of the WEAR dataset.}
    \label{fig:recordingplan1}
\end{figure}

\begin{figure}
    \begin{center}
        \fbox{\includegraphics[page=2, height=1.2\linewidth]{figures/recording_plan.pdf}}
    \end{center}
    \caption{Second page of the recording plan of the WEAR dataset. Note that pictures are blurred for anonymization and are short video-clips in the original document.}
    \label{fig:recordingplan2}
\end{figure}

\begin{figure}
    \begin{center}
        \fbox{\includegraphics[page=3, height=1.2\linewidth]{figures/recording_plan.pdf}}
    \end{center}
    \caption{Third page of the recording plan of the WEAR dataset. Note that pictures are blurred for anonymization and are short video-clips in the original document.}
    \label{fig:recordingplan3}
\end{figure}

\begin{figure}
    \begin{center}
        \fbox{\includegraphics[page=4, height=1.2\linewidth]{figures/recording_plan.pdf}}
    \end{center}
    \caption{Fourth page of the recording plan of the WEAR dataset. Note that pictures are blurred for anonymization and are short video-clips in the original document.}
    \label{fig:recordingplan4}
\end{figure}

\begin{figure}
    \begin{center}
        \fbox{\includegraphics[page=5, height=1.2\linewidth]{figures/recording_plan.pdf}}
    \end{center}
    \caption{Fifth page of the recording plan of the WEAR dataset. Note that pictures are blurred for anonymization and are short video-clips in the original document.}
    \label{fig:recordingplan5}
\end{figure}

\clearpage 
{\small
\bibliographystyle{plainnat}
\bibliography{main}
}